%% file: arxiv.tex
\documentclass{article}

\usepackage[utf8]{inputenc} %
\usepackage[T1]{fontenc}    %
\usepackage{microtype}
\usepackage{graphicx}
\usepackage{booktabs} %
\usepackage[aboveskip=2pt]{subcaption} %

\usepackage{hyperref}

\usepackage[accepted]{icml2023}

\usepackage[utf8]{inputenc} %
\usepackage[T1]{fontenc}    %
\usepackage{url}            %
\usepackage{booktabs}       %
\usepackage{amsfonts}       %
\usepackage{nicefrac}       %
\usepackage{microtype}      %
\usepackage{xcolor}         %
\usepackage{amsmath}
\usepackage{mathrsfs}
\usepackage{amsthm}
\usepackage{amssymb}
\usepackage{array}
\usepackage{graphicx} 
\usepackage{subcaption}
\usepackage[nameinlink,capitalize]{cleveref}

\usepackage[capitalize,nameinlink]{cleveref}
\crefname{section}{Sec.}{Secs.}
\crefname{appendix}{App.}{Apps.}
\crefname{algorithm}{Alg.}{Algs.}

\usepackage{tikz,pgfplots}
\pgfplotsset{compat=1.16}
\usetikzlibrary{patterns,shapes.arrows,calc,arrows.meta}
\usetikzlibrary{shapes.geometric}
\usetikzlibrary{external}
\usepackage{ifthen}

\newlength{\figurewidth}
\newlength{\figureheight}

\pgfdeclareplotmark{mystar}{
    \node[star,star point ratio=2.25,minimum size=7pt,
          inner sep=0pt,draw=white,solid,fill=black,rounded corners=.5pt, anchor=center] {};
}

\pgfplotsset{every axis/.append style={
		legend style={inner xsep=1pt, inner ysep=0.5pt, nodes={inner sep=1pt, text depth=0.1em},draw=none,fill=none}
}}    

\newcommand*{\lmlVIf}{\mathcal{L}_{\text{VI}}}
\newcommand*{\lmlVIs}{\mathcal{L}_{\text{VI}}^\text{sparse}}
\newcommand*{\lmlEPf}{\mathcal{L}_{\text{EP}}}
\newcommand*{\lmlEPs}{\mathcal{L}_{\text{EP}}^\text{sparse}}
\renewcommand{\mid}{\,|\,}

\renewcommand{\paragraph}[1]{{\bf #1}~~}

\usepackage{xspace}
\newcommand{\eg}{\textit{e.g.\@}\xspace}
\newcommand{\ie}{\textit{i.e.\@}\xspace}
\newcommand{\cf}{\textit{cf.\@}\xspace}

\newcommand{\etal}{\textit{et~al.\@}\xspace}

\usepackage{bm}
\newcommand{\mathbold}[1]{\bm{#1}}
\newcommand{\mbf}[1]{\mathbf{#1}}

\newcommand{\vtheta}[0]{\mathbold{\theta}}
\newcommand{\veta}[0]{\mathbold{\eta}}
\newcommand{\vxi}[0]{\mathbold{\xi}}
\newcommand{\vxiu}[0]{\vxi_{\vu}}
\newcommand{\vlambda}[0]{\mathbold{\lambda}}
\newcommand{\vlambdau}[0]{\vlambda_{\vu}}
\newcommand{\vzeta}[0]{\mathbold{\zeta}}
\newcommand{\vzetau}[0]{\mathbold{\zeta}_{\vu}}

\newcommand{\vmu}[0]{\mathbold{\mu}}

\newcommand{\valpha}[0]{\mathbold{\alpha}}
\newcommand{\vbeta}[0]{\mathbold{\beta}}

\newcommand{\MLambda}[0]{\mathbold{\Lambda}}

\newcommand{\MZ}{\mbf{Z}}

\newcommand{\MX}{\mbf{X}}
\newcommand{\MA}{\mbf{A}}
\newcommand{\ML}{\mbf{L}}
\newcommand{\MK}{\mbf{K}}
\newcommand{\MS}{\mbf{S}}
\newcommand{\MT}{\mbf{T}}

\newcommand{\vz}{\mbf{z}}
\newcommand{\vf}{\mbf{f}}
\newcommand{\vm}{\mbf{m}}
\newcommand{\vu}{\mbf{u}}
\newcommand{\vx}{\mbf{x}}
\newcommand{\vy}{\mbf{y}}
\newcommand{\va}{\mbf{a}}
\newcommand{\vk}{\mbf{k}}

\newcommand{\diff}{\,\mathrm{d}}
\newcommand{\diag}{\operatorname{diag}}
\newcommand{\KL}[2]{\operatorname{D_{\mathrm{KL}}}\big[#1 \,\big\|\, #2 \big]}
\newcommand{\epsite}[1]{t_{#1}(f_{#1}; \mathbold{\zeta}_{#1})}
\newcommand{\dualsite}[1]{t_{#1}(f_{#1}; \mathbold{\lambda}_{#1})}
\newcommand{\posterior}{p(\vf  \mid \mathbf{y}; \vtheta)}
\newcommand{\prior}{p(\vf ;\vtheta)}
\newcommand{\likelihoodi}{p(y_i \mid f_i ;\vtheta)}
\newcommand{\likelihoodfull}{p(\vy \mid \vf ;\vtheta)}

\newcommand{\MKuu}{\MK_{\vu\vu}}

\newcommand\rurl[1]{\href{http://#1}{\nolinkurl{#1}}}

\usepackage{tabularx}

\begin{document}

\tikzexternaldisable

\twocolumn[

\icmltitlerunning{Improving Hyperparameter Learning under Approximate Inference in Gaussian Process Models}
\icmltitle{Improving Hyperparameter Learning under Approximate Inference \\ in Gaussian Process Models}

\begin{icmlauthorlist}
\icmlauthor{Rui Li}{aalto}
\icmlauthor{ST John}{aalto}
\icmlauthor{Arno Solin}{aalto}
\end{icmlauthorlist}

\icmlaffiliation{aalto}{Department of Computer Science, Aalto University, Finland, and Finnish Center for Artificial Intelligence (FCAI)}

\icmlcorrespondingauthor{Rui Li}{rui.li@aalto.fi}

\icmlkeywords{Machine Learning, Gaussian Processes, Approximate Inference, Hyperparameter Learning}

\vskip 0.3in
]

\printAffiliationsAndNotice{}  %

\begin{abstract}
Approximate \emph{inference} in Gaussian process (GP) models with non-conjugate likelihoods gets entangled with the \emph{learning} of the model hyperparameters.
We improve hyperparameter learning in GP models and focus on the interplay between variational inference (VI) and the learning target. While VI's lower bound to the marginal likelihood is a suitable objective for inferring the approximate posterior, we show that a direct approximation of the marginal likelihood as in Expectation Propagation (EP) is a better learning objective for hyperparameter optimization. We design a hybrid training procedure to bring the best of both worlds: it leverages conjugate-computation VI for inference and uses an EP-like marginal likelihood approximation for hyperparameter learning. We compare VI, EP, Laplace approximation, and our proposed training procedure and empirically demonstrate the effectiveness of our proposal across a wide range of data sets.\looseness-1

\end{abstract}

\section{Introduction}

Gaussian processes \citep[GPs,][]{gpbook} provide a plug-and-play approach for inference and learning, with principled ways of incorporating prior knowledge over functions and quantifying uncertainty. %
While GP regression under a conjugate (Gaussian) likelihood can be carried out elegantly in closed form, we focus on the non-conjugate case, where exact inference is intractable. 
Training of the GP consists of \emph{inferring} the approximate posterior and \emph{learning} the hyperparameters of the model. For clarity, by \emph{training} we refer to the combination of inference and learning.\looseness-1

\begin{figure}[t!]
  \centering\scriptsize
  \setlength{\figurewidth}{.87\columnwidth}
  \setlength{\figureheight}{.33\figurewidth}
  \pgfplotsset{scale only axis,axis on top,y tick label style={rotate=90,font=\tiny},x tick label style={font=\tiny},grid style={line width=.1pt, draw=gray!10,dashed},grid,ytick={-.4,-.3,-.2,-.1}}
  \pgfplotsset{clip mode=individual}
  \begin{subfigure}[t]{\columnwidth}
    \pgfplotsset{xticklabels = {}, ylabel={\parbox{3cm}{\centering Marginal likelihood \\ $\log p(\vy; \theta)$}}}
    \input{./fig/tikz/fig1a.tex}
  \end{subfigure}\\[-30pt]
  \begin{subfigure}[t]{\columnwidth}
    \pgfplotsset{xlabel={Characteristic lengthscale, $\log \theta$}, ylabel={\parbox{3cm}{\centering Predictive density \\ $\log p(\vy_*; \theta)$}}}
    \input{./fig/tikz/fig1b.tex}
  \end{subfigure}\\[-8pt]
  \caption{{\bf Practical benefits} on {\sc Ionosphere}: Marginal likelihood (top) acts as a training proxy for predictive density (bottom) of unseen future data. Our training objective produces a better point for prediction that also matches the MCMC baseline.\looseness-1}
  \label{fig:teaser}
\end{figure}
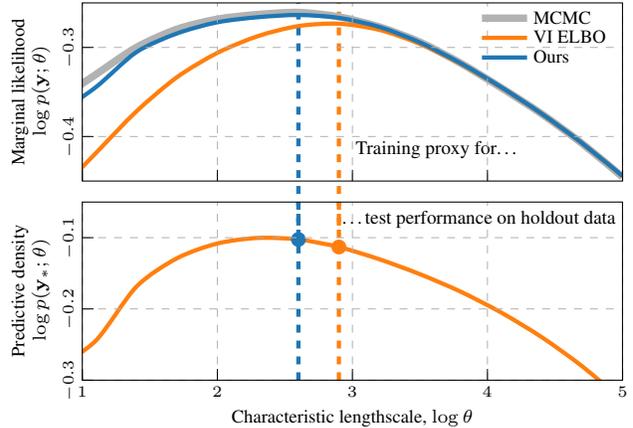

In a supervised learning setting, GPs are typically trained to optimize performance on the training samples \citep[as in empirical risk minimization,][]{vapnick1998statistical}.
Under the GP paradigm, the go-to solution to learning is finding $\vtheta^\star$ that maximizes the marginal likelihood. The marginal likelihood summarizes the probability that we would generate the observations $\vy$ with the model parameters $\vtheta$ if we would sample over the prior. It is formed by {\em marginalizing} over the latent functions from the GP prior, thus also known as the {\em evidence}. Even if this does not capture all aspects of {\em generalization} \citep[see discussion in][]{vehtari2016bayesian,lotfi2022bayesian}, it is still used as a practical proxy for performance on unseen test points (see \cref{fig:teaser} for the proxy and test performance on the {\sc ionosphere} benchmark data set).%

\begin{figure*}[t!]
\centering\footnotesize
\setlength{\figurewidth}{0.17\textwidth}
\setlength{\figureheight}{1.\figurewidth}
\pgfplotsset{grid style={dashed, white},scale only axis,xlabel near ticks, tick align=inside, ylabel near ticks, axis on top, ticklabel style = {font=\tiny, inner sep=3pt}, ytick={-1,1,3,5}, xtick={-1,1,3,5},ylabel style={yshift=-1em, align=center}, grid=both, minor tick num=1}
\tikzexternalenable  
\tikzsetnextfilename{main-figure0} 
\begin{tikzpicture}[inner sep=0, outer sep=0, remember picture]

  \def\data{./fig/tikz/ionosphere}
  \def\myxshift{0.6cm}
  \def\myxscaler{1.1}
  \def\myyscaler{-1.1}

  \foreach \x/\name [count=\i from 0] in {MCMC_mean_lml/MCMC,LA_mean_lml/LA, EP_mean_lml/EP,CVI_mean_lml/VI,CVI_mean_lml_ep/Ours} {
    \pgfplotsset{title=\name,ylabel={},xticklabels={}}
    \ifthenelse{\i > 0}{\pgfplotsset{yticklabels={}}}{}
    \begin{scope}[shift={(\myxscaler*\i*\figurewidth+\myxshift,0)}]
      \input{\data/\x.tex}
    \end{scope}
  }

  \node[rotate=90,align=center] at (-.05\figureheight,.5\figureheight) {\textbf{Marginal likelihood}\\[.6ex]$\log \sigma$};
  \node[rotate=90,align=center] at (-.05\figureheight,-.6\figureheight) {\textbf{Predictive density}\\[.6ex]$\log \sigma$};

  \foreach \x [count=\i from 0] in {MCMC_nlpd,LA_nlpd,EP_nlpd} {
    \pgfplotsset{title={},xlabel={$\log \ell$}}
    \ifthenelse{\i > 0}{\pgfplotsset{yticklabels={}}}{}
    \begin{scope}[shift={(\myxscaler*\i*\figurewidth+\myxshift,\myyscaler*\figureheight)}]
      \input{\data/\x.tex}
    \end{scope}
  }  

  \pgfplotsset{title={},xlabel={$\log \ell$}}
  \begin{scope}[shift={(\myxscaler*3.5*\figurewidth+\myxshift,\myyscaler*\figureheight)}]
    \input{\data/CVI_nlpd.tex}
  \end{scope}

  \draw[blue, shorten >=.2cm,shorten <=.2cm,-Stealth] (CVI_mean_lml_ep_t) to[out=250, in = 50] (CVI_mean_lml_ep_b);
  \draw[blue, shorten >=.2cm,shorten <=.2cm,-Stealth] (CVI_mean_lml_t) to[out=250, in = 200] (CVI_mean_lml_b);
\end{tikzpicture}%
\vspace*{-1em}
\caption{\textbf{Log marginal likelihood / predictive density surfaces} for the {\sc ionosphere} data set by varying kernel magnitude $\sigma$ and lengthscale $\ell$. The colour scale is the same in all plots: $-0.8$~\includegraphics[width=1cm,height=.65em]{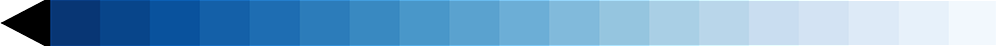}~$0$ (normalized by $n$). Optimal hyperparameters are shown by a black marker. EP and our EP-like marginal likelihood estimation match the MCMC baseline better than VI or LA, thus providing a learning proxy. For prediction, our method still leverages the same variational representation as VI.}
\label{fig:ionosphere_contour}
\tikzexternaldisable 
\end{figure*}

Under approximate inference, the marginalization step entangles the representation of the posterior with the learning target evaluation. The common approach is to assume an approximative Gaussian form for the posterior, so that the inference problem turns into finding a `good' parameterization for the Gaussian \citep[see][for a recent discussion on linearization/Gaussianization approaches]{wilkinson2021bayes}. The simplest approach is the so-called Laplace's approximation \citep[LA,][]{LA}, which uses a second-order Taylor approximation. It is efficient but not very accurate. Variational inference \citep[VI,][]{VI} and expectation propagation \citep[EP,][]{EP} are two commonly used approximate inference methods for non-conjugate GP models, which have complementary advantages: VI optimizes a lower bound of the marginal likelihood, is easy to implement, straightforward to use, and the convex optimization problem is guaranteed to converge. However, it is known to underestimate variance \citep{powerep}. EP on the other hand requires implementation-wise tuning per likelihood and is not guaranteed to converge \citep{epasawayoflife}. However, it does provide a good approximation for the marginal likelihood \citep{05classification, 08classification}.\looseness-1

For model performance on unseen test data, the learning of hyperparameters plays a crucial role. Thus we strongly advocate against the common practice of jointly optimizing variational and hyperparameters using the ELBO, as the training target is only representative for the variational parameters.
We build on work by \citet{cvi} and \citet{dual} that separate the learning of hyperparameters from inferring the variational parameters, and capture a link between VI and EP: the approximate posterior obtained through VI has exactly the same structure as the approximate posterior of EP. We obtain an EP-like marginal likelihood estimate from the VI approximate posterior for full and sparse GPs with no added computational cost. We propose a hybrid training procedure that combines the complementary advantages of natural-gradient VI and EP.\looseness-1

The contributions of this paper are as follows.
{\em (i)}~We improve generalizability in non-conjugate GP models with no extra computational cost by augmenting VI with an EP-like learning target for hyperparameter learning.
{\em (ii)}~We demonstrate our EP-like learning target is closer to an MCMC baseline and thus provides a better learning objective. We empirically compare the quality of the approximate marginal likelihood in LA, EP, and VI, and our proposed learning target.
{\em (iii)}~We show our method improves generalizability via experiments in binary classification for full and sparse GP models and robust regression.

\section{Approximate Inference}
\label{sec:background}

In this section, we review common approximate inference methods in Gaussian process (GP) models.
GP models put a GP prior over functions:\looseness-1
\begin{equation}
  \text{GP prior:} \quad f(\vx) \sim \mathcal{GP}(\mu(\vx), \kappa(\vx,\vx')),
\end{equation}  
where $\vx \in \mathcal{X} \subset \mathbb{R}^d$ is an input vector, $\mu(\vx)$ is the mean function, and $\kappa(\vx,\vx')$ is the covariance (kernel) function. This GP prior is linked to the data set $\mathcal{D}=(\MX, \vy)=\{(\vx_i, y_i)\}_{i=1}^n$ of input--output pairs through a likelihood function that maps the latent function value $f(\vx)$ to the observations. We assume the likelihood factorizes over observations:
\begin{equation}\textstyle
  \text{Likelihood:}\quad \vy \mid \vf \sim \prod_{i=1}^{n} p(y_i \mid f(\vx_i)).
\end{equation}
The posterior is given by $\posterior \propto \likelihoodfull \, \prior$, where $\vtheta$ denotes the model (hyper)parameters of the likelihood, mean function, and kernel, and $\vf $ is the vector of function values evaluated at the inputs. Prediction at a new test input $\vx_*$ is obtained by computing the predictive distribution $p(f(\vx_*)\mid\mathcal{D}, \vx_*)$.

\paragraph{Probabilistic inference} For (conjugate) Gaussian likelihoods, $p(y_i \mid f_i) = \mathrm{N}(y_i \mid f(\vx_i), \sigma_\text{n}^2)$, the posterior is available in closed form as a Gaussian distribution. For non-Gaussian likelihood models the inference problem needs to be approached with \emph{approximative} inference methods. Sampling schemes (see \cref{sec:mcmc} for our baseline solution) can tackle this, but for efficient inference one typically employs an approximative Gaussian posterior of the form
\begin{equation}\label{eq:posterior}
  \text{Approximate posterior:} \quad q(\vf) = \mathrm{N}(\vm,\MS).  
\end{equation}
Its `optimal' parameterization \citep{VI} is given in terms of $2n$ parameters $(\valpha,\vbeta)$ such that $\vm = \MK\valpha$ and $\MS = (\MK^{-1} + \diag(\vbeta))^{-1}$, where $\MK$ is an $n \times n$ matrix with $\kappa(\vx_i,\vx_j)$ as the $ij$\textsuperscript{th} entry.
The inference problem thus turns into (efficiently) finding a (good) representation of the posterior in terms of \cref{eq:posterior} by minimizing some measure of error. Typical approaches for this are the Laplace approximation (local linearisation of the problem), expectation propagation (approximately minimizing $\KL{p(\vf\mid\vy)}{q(\vf)}$ from approximate to true posterior), or variational inference (minimizing $\KL{q(\vf)}{p(\vf\mid\vy)}$). EP is expected to be the most accurate method \citep[see discussion in][]{vehtari2016bayesian} and Laplace to have the smallest computational overhead.

\paragraph{Learning under the GP paradigm} In probabilistic machine learning, `learning' typically amounts to finding point estimates for the hyperparameters $\vtheta$ in the likelihood, mean function, and kernel by optimizing w.r.t.\ the log marginal likelihood:
\begin{equation}
  \text{Learning target:}\quad
  \vtheta^\star = \arg \max_{\vtheta} \log p(\vy; \vtheta). 
\end{equation}
For Gaussian likelihoods, the marginal likelihood is available in closed form. For non-conjugate models, we can only optimize a proxy to the marginal likelihood $p(\vy; \vtheta) = \int \likelihoodfull \, \prior \diff\vf$, which depends on the approximate inference scheme and how it represents the posterior.

\subsection{Laplace Approximation (LA)}
A local Taylor expansion of the log posterior gives the Laplace approximation \citep[LA,][]{LA}. By defining
$ \Psi(\vf)=\log (p(\vy \mid \vf)\,\prior)$,
the approximate posterior $q(\vf)$ is obtained through a second-order Taylor expansion of $\Psi(\vf)$ around its maximum at $\hat{\vf}=\arg \max_{\vf} \Psi(\vf)$ (the posterior mode):
$\posterior \propto \exp(\Psi(\vf)) \approx \exp\big(\Psi(\hat{\vf})+\frac12 (\vf-\hat{\vf})^\top \nabla^2 \Psi(\vf) |_{\vf=\hat{\vf}} (\vf-\hat{\vf})\big)$. This is proportional to $\mathrm{N}(\vf \mid \hat{\vf}, \MA^{-1}) = q(\vf)$,
where $\MA= -\nabla^2 \Psi(\vf)|_{\vf=\hat{\vf}}$ is the Hessian of the negative log posterior at $\hat{\vf}$.
The log marginal likelihood is approximated as\looseness-1
\begin{multline}
	\textstyle\log p(\vy; \vtheta) = \log \int \exp (\Psi(\vf)) \diff \vf  \\
	\approx \textstyle\log \int \exp\big(\Psi(\hat{\vf})-\frac12 (\vf-\hat{\vf})^\top \MA(\vf-\hat{\vf})\big) \diff \vf .
\end{multline}

\subsection{Expectation Propagation (EP)}
Expectation Propagation \citep[EP,][]{EP} is based on an approximation $q(\vf)$ that factorizes in the same way as the target posterior $\posterior \propto \prior \prod_{i=1}^n \likelihoodi$: each likelihood term is approximated with a \emph{site} function $\epsite{i}$, and
\begin{equation}\label{eq:full_ep_post}
	\textstyle q(\vf ; \vtheta, \vzeta) \propto \prior \prod_{i=1}^n \epsite{i}.
\end{equation}
For GP models, the sites $\epsite{i}$ are chosen to be (unnormalized) Gaussians, and hence the global approximation $q(\vf)$ is also Gaussian.
EP aims to minimize $\KL{\posterior}{q(\vf ; \vtheta, \vzeta)}$ w.r.t.\ $\vzeta$. This KL cannot be computed directly. Instead, EP updates the sites in an iterative fashion; the parameters of one site $\vzeta_i$ are tuned by minimizing the local Kullback--Leibler divergence
\begin{multline}\label{eq:full_ep_infer_obj}
	\textstyle\KL{\likelihoodi \, \prior \, \prod_{j \neq i} \epsite{j}\\}{\epsite{i}\,\prior\textstyle\prod_{j \neq i} \epsite{j}} ,
\end{multline}
where in the first argument the $n-1$ other likelihood terms have been replaced by their current site approximation. The optimal values of $\vzeta_i$ in this step can be determined by matching the first two moments.
This iterative process often works well in practice, but can be numerically unstable (\eg, for Student-$t$ likelihood) and is not guaranteed to converge in the general case \citep[see][]{epasawayoflife}.

The log marginal likelihood is directly approximated as
\begin{equation}\label{eq:full_ep_energy}
	\log p(\mathbf{y}; \vtheta) 
	\approx \lmlEPf (\vzeta, \vtheta)
	= \log \int p(\vf ;\vtheta ) \prod_{i=1}^n \epsite{i} \diff\vf ,
\end{equation}	
which is known to lead to a good objective for learning hyperparameters \citep[see][]{aki}.

\subsection{Variational Inference (VI)} 
\label{sec:VI}
Variational Inference \citep[VI,][]{VI} approximates the GP posterior $\posterior$ with a Gaussian distribution $q(\vf; \vxi)$ parameterized by $\vxi$. VI minimizes the reverse KL $\KL{q(\vf; \vxi)}{\posterior}$ by maximizing the following evidence lower bound (ELBO):
\begin{multline}\label{eq:full_elbo}
\!\!\!\!\! \log p(\vy; \vtheta) \geq \lmlVIf(\vxi, \vtheta ) = \sum_{i=1}^n \mathbb{E}_{q(f_i; \vxi_i)}\big[\log p(y_i \mid f_i; \vtheta)\big]\\ -\KL{q(\vf ; \vxi)} {p(\vf ; \vtheta)},
\end{multline}
w.r.t.\ variational parameters $\vxi$. VI optimizes a lower bound on the marginal likelihood, so is guaranteed to converge, which is a strength over EP.
As known from \citet{hensman2013gaussian} and motivated by \citet{khan2013fast}, it has been desirable to {\em not} use the optimal parameterization in terms of $2n$ parameters, as the resulting optimization problem is non-convex. Instead, it is common to declare a variational distribution over the full posterior, $q(\vf; \vxi)=\mathrm{N}(\vm, \MS)$, and optimize the ELBO w.r.t.\ this mean--covariance parameterization\footnote{In practice it may be beneficial to optimize in the \emph{whitened} (or \emph{non-centered}) parameterization $\vxi = (\vm', \MS')$ s.t.~$\vm = \ML \vm'$ and $\MS = \ML \MS' \ML^\top$, where $\ML = \operatorname{Cholesky}(\MK)$ \citep{gorinova20a}.} $\vxi = (\vm, \MS)$ %
using a general-purpose optimizer \citep[\eg\ Adam,][]{adam}.

In practice, the same lower bound $\lmlVIf(\vxi, \vtheta)$ is used to optimize variational parameters as well as hyperparameters, \ie, inference and learning are coupled into a single optimization.
This approach is commonplace, even though it is well-known to result in biased hyperparameters \citep{05classification, 08classification, powerep}.

\section{Learning in the Dual Parameterization}
\label{sec:methods}
We design a hybrid training procedure that augments VI with an EP-like learning target for hyperparameter learning.
Our work builds upon the dual parameterization \citep{cvi}. Because the Gaussian distribution is part of the exponential family, we can write the approximate posterior as $q(\vf) = \mathrm{N}(\vm, \MS) = \exp \big(\veta^{\top} \MT(\vf )-a(\veta)\big)$, where $\veta=(\MS^{-1} \vm,-\frac12 \MS^{-1})$, $\MT(\vf )=(\vf , \vf  \vf ^{\top})$ are the sufficient statistics, and $\exp(-a(\veta))$ is a normalization term. This leads to two additional parameterizations of $q(\vf)$: using the natural parameters $\veta$, or using the expectation parameters $\vmu=\mathbb{E}_{q(\vf)}[\MT(\vf)]=(\vm, \MS+\vm \vm^{\top})$. 

\citet{cvi} showed that in the natural parameterization of the approximate posterior, {\em natural gradient descent} \citep[NGD,][]{amari1998natural} (in the natural parameters space $\veta$) will have the same computational cost as ordinary gradient descent on $\vxi = (\vm, \MS)$.
The approximate posterior under this parameterization is  %
\looseness-1
\begin{equation} \label{eq:full_vi_post}
	q(\vf ; \vlambda, \vtheta) \propto p(\vf ; \vtheta) \textstyle \prod_{i=1}^n \underbrace{\exp{\langle \vlambda_i, \MT(f_i)\rangle}}_{\triangleq~\dualsite{i}},	
\end{equation}
where $\vlambda_i=\nabla_{\vmu_i} \mathbb{E}_{q(f_i; \vlambda_i,\vtheta)}[\log \likelihoodi]$. 
The natural parameters of the approximate posterior $q(\vf)$ are $\veta=\vlambda_0 + \vlambda$, where $\vlambda_0=(\mathbold{0}, -\frac{1}{2} \MK^{-1})$ are the natural parameters of the prior $\prior$ and $\vlambda$ are the parameters of the likelihood approximation term $t(\vf)$. Then, we could also parameterize the approximate posterior with $\vlambda$, to which we refer as the `dual' $\vlambda$ parameterization.

Crucially, the approximate posterior \eqref{eq:full_vi_post} has the same form as its EP counterpart \eqref{eq:full_ep_post}. This links EP with VI, which is the starting point for our proposed learning objective. The similarity {\it per se} has been visible in, \eg, \citet{chang2020fast,dual}, but it had not been explored further.\looseness-1

\begin{algorithm}[t!]
\caption{Training procedure for improved hyperparameter learning by a VEM-style iteration.}
\label{alg:training_procedure}
\renewcommand{\algorithmicrepeat}{\textbf{do}}   
\renewcommand{\algorithmicuntil}{\textbf{while}}   
\begin{algorithmic}
\STATE  Initialize var.\ parameters $\vlambda^{(0)}$ and hyperparameters $\vtheta^{(0)}$ 
\STATE  Specify total training iteration $K$, the number of E-steps and M-steps per iteration $J_\text{E}$, $J_\text{M}$
\FOR{$k=1, 2, \ldots, K$}

    \STATE \emph{With $J_\text{E}$ nat.\ grad.\ steps and learning rate $\rho_\text{E}$, optimize}
        \STATE $\quad\vlambda^{(k)} \leftarrow \arg \max _{\vlambda} \lmlVIf \big(\vlambda, \vtheta^{(k-1)} \big)$ 

    \STATE \emph{With $J_\text{M}$ grad.\ steps and learning rate $\rho_\text{M}$, optimize}
        \STATE $\quad\vtheta^{(k)} \leftarrow \arg \max _{\vtheta} \lmlEPf \big(\vlambda^{(k)}, \vtheta \big)$ 
    \IF{$\lmlEPf \big(\vlambda^{(k)}, \vtheta^{(k)} \big) < \lmlEPf \big(\vlambda^{(k-1)}, \vtheta^{(k-1)} \big) $}
       \STATE \textbf{return} $\vlambda^{(k-1)}, \vtheta^{(k-1)}$
    \ENDIF
\ENDFOR
\STATE \textbf{return} $\vlambda^{(k)}, \vtheta^{(k)}$
\end{algorithmic}
\end{algorithm}

\begin{figure*}[t!]
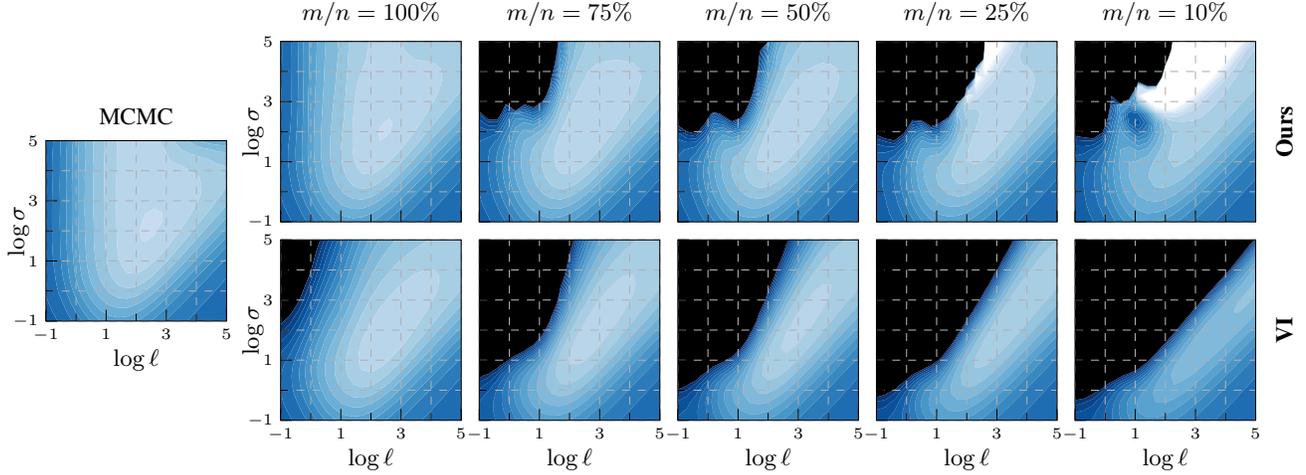

\centering\footnotesize
\setlength{\figurewidth}{0.14\textwidth}
\setlength{\figureheight}{1.\figurewidth}
\pgfplotsset{grid style={dashed, white},scale only axis,xlabel near ticks, tick align=inside, ylabel near ticks, axis on top, ticklabel style = {font=\tiny, inner sep=3pt}, ytick={-1,1,3,5}, xtick={-1,1,3,5},ylabel style={yshift=-1em, align=center}, grid=both, minor tick num=1}
\tikzexternalenable  
\tikzsetnextfilename{main-figure1} 
\begin{tikzpicture}[inner sep=0, outer sep=0, remember picture]

  \def\data{./fig/tikz/ionosphere}
  \def\myxshift{0.6cm}
  \def\myxscaler{1.1}
  \def\myyscaler{-1.1}

  \foreach \x/\name [count=\i from 0] in {MCMC_Z_mean_lml/MCMC} {
    \pgfplotsset{title=\name,xlabel={$\log\ell$},ylabel={$\log\sigma$}}
    \begin{scope}[shift={(\myxscaler*\i*\figurewidth+\myxshift-0.2\figurewidth,0.5*\myyscaler*\figureheight)}]
      \input{\data/\x.tex}
    \end{scope}
  }

  \foreach \x/\name [count=\i from 1] in {CVI_Z_mean_lml_ep/100\%, epcvi_Z_3/75\%, epcvi_Z_2/50\%, epcvi_Z_1/25\%, epcvi_Z_0/10\%} {
    \pgfplotsset{title={$m/n=\name$},xticklabels={}}
    \ifthenelse{\i > 1}{\pgfplotsset{yticklabels={}}}{\pgfplotsset{ylabel={$\log\sigma$}}}
    \begin{scope}[shift={(\myxscaler*\i*\figurewidth+\myxshift,0)}]
      \input{\data/\x.tex}
    \end{scope}
  }
  \node[rotate=90,align=center] at (7.0\figureheight,.5\figureheight) {\textbf{Ours}\\[.6ex]};
  \node[rotate=90,align=center] at (7.0\figureheight,-.6\figureheight) {\textbf{VI}\\[.6ex]};

  \foreach \x/\name [count=\i from 1] in {CVI_Z_mean_lml/100\%, cvi_Z_3/75\%, cvi_Z_2/50\%, cvi_Z_1/25\%, cvi_Z_0/10\%} {
    \pgfplotsset{title={},xlabel={$\log \ell$}}
    \ifthenelse{\i > 1}{\pgfplotsset{yticklabels={}}}{\pgfplotsset{ylabel={$\log\sigma$}}}
    \begin{scope}[shift={(\myxscaler*\i*\figurewidth+\myxshift,\myyscaler*\figureheight)}]
      \input{\data/\x.tex}
    \end{scope}

 }
\end{tikzpicture}
\vspace*{-1em}
\caption{\textbf{Sparse approximation:} log marginal likelihood surfaces for the {\sc ionosphere} data set, changing the fraction $m/n$ of the number of inducing points $m$ vs.\ $n=351$ data points. The colour scale is the same in all plots: $-0.8$~\includegraphics[width=1cm,height=.65em]{fig/colourbar.png}~$0$; values below the predefined range are plotted as black. For moderate sparsification, our EP-like marginal likelihood estimation (top) matches the full MCMC baseline better than VI (bottom). For extreme sparsification ($10\%$: $m=35$), neither approximation resembles the full surface.\looseness-1}
\label{fig:reduce_Z}
\tikzexternaldisable  
\end{figure*}

\subsection{Our Proposed Objective for Learning} 
\label{sec:objective}
Natural gradient descent can efficiently optimize the variational parameters $\vlambda$, and we can combine it with another optimizer for the hyperparameters, leading to a natural separation of the inference and learning steps \citep{salimbeni2018natural}. As discussed by \citet{dual}, this can be seen as a Variational Expectation--Maximization (VEM) procedure. Under this setup, inference/learning is performed by alternating between optimizing the variational distribution in the $\vlambda$ parameterization and taking gradient steps for finding $\vtheta$ by iterating the following steps at the $k$\textsuperscript{th} iteration:
\begin{equation*}
\begin{aligned}
&\text{E-step (inference):} \quad \vlambda^{(k+1)} \leftarrow \arg \max _{\vlambda} \mathcal{L}_\mathrm{E}(\vlambda, \vtheta^{(k)}), \\
&\text{M-step (learning):} \quad \vtheta^{(k+1)} \leftarrow \arg \max _{\vtheta} \mathcal{L}_\mathrm{M}(\vlambda^{(k+1)}, \vtheta),
\end{aligned}
\end{equation*}
where the objective for both the inference and learning steps is the ELBO in \cref{eq:full_elbo} under the $\vlambda$ parameterization: $\mathcal{L}_\mathrm{E} \equiv \mathcal{L}_\mathrm{M} \equiv \lmlVIf$. Note: Even if the parameterization and optimization procedure are different, the inference and learning objective are the same as in \cref{sec:VI} and typically expected to converge at the same optima.

VEM deals with the variational inference problem by casting inference into an optimization problem that is solved by NGD, which appears both principled and efficient. We conjecture that the ELBO in \cref{eq:full_elbo} is not the best objective. Conveniently, the dual parameterization in the VI posterior \cref{eq:full_vi_post} is formed as a product of the prior and Gaussian sites $\dualsite{i}$ just as in EP (\cf\ \cref{eq:full_ep_post}). This provides a representation of the posterior that is directly EP-like and allows us to estimate the log marginal likelihood by plugging $\vlambda_i$ from \cref{eq:full_vi_post} into $\vzeta_i$ in \cref{eq:full_ep_energy}:
\begin{equation}\label{eq:epobj}
\mathcal{L}_\mathrm{M} \equiv \mathcal{L}_\text{EP} (\vlambda, \vtheta)= \log \!\!\int\!\! \prior \textstyle\prod_{i=1}^n \dualsite{i} \diff \vf ,
\end{equation}
giving the target for the M-step.

Our hybrid training procedure uses the variational objective in the E-step to ensure a good representation of the posterior. Then in the M-step, we use an EP-like (and thus closer to marginal likelihood) objective for hyperparameter learning, at no additional computational cost. The algorithm is sketched out in \cref{alg:training_procedure}. Although our procedure requires implementing two training objectives, this is not likelihood-specific and has minimal implementation overhead.

\section{Sparse Approximation for Large Data Sets}
\label{sec:sparse}
Regardless of conjugacy, inference in GP models for large-scale data sets is challenging due to an $\mathcal{O}(n^3)$ computational bottleneck. In this section, we extend our hybrid training procedure to the sparse case.

A common approach to tackle this scalability issue is to summarize the information contained in the original data set into a smaller but more effective \emph{pseudo}-data set, making the computational complexity tractable \citep[see \cref{fig:reduce_Z} and][]{05overview}. The pseudo-inputs are referred to as inducing points and denoted as $\MZ=\{\vz_{i}\}_{i=1}^{m}$, where $m \ll n$ \citep{seeger2003bayesian, csato2002gaussian, 05overview, williams2001using}. The pseudo-outputs are referred to as inducing variables and denoted as $\vu=f(\MZ)$. \looseness-1

One common choice for the form of approximate posterior, as first introduced in \citet{VFE}, is $q(\vf, \vu; \vxiu, \vtheta) = p(\vf \mid \vu; \vtheta)\, q(\vu;\vxiu)$, where $p(\vf \mid \vu)$ is the GP conditional and $q(\vu;\vxiu)=\mathrm{N}(\vm_{\vu}, \MS_{\vu})$ the approximate posterior in $\vu$. In this form, $\vy$ can only affect $\vf$ through $\vu$, which means the information in the original data set is summarized in $\vu$. The marginal posterior over the function $f(\cdot)$ is
\begin{equation}\label{eq:svgp_marginals}
  q_{\vu}(f(\cdot); \vxiu, \vtheta)= \int p(f(\cdot) \mid \vu; \vtheta) \, q(\vu; \vxiu) \, \diff \vu , 
\end{equation}
where $p(f(\cdot) \mid \vu;\vtheta)$ is the distribution of the GP prior conditioned on $f(\MZ) = \vu$. Variational parameters $\vxi_{\vu}=(\vm_{\vu}, \MS_{\vu})$ can be inferred by optimizing the following ELBO:
\begin{multline}\label{eq:sparse_elbo}
	\log p(\vy; \vtheta) \geq \lmlVIs(\vxiu, \vtheta) \\ =\textstyle \sum_{i=1}^n \mathbb{E}_{q_{\vu}(f_i; \vxi_{\vu}, \vtheta)}[\log p(y_i \mid f_i; \vtheta)] \\ -\KL{q(\vu;\vxiu)}{p(\vu;\vtheta)},
\end{multline}
where $q_{\vu}(f_i; \vxi_{\vu}, \vtheta)=\mathrm{N}(f_i \mid \va_i^{\top} \vm_{\vu}, \kappa_{i i}-\va_i^{\top}(\mathbf{K}_{\mathbf{u u}}-\mathbf{S}_{\vu}) \va_i)$, $\va_i=\MKuu^{-1} \vk_{\vu,i}$, $\MKuu = \kappa(\MZ, \MZ)$, and $\vk_{\vu,i}=\kappa(\MZ, \vx_i)$ for the $i$\textsuperscript{th} data sample. 
Now under the dual parameterization, we denote the converged dual parameters of the posterior marginal $q_{\vu}^*(f_i)$ as $\vlambda_i^*$. \citet{dual} suggest designing a similar VEM procedure that exploits the structure of the $q(\vu)$ in terms of the $2n$ dual parameters. The natural parameters of $q^*(\vu)$ are \looseness-1
\begin{align}
    &(\MS_{\vu}^*)^{-1}\vm_{\vu}^* = \MK_{\vu\vu}^{-1} \underbrace{ \big( \textstyle\sum_{i=1}^{n} \vk_{\vu i} {\lambda}_{1,i}^*\big) }_{= \bar{\vlambda}_1^*} \\
    &(\MS_{\vu}^*)^{-1} = \MK_{\vu\vu}^{-1} + \MK_{\vu\vu}^{-1} , \underbrace{\big(\textstyle\sum_{i=1}^{n} \vk_{\vu i} {\lambda}_{2,i}^* \vk_{\vu, i}^\top \big)}_{\bar{\MLambda}^*_2} \MK_{\vu\vu}^{-1} , %
\end{align}
where the quantities $\MK_{\vu\vu}$ and $\vk_{\vu,i}$ directly depend on $\vtheta$ and we can express the ELBO as the partition function of a Gaussian distribution. 
\citet{dual} use a \emph{tied} parameterization \citep[motivated by site-tying in EP, see][]{powerep, li2015stochastic} that relaxes the need of storing all the $\{\vlambda_i^*\}_{i=1}^n$ and instead stores only $\bar{\vlambda}^*_1$ (length $m$) and $\bar{\MLambda}^*_2$ (size $m\times m$), which avoids the storage issue for large data sets. This extends the results of \citeauthor{cvi} to the sparse case where the resulting approximate posterior is
\tikzexternalenable
\begin{multline}\label{eq:sparse_vi_post}
  q(\vf, \vu; \vlambdau, \vtheta) \propto p(\vf \mid \vu; \vtheta)\,p(\vu;\vtheta) \\ \times \textstyle \prod_{i=1}^n \underbrace{\exp{\langle\vlambda_{\vu,i}, \MT(\va_i^{\top} \vu)\rangle}}_{\triangleq~t_i(\vu; \vlambda_{\vu,i})},
\end{multline}
where $\vlambda_{\vu,i}=\nabla_{\mathbold{\mu}_{\vu, i}} \mathbb{E}_{q_{\vu}(f_i; \vlambda_{\vu,i}, \vtheta)}[\log \likelihoodi]$. This gives rise to the sparse E-step for inference under the sparse VEM scheme, where $\mathcal{L}_\mathrm{E}^\text{sparse} \equiv \lmlVIs$.

\paragraph{Our proposed sparse objective for learning}
In EP, the tied representation for constraining the problem to a summary $\vzetau$ that scales in $m$ rather than $n$ gives rise to a sparse expectation propagation approach \citep{powerep}, where the log marginal likelihood is approximated as
\begin{align}\label{eq:sparse_ep_energy}
	&\log p(\vy;\vtheta) 
	\approx \lmlEPs (\vzetau, \vtheta) \nonumber =  \log \int q(\vf, \vu; \vzeta_{\vu}, \vtheta) \diff \vf \diff \vu \\
	&~= \log \int p(\vf \mid \vu;\vtheta) \, p(\vu; \vtheta) \prod_{i=1}^n {t_i(\vu;\vzeta_{\vu,i})} \diff \vf \diff \vu.%
\end{align}
Under dual parameterization VI, the approximate posterior \cref{eq:sparse_vi_post}  has the same structure as the EP approximate posterior in \cref{eq:sparse_ep_energy}. An EP-like estimate of the log marginal likelihood can thus be calculated by injecting $\vlambda_{\vu,i}$ from \cref{eq:sparse_vi_post} into $\vzeta_{\vu,i}$ in \cref{eq:sparse_ep_energy}, thus giving $\mathcal{L}_\mathrm{M}^\text{sparse} \equiv \lmlEPs(\vlambda_{\vu}, \vtheta)$ which is a sparse EP-like learning objective under sparse variational inference. Note: $\lmlEPs(\vlambda_{\vu}, \vtheta)$ has the same computational cost as VI.

\section{Experiments}
\label{sec:experiments}
We provide a range of experiments, in which we demonstrate {\em effectiveness} and {\em practicality} of the proposed approach, and highlight similarities and differences between learning under the three most common approximative inference methods (LA, EP, and VI).\looseness-1 

As the log marginal likelihood is a surrogate for the generalization ability of the model to unseen data, we evaluate the marginal likelihood estimations of different methods (\cref{sec:lml_quality}) to see whether our EP-like marginal likelihood provides a better learning target. We then evaluate our hybrid training on non-conjugate tasks in binary classification and Student-$t$ regression on small and mid-sized data sets (\cref{sec:benchmarks,sec:robust}). 

How the sparsity affects the learning target is not obvious, therefore in \cref{sec:reduce_Z} we first investigate the influence of sparse approximation on the learning target. We then evaluate our hybrid training procedure on binary classification tasks with sparse approximation. For all experiments in the main paper, we use an isotropic Mat\'ern-$\nicefrac52$ kernel. %
We also provide results under automatic relevance determination (ARD) with the same kernel (in \cref{app:classification_ablation}), where we only include results for data sets that could be confirmed to have converged.\looseness-1

We implement the variational methods in GPflow~\citep{GPflow:2017}, use reference implementations of LA and EP from GPy~\citep{gpy2014}, and base our MCMC implementation on the GPML toolbox \citep{rasmussen2010gaussian}. Additionally, we use the GPstuff toolbox \citep{vanhatalo2013gpstuff} for the custom LA and EP implementation for the Student-$t$ likelihood. We implement EP and VI convergence checks; details in \cref{app:classification_ablation}. \looseness -2

\begin{table*}[h!]
\centering
\caption{\textbf{Binary classification:} log predictive density (higher is better) on different data sets from the {\em Bayesian benchmarks} over 5-fold cross-validation with 10 different seeds. Best results and those not statistically significantly different from them under a paired $t$-test are \textbf{bolded}. We provide MCMC results for reference and exclude it from bolding. MCMC gives the best results on all data sets except Balloons. All inference methods perform well overall, while our training objective delivers the most reliable performance.}
\setlength{\tabcolsep}{7pt}
\scriptsize
\input{table/seed_test_lpd.tex}
\label{table:classification_full_lpd_10seed}
\end{table*}

\subsection{Quality of Marginal Likelihood Approximations}
\label{sec:lml_quality}
We compare the quality of marginal likelihood approximations of LA, VI, EP, and our EP-like VI with gold-standard MCMC. We demonstrate this on a binary classification task on the {\sc ionosphere} data set, with {\sc sonar}, {\sc usps}, {\sc parkinsons}, and {\sc monks-2} in \cref{app:additional_contour}. We estimate the log marginal likelihood on a $21 \times 21$ grid of values for the log hyperparameters $\log \vtheta=(\log \ell, \log \sigma)$ and plot the contour on the grid. For each hyperparameter setting, we fix the hyperparameters and evaluate the approximate log marginal likelihood based on the inferred approximate posterior.

\paragraph{Markov Chain Monte Carlo baseline} \label{sec:mcmc} MCMC is exact in the limit of long runs and thus provides a gold standard for log marginal likelihood estimation. \citet{05classification} and \citet{08classification} proposed a sampling scheme based on annealed importance sampling \citep[AIS,][]{AIS} for obtaining a good estimate of the marginal likelihood (see \cref{app:mcmc} for details). The baseline was computed by running $21 \times 21 =441$ jobs in parallel on a cluster. \looseness-1

\paragraph{Experiment results} As shown in the top row of \cref{fig:ionosphere_contour} on the {\sc ionosphere} benchmark data set, the marginal likelihood estimation of EP closely matches the MCMC baseline, whereas that of VI looks clearly different. Notably, when we estimate the marginal likelihood by $\mathcal{L}_\text{EP}(\vlambda,\vtheta)$ using the `site' parameters of dual VI (Ours), the contour shapes become much closer to the MCMC result, demonstrating the improvement of using this EP-like marginal likelihood estimation. To investigate whether the improved marginal likelihood estimation also leads to better generalization, we select the optimal hyperparameter location across the grid values and compare the log predictive density on the test set (bottom row of \cref{fig:ionosphere_contour}). The optimal hyperparameter location of EP-like VI (Ours) is closer to MCMC than VI and generalizes well. We show the same analysis for different data sets covering different types of classification tasks (from general classification to small images) in \cref{app:additional_contour}, with the same conclusion. In \cref{fig:sonar_contour,fig:usps_contour,fig:parkinsons_contour,fig:monks_contour}, VI and EP conform to their stereotypes of being over- and under-confident, respectively, while Ours tends to have slightly better calibration.

For completeness, and motivated by the seminal work of \citet{05classification}, we provide back-to-back comparisons of both marginal log likelihood and predictive density surfaces also for LA. In \cref{fig:ionosphere_contour}, the marginal likelihood surface of LA resembles that of VI, while for the predictive density surface VI more closely resembles MCMC compared to LA.

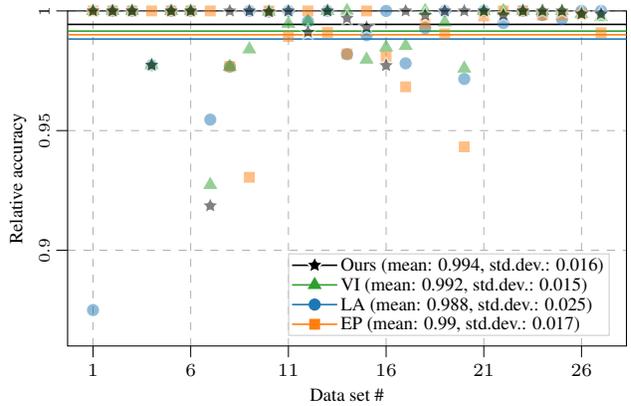
\begin{figure}[t!]
  \tikzexternaldisable
  \centering\scriptsize
  \setlength{\figurewidth}{.9\columnwidth}
  \setlength{\figureheight}{.6\figurewidth}
  \pgfplotsset{scale only axis,ylabel near ticks, y tick label style={rotate=90,font=\tiny},y tick label style={font=\tiny},xtick={1,6,11,16,21,26},grid style={line width=.1pt, draw=gray!10,dashed},grid,legend style={fill=white}}
  \input{./fig/tikz/classification_relative_acc.tex}\\[-1.5em]
  \caption{Mean relative accuracy compared to best method on each data set of \cref{table:classification_full_lpd_10seed} (over 5-fold CV repeated with 10 seeds). The horizontal lines indicate mean across all data sets; see legend for mean and standard deviation. Our approach yields reliable training, with the highest average relative accuracy and the least outliers.}
  \vspace*{-1em}
  \label{fig:relacc}
  \tikzexternalenable
\end{figure}

\subsection{Non-Conjugate Tasks in Bayesian Benchmarks}
\label{sec:benchmarks}
The log marginal likelihood is a surrogate for the generalization ability of the model to unseen data. To explicitly evaluate the generalization ability of our hybrid training procedure, we compare it against LA, EP, and VI on commonly used benchmark classification tasks. We use common small and mid-sized data sets ($n \leq 1000$) and do full GP inference with 5-fold cross-validation. We use the {\em Bayesian Benchmarks} suite (\rurl{github.com/secondmind-labs/bayesian_benchmarks}) for evaluating the methods.

\paragraph{Evaluation on binary classification}
We consider binary classification with a Bernoulli likelihood on 27 data sets from the UCI repository \citep{UCI}. For all approximate inference methods, we set the same number of maximum training iterations and use the relative changes in the parameters of the model as convergence criteria. For our hybrid training procedure due to the conflicting objectives in E- and M-steps discussed in \cref{sec:conflict_e_m}, we use a decrease in the EP learning objective as an additional convergence criterion. As a gold-standard baseline, we include MCMC results. For MCMC, we use a log-uniform hyperpriors to ensure a close match to the model setup in the other models.
 
To reduce the variance introduced by the training--test set split, we repeat the 5-fold CV with ten different seeds. The performance on the test set is given in \cref{table:classification_full_lpd_10seed} and \cref{fig:relacc}. As shown in \cref{table:classification_full_lpd_10seed} for log predictive density, LA and EP have very similar performance to VI-based methods (VI and Ours) on most data sets. This empirically demonstrates that for binary classification on small and mid-sized data sets EP and LA generalize well. Our hybrid training procedure achieves the same test performance as VI on most data sets and outperforms VI on three data sets. It empirically demonstrates that when no sparse approximation is required, by using an improved estimation of the marginal likelihood for hyperparameter learning, we could potentially have better generalization ability at no additional computational cost. As the gold standard, MCMC gives the best results; notably, the gap between MCMC and approximate inference methods is relatively small on small data sets. In practice we often favour methods with stable performance over different data sets, \ie, they might not always give the best performance but we can expect consistently good performances. To investigate the reliability of different methods, in \cref{fig:relacc} we plot the relative accuracy (on each data set we divide the results of each method by the highest accuracy on that data set) for individual data sets and the mean relative accuracy of each method across all data sets. Our approach achieves the most consistent performance on all data sets and thus yields reliable training. We include additional result tables with the same conclusion in \cref{app:classification_additional_tables}, including experiments with an ARD kernel and checks for initializing other methods with our optimal hyperparameters.

\begin{table}[!t]
\centering\scriptsize
\caption{\textbf{Robust regression tasks:} log predictive density (higher is better) with a Student-$t$ likelihood on different data sets (mean $\pm$ standard deviation over 5-fold cross-validation). The best results and those not statistically significantly different from them under a paired $t$-test are \textbf{bolded}. Our objective performs well overall.}
\setlength{\tabcolsep}{2pt}
\input{table/t_test_lpd.tex}
\label{table:t_full_lpd}
\end{table}

\subsection{Robust (Student-$t$) Regression}
\label{sec:robust}
We further test our hybrid training procedure on a more challenging robust regression task with a Student-$t$ likelihood, a model which is not log-concave. In the likelihood, we fix the degrees of freedom, $\nu=3$, and only train the noise scale together with hyperparameters. For LA and EP we follow the methods designed by \citet{aki}. For VI and our EP-like VI, to make the training procedure numerically stable we crop the gradient w.r.t.\ the second element of the natural parameters to prevent the approximate posterior covariance from becoming negative. We test on three benchmark data sets previously used for robust regression: the simulated data from \citet{odata}, the Boston housing regression task, and the stock data from \citet{solin2015state}. \citet{aki} point out that in Student-$t$ regression EP provides a good approximation for marginal likelihood and, as shown in \cref{table:t_full_lpd}, by using an EP-like marginal likelihood estimation for hyperparameters learning our hybrid training procedure generalizes better than vanilla VI. The MCMC gold standard results for {\sc neal}, {\sc boston} and {\sc stock} are $0.309 {\pm} 0.454$, $-0.191 {\pm} 0.051$, and $1.586 {\pm} 0.034$, respectively.

\subsection{Evaluation under Sparse Approximation}
\label{sec:reduce_Z}
It is not obvious how a sparse approximation would affect the quality of marginal likelihood approximation. To be able to compare with the MCMC baseline on the full data set, we first analyse the influence of sparsification on {\sc ionosphere}. %
We choose 75\%, 50\%, 25\%, and 10\% random subsets of training data as inducing points.
We estimate the log marginal likelihood as in \cref{sec:lml_quality} on a grid of values for the log hyperparameters. \cref{fig:reduce_Z} shows the resulting contour surfaces. Unsurprisingly, as we reduce the number of inducing points, the estimation of the log marginal likelihood becomes less accurate. For moderate sparsification ($75\%$, $50\%$), our EP-like marginal likelihood estimation matches the full MCMC baseline better than VI.
For more extreme sparsification, both approximations show significant biases. This is because, with very few inducing points, only larger lengthscales make sense. To match the low-complexity approximation, large lengthscales are required and the ground truth marginal likelihood provided by MCMC becomes irrelevant.

\paragraph{Evaluation on large-scale binary classification}
We compare LA, EP, VI, and our proposed training procedure on five data sets from 2k to 19k data points (for details, see \cref{table:sparse_dataset}). We use k-means to select 500 inducing points from the input data and keep them fixed. The test set performance is given in \cref{table:classification_sparse_lpd}. 
Here we are in the regime of $m/n$ in the range of $25\%$ to $2.5\%$, where the log marginal likelihood surface approximations of both VI and our EP-like approximation are likely to be biased away from the true marginal likelihood surface, and the approximate sparse model can no longer be considered a surrogate of the true model.\looseness-1

\begin{table}[!t]
\centering\scriptsize
\caption{\textbf{Sparse approximation} for classification on different data sets: log predictive density (mean $\pm$ standard deviation). Higher is better. Results that are statistically significantly different under a paired $t$-test are \textbf{bolded}.}
\setlength{\tabcolsep}{3pt}
\input{table/isotropic_sparse_test_lpd.tex}
\label{table:classification_sparse_lpd}
\end{table}

\subsection{Practical Considerations}
\label{sec:conflict_e_m}
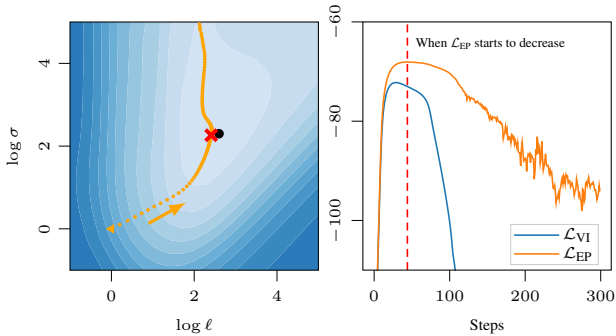
\begin{figure}
  \tikzexternaldisable
  \setlength{\figurewidth}{.4\columnwidth}
  \setlength{\figureheight}{1\figurewidth}
  \pgfplotsset{scale only axis,axis on top,y tick label style={rotate=90,font=\tiny}, legend style={font=\tiny, at={(.95,0.02)},anchor=south east},x tick label style={font=\tiny}, xlabel near ticks, ylabel near ticks}
   \begin{subfigure}[t]{.48\columnwidth}
  \pgfplotsset{x label style={font=\tiny},y label style={font=\tiny}, xlabel={$\log \ell$}, ylabel={$\log \sigma$}}
    \input{./fig/tikz/fig4b.tex}
    \definecolor{orange}{RGB}{255,165,0}
    \tikz[remember picture]\coordinate (coord) at (0,0);
    \tikz[overlay,remember picture]\draw[yshift=1cm,->,orange,very thick,-latex] ($(coord) + (2,2)$) -- ++(0.5,0.28);
  \end{subfigure} 
  \hfill
  \begin{subfigure}[t]{.48\columnwidth}
  \pgfplotsset{x label style={font=\tiny}, xlabel={Steps}}
    \input{./fig/tikz/fig4a.tex}
  \end{subfigure}\\[-2em]
  \caption{Interplay between the two optimization targets can result in overshooting the optimum (\ref{dot}). The optimizer starts at \ref{triangle} and once it passes \ref{cross} (dashed red; stopping point), $\mathcal{L}_{\text{EP}}$ starts to decrease. After that, the optimization becomes increasingly unstable.}
  \label{fig:opt}
   \tikzexternalenable
\end{figure}

VEM with the same objective for E- and M-step is analogous to coordinate ascent and hence guaranteed to always improve the objective. With $\mathcal{L}_\text{VI}$ for E-step and $\mathcal{L}_\text{EP}$ for M-step, this guarantee no longer holds. This can introduce interplay between the two targets, and in our experiments, we observed that (with certain data-splits/model setups) the optimization can overshoot past the optimum and then becomes increasingly unstable, see \cref{fig:opt}. In our experiments this conflict between $\mathcal{L}_\text{E}=\mathcal{L}_\text{VI}$ and $\mathcal{L}_\text{M}=\mathcal{L}_\text{EP}$ occurred in about 54\% of cases.
We address this issue by ending optimization once the $\mathcal{L}_\text{EP}$ objective starts to decrease (see \cref{alg:training_procedure}). Note that this is solely based on the training data and does not require any additional validation data or tuning per data set.

\section{Discussion and Conclusions}
In GP models, the training separates into inference and learning which are typically both cast into an optimization problem. 
In this paper, we improved hyperparameter learning in non-conjugate GP models by augmenting VI with an EP-like learning target. Our hybrid training procedure builds upon the dual variational GP formulation, introduced by \citet{cvi} and extended to sparse GPs in \citet{dual}, which provides a link between VI and EP. We used the representation of the posterior from VI to obtain an EP-like approximation of the marginal likelihood for hyperparameter optimization---without any added computational complexity or computation time. 

In the experiments, we evaluated our hybrid training procedure on binary classification tasks and robust regression. For full (non-sparse) models, the extensive results (over 1350 runs per method) show clear benefits in stability, reliability, and performance for our method. This shows the benefits of decoupling inference and learning.
When more hyperparameters are present, as shown in \cref{table:classification_full_lpd_ard} and \cref{table:classification_full_acc_ard} in the appendix, our method has the same performance on log predictive density as VI. However, as shown in \cref{fig:relacc_lpd}, our method is still more reliable than VI.
For sparse problems, similar empirical benefits could not be demonstrated. 

We provide a reference implementation of the methods and code to reproduce the experiments at \url{https://github.com/AaltoML/improved-hyperparameter-learning}.

\section*{Acknowledgements}
This work was supported by funding from the Academy of Finland (grant id 339730) and the Finnish Center for Artificial Intelligence (FCAI). We acknowledge the computational resources provided by the Aalto Science-IT project and CSC -- IT Center for Science, Finland. We thank Aidan Scannell for his constructive comments on the manuscript.

\bibliographystyle{icml2023}

\clearpage

\appendix

\onecolumn

\section*{Appendix}

In the supplementary material, we include technical details of the methods that were omitted for brevity in the main paper (\cref{app:app_details}). Additionally, we provide details on the experiments and evaluation setup (\cref{app:computation}--\ref{app:sparse_app}) for reproducing the results in the main paper, and include further result tables and figures that extend the evaluation. The codes for the methods proposed in this paper are included as a separate supplement.

\section{Method Details}
\label{app:app_details}
\paragraph{Sparse energy}
To extend the presentation in \cref{sec:sparse}, we derive how to obtain the sparse EP marginal likelihood estimation from the VI approximate posterior.
Following Eq.~(77) in \citet{powerep}, the sites $t_i(\vu;\vzeta_{\vu,i})$ in \cref{eq:sparse_ep_energy} are
\begin{equation}
	t_i(\vu;\vzeta_{\vu,i}) \propto \exp \langle\vu^{\top} \va_i \vzeta_{\vu, i, 2} \vzeta_{\vu, i, 1}-\frac{1}{2} \vu^{\top} \va_i \vzeta_{\vu, i, 2} \va_i^{\top} \vu \rangle,
\end{equation}
where $\va_i=\MKuu^{-1} \vk_{\vu,i}$, $\MKuu = \kappa(\MZ, \MZ)$, and $\vk_{\vu,i}=\kappa(\MZ, \vx_i)$ for the $i$\textsuperscript{th} data sample.
Note that $\vu^{\top} \va_i = \va_i^{\top} \vu$ is a scalar.

As $t_i(\vu; \vlambda_{\vu,i})$ in \cref{eq:sparse_vi_post} is given by
\begin{align}
	t_i(\vu; \vlambda_{\vu,i}) &= \exp{\langle\vlambda_{\vu,i}, \MT(\va_i^{\top} \vu)\rangle} \notag\\
	&=\exp \langle \vlambda_{\vu,i,1}\va_i^{\top} \vu + \vlambda_{\vu,i,2} (\va_i^{\top} \vu)^2 \rangle,
\end{align}
we have the following correspondence between $\vzetau$ and $\vlambdau$:
\begin{equation}\label{eq:epviparameter}
	\vzeta_{\vu, i, 2} \vzeta_{\vu, i, 1}  \Leftrightarrow   \vlambda_{\vu, i, 1} \qquad \text{and} \qquad -\frac12\vzeta_{\vu, i, 2}  \Leftrightarrow   \vlambda_{\vu, i, 2}.
\end{equation}
Following Eq.~(126) in \citet{powerep}, the sparse EP energy is (we omit $\vtheta$ to make notation simpler)
\begin{align}
 \lmlEPs (\vzetau, \vtheta) &= \frac{1}{2} \log |\MS_{\vu}|+\frac{1}{2} \vm_{\vu}^{\top} \MS_{\vu}^{-1} \vm_{\vu}-\frac{1}{2} \log |\mathbf{K}_{\mathbf{u u}}|+\frac{1}{\alpha} \sum_n \log \mathcal{Z}_{\mathrm{tilted}, i} \nonumber\\
&\quad+\sum_n\big[-\frac{1}{2 \alpha} \log (1-\mathbf{a}_i^{\top} \alpha \vzeta_{\vu, i, 2} \MS_{\vu} \mathbf{a}_i)+\frac{1}{2} \frac{\vm_{\vu}^{\top} \mathbf{a}_i \vzeta_{\vu, i, 2} \mathbf{a}_i^{\top} \vm_{\vu}}{1-\mathbf{a}_i^{\top} \alpha \vzeta_{\vu, i, 2} \MS_{\vu} \mathbf{a}_i} \nonumber\\
&\quad+\frac{1}{2} \vzeta_{\vu, i, 1} \vzeta_{\vu, i, 2} \mathbf{a}_i^{\top} \mathbf{V}_{\mathrm{cav}, i} \mathbf{a}_i \alpha \vzeta_{\vu, i, 2} \vzeta_{\vu, i, 1}-\vzeta_{\vu, i, 1} \vzeta_{\vu, i, 2} \mathbf{a}_i^{\top} \mathbf{V}_{\mathrm{cav}, i}\MS_{\vu}^{-1} \vm_{\vu}\big], \label{eq:written_out_sparse_ep_energy}
\end{align}
where the different terms are defined by
\begin{align}
\MS_{\mathrm{cav}, i} &= \MS_{\vu}+\frac{\MS_{\vu} \va_i \alpha \vzeta_{\vu,i,2} \va_i^{\top} \MS_{\vu}}{1-\va_i^{\top} \, \alpha \, \vzeta_{\vu,i,2} \MS_{\vu} \va_i}, \\
\MS_{\mathrm{cav}, i}^{-1} \vm_{\mathrm{cav}, i}&=\MS_{\vu}^{-1}  \vm_{\vu}-\alpha \, \va_i \vzeta_{\vu,i,2} \vzeta_{\vu,i,1}, \\
\log \mathcal{Z}_{\mathrm{tilted}, i} &= \log \int q_{\text{cav},i}(f_i) \, p^\alpha\left(y_i \mid f_i\right) \diff f_i, \\
q_{\text{cav},i}(f_i) &= \int p\left(f_i \mid \vu\right) \mathrm{N}(\vm_{\mathrm{cav}, i}, \MS_{\mathrm{cav}, i}) \diff \vu.
\end{align}
By substituting $\vlambda_{\vu,i}$ into $\vzeta_{\vu,i}$ in \cref{eq:written_out_sparse_ep_energy} using \cref{eq:epviparameter}, we obtain the sparse EP marginal likelihood approximation with the VI approximate posterior. When $\alpha=1$, Power-EP reduces to normal EP, which we use in our experiments.

\section{Computational Details}
\label{app:computation}
All experiments ran on a cluster, which allowed us to parallelize jobs. This played a central role especially for the MCMC baseline results for the marginal likelihood surfaces, where we split into 441 separate jobs (per hyperparameter value combination), each of which were allocated 1--3 CPU cores and 1~Gb memory and ran 8--40~h depending on data set size.\looseness-1

\section{Quality of Marginal Likelihood Approximation}
Inspired by the work by \citet{05classification} and \citet{08classification}, we compare the quality of marginal likelihood approximations of LA, VI, EP, and our EP-like VI with a `ground' truth obtained by annealed importance sampling \citep[AIS,][]{AIS}. We demonstrate this on binary classification tasks, where we estimate the log marginal likelihood on a $21 \times 21$ grid of values for the log hyperparameters $\log \vtheta=(\log \ell, \log \sigma)$ and plot the contour on the grid. For each hyperparameter setting, we fix the hyperparameters and evaluate the approximate log marginal likelihood based on the inferred approximate posterior. We then also visualize the log predictive density on hold-out test data as a similar contour plot, showing what the performance of the model would have been if the hyperparameters would have been chosen based on the log marginal likelihood surface in question under the specific inference scheme.

\subsection{Markov Chain Monte Carlo Baseline}
\label{app:mcmc}
As in previous work, we use an MCMC approach as the gold-standard baseline. We use the annealed importance sampling \citep{AIS} approach from \citet{05classification} and \citet{08classification} that defines a sequence of $t=0,1,\ldots,T$ steps
	$Z_t= \int \likelihoodfull^{\tau(t)} \prior \diff \vf$,
where $\tau(t)=(t/T)^4$ (such that $\tau(0)=0$ and $\tau(T)=1$). The marginal likelihood can be rewritten as
\begin{equation}
	p(\vy;\vtheta)=\frac{Z_T}{Z_0}=\frac{Z_T}{Z_{T-1}} \frac{Z_{T-1}}{Z_{T-2}} \cdots \frac{Z_1}{Z_0},
\end{equation}
where $\nicefrac{Z_t}{Z_{t-1}}$ is approximated by importance sampling using samples from $q_t(\vf ) \propto p(\vy \mid \vf ; \vtheta)^{\tau(t-1)} \, \prior$:
\begin{align}
\frac{Z_t}{Z_{t-1}} &= \frac{\int p(\mathbf{y} \mid \vf ; \vtheta)^{\tau(t)} \prior \diff \vf }{Z_{t-1}} \nonumber \\
&= \int \frac{p(\mathbf{y} \mid \vf ; \vtheta)^{\tau(t)}}{p(\mathbf{y} \mid \vf ; \vtheta)^{\tau(t-1)}} \frac{p(\mathbf{y} \mid \vf ; \vtheta)^{\tau(t-1)}\prior }{Z_{t-1}} \diff \vf \nonumber \\
& \approx \frac{1}{S} \sum_{s=1}^S p(\mathbf{y} \mid \vf_t^{(s)}; \vtheta)^{\tau(t)-\tau(t-1)}, \quad \text{where} \quad
 \vf_t^{(s)} \sim \frac{p(\mathbf{y} \mid \vf ; \vtheta)^{\tau(t-1)}\prior }{Z_{t-1}}.
\end{align}
In practice, instead of sampling $\vf_t$ from $\frac{p(\mathbf{y} \mid \vf)^{\tau(t-1)}\mathrm{N}(\vf \mid \vm, \mathbf{K})}{Z_{t-1}}$ directly, we use a parameterisation in terms of $\boldsymbol{\alpha}=\mathbf{K}^{-1}(\vf_t-\vm)$ and sample from $\boldsymbol{\alpha} \sim P(\boldsymbol{\alpha})=\frac{p(\mathbf{y} \mid \mathbf{K}\boldsymbol{\alpha} + \vm)^{\tau(t-1)}\mathrm{N}(\boldsymbol{\alpha} \mid \boldsymbol{0}, \mathbf{K}^{-1})}{Z_a}$ to increase numerical stability since $\log P(\boldsymbol{\alpha})$ and its gradient can be computed safely. We use elliptical slice sampling \cite{murray2010elliptical}. %
Now, by using a single sample $S=1$ and a large number of steps $T$, the estimation of log marginal likelihood can be written as 
\begin{equation}
	\log p(\vy;\vtheta) = \sum_{t=1}^{T} \log \frac{Z_t}{Z_{t-1}} \approx \sum_{t=1}^{T} (\tau(t)-\tau(t-1))\log p(\mathbf{y} \mid \vf_t; \vtheta).
\end{equation}
Following \citet{05classification}, we set $T=8000$ and combine three estimates of log marginal likelihood by their geometric mean.

\subsection{Experiment Results}
\label{app:additional_contour}
\begin{figure*}[t!]
\centering\footnotesize
\setlength{\figurewidth}{0.17\textwidth}
\setlength{\figureheight}{1.\figurewidth}
\pgfplotsset{grid style={dashed, white},scale only axis,xlabel near ticks, tick align=inside, ylabel near ticks, axis on top, ticklabel style = {font=\tiny, inner sep=3pt}, ytick={-1,1,3,5}, xtick={-1,1,3,5},ylabel style={yshift=-1em, align=center}, grid=both, minor tick num=1}
\tikzexternalenable  
\tikzsetnextfilename{main-figure2} 
\begin{tikzpicture}[inner sep=0, outer sep=0, remember picture]

  \def\data{./fig/tikz/sonar}
  \def\myxshift{0.6cm}
  \def\myxscaler{1.1}
  \def\myyscaler{-1.1}

  \foreach \x/\name [count=\i from 0] in {MCMC_mean_lml/MCMC,LA_mean_lml/LA, EP_mean_lml/EP,CVI_mean_lml/VI,CVI_mean_lml_ep/Ours} {
    \pgfplotsset{title=\name,ylabel={},xticklabels={}}
    \ifthenelse{\i > 0}{\pgfplotsset{yticklabels={}}}{}
    \begin{scope}[shift={(\myxscaler*\i*\figurewidth+\myxshift,0)}]
      \input{\data/\x.tex}
    \end{scope}
  }

  \node[rotate=90,align=center] at (-.05\figureheight,.5\figureheight) {\textbf{Marginal likelihood}\\[.6ex]$\log \sigma$};
  \node[rotate=90,align=center] at (-.05\figureheight,-.6\figureheight) {\textbf{Predictive density}\\[.6ex]$\log \sigma$};

  \foreach \x [count=\i from 0] in {MCMC_nlpd,LA_nlpd,EP_nlpd} {
    \pgfplotsset{title={},xlabel={$\log \ell$}}
    \ifthenelse{\i > 0}{\pgfplotsset{yticklabels={}}}{}
    \begin{scope}[shift={(\myxscaler*\i*\figurewidth+\myxshift,\myyscaler*\figureheight)}]
      \input{\data/\x.tex}
    \end{scope}
  }  

  \pgfplotsset{title={},xlabel={$\log \ell$}}
  \begin{scope}[shift={(\myxscaler*3.5*\figurewidth+\myxshift,\myyscaler*\figureheight)}]
    \input{\data/CVI_nlpd.tex}
  \end{scope}

  \draw[blue, shorten >=.2cm,shorten <=.2cm,-Stealth] (CVI_mean_lml_ep_t) to[out=250, in = 50] (CVI_mean_lml_ep_b);
  \draw[blue, shorten >=.2cm,shorten <=.2cm,-Stealth] (CVI_mean_lml_t) to[out=250, in = 200] (CVI_mean_lml_b);
\end{tikzpicture}
\caption{\textbf{Log marginal likelihood / predictive density surfaces} for the {\sc sonar} data set by varying kernel magnitude $\sigma$ and lengthscale $\ell$. The colour scale is the same in all plots: $-1.0$~\includegraphics[width=1cm,height=.65em]{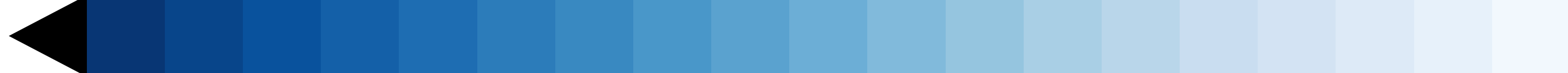}~$-0.2$ (normalized by $n$). Optimal hyperparameters of each method shown by a black marker. EP and our EP-like marginal likelihood estimation match the MCMC baseline better than VI or LA, thus providing a learning proxy. For prediction, our method still leverages the same variational representation as VI.}
\label{fig:sonar_contour}
\tikzexternaldisable
\end{figure*}

\begin{figure*}[t!]
\centering\footnotesize
\setlength{\figurewidth}{0.17\textwidth}
\setlength{\figureheight}{1.\figurewidth}
\pgfplotsset{grid style={dashed, white},scale only axis,xlabel near ticks, tick align=inside, ylabel near ticks, axis on top, ticklabel style = {font=\tiny, inner sep=3pt}, ytick={-1,1,3,5}, xtick={-1,1,3,5},ylabel style={yshift=-1em, align=center}, grid=both, minor tick num=1}
\tikzexternalenable  
\tikzsetnextfilename{main-figure3} 
\begin{tikzpicture}[inner sep=0, outer sep=0, remember picture]

  \def\data{./fig/tikz/usps}
  \def\myxshift{0.6cm}
  \def\myxscaler{1.1}
  \def\myyscaler{-1.1}

  \foreach \x/\name [count=\i from 0] in {MCMC_mean_lml/MCMC,LA_mean_lml/LA, EP_mean_lml/EP,CVI_mean_lml/VI,CVI_mean_lml_ep/Ours} {
    \pgfplotsset{title=\name,ylabel={},xticklabels={}}
    \ifthenelse{\i > 0}{\pgfplotsset{yticklabels={}}}{}
    \begin{scope}[shift={(\myxscaler*\i*\figurewidth+\myxshift,0)}]
      \input{\data/\x.tex}
    \end{scope}
  }

  \node[rotate=90,align=center] at (-.05\figureheight,.5\figureheight) {\textbf{Marginal likelihood}\\[.6ex]$\log \sigma$};
  \node[rotate=90,align=center] at (-.05\figureheight,-.6\figureheight) {\textbf{Predictive density}\\[.6ex]$\log \sigma$};

  \foreach \x [count=\i from 0] in {MCMC_nlpd,LA_nlpd,EP_nlpd} {
    \pgfplotsset{title={},xlabel={$\log \ell$}}
    \ifthenelse{\i > 0}{\pgfplotsset{yticklabels={}}}{}
    \begin{scope}[shift={(\myxscaler*\i*\figurewidth+\myxshift,\myyscaler*\figureheight)}]
      \input{\data/\x.tex}
    \end{scope}
  }  

  \pgfplotsset{title={},xlabel={$\log \ell$}}
  \begin{scope}[shift={(\myxscaler*3.5*\figurewidth+\myxshift,\myyscaler*\figureheight)}]
    \input{\data/CVI_nlpd.tex}
  \end{scope}

  \draw[blue, shorten >=.2cm,shorten <=.2cm,-Stealth] (CVI_mean_lml_ep_t) to[out=250, in = 50] (CVI_mean_lml_ep_b);
  \draw[blue, shorten >=.2cm,shorten <=.2cm,-Stealth] (CVI_mean_lml_t) to[out=250, in = 200] (CVI_mean_lml_b);
\end{tikzpicture}
\caption{\textbf{Log marginal likelihood / predictive density surfaces} for the {\sc usps} data set (MNIST-like digits image) by varying kernel magnitude $\sigma$ and lengthscale $\ell$. The colour scale is the same in all plots: $-1.0$~\includegraphics[width=1cm,height=.65em]{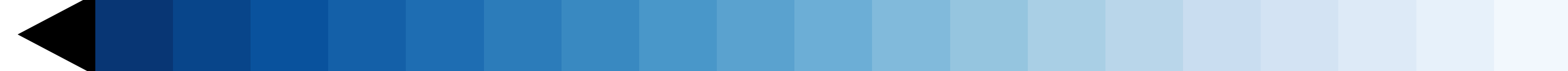}~$0$ (normalized by $n$). Optimal hyperparameters shown by a black marker. EP and our EP-like marginal likelihood estimation match the MCMC baseline better than VI or LA, thus providing a learning proxy. For prediction, our method still leverages the same variational representation as VI.}
\label{fig:usps_contour}
\tikzexternaldisable
\end{figure*}

\begin{figure*}[t!]
\centering\footnotesize
\setlength{\figurewidth}{0.17\textwidth}
\setlength{\figureheight}{1.\figurewidth}
\pgfplotsset{grid style={dashed, white},scale only axis,xlabel near ticks, tick align=inside, ylabel near ticks, axis on top, ticklabel style = {font=\tiny, inner sep=3pt}, ytick={-1,1,3,5}, xtick={-1,1,3,5},ylabel style={yshift=-1em, align=center}, grid=both, minor tick num=1}
\tikzexternalenable  
\tikzsetnextfilename{main-figure4} 
\begin{tikzpicture}[inner sep=0, outer sep=0, remember picture]

  \def\data{./fig/tikz/parkinsons}
  \def\myxshift{0.6cm}
  \def\myxscaler{1.1}
  \def\myyscaler{-1.1}

  \foreach \x/\name [count=\i from 0] in {MCMC_mean_lml/MCMC,LA_mean_lml/LA, EP_mean_lml/EP,CVI_mean_lml/VI,CVI_mean_lml_ep/Ours} {
    \pgfplotsset{title=\name,ylabel={},xticklabels={}}
    \ifthenelse{\i > 0}{\pgfplotsset{yticklabels={}}}{}
    \begin{scope}[shift={(\myxscaler*\i*\figurewidth+\myxshift,0)}]
      \input{\data/\x.tex}
    \end{scope}
  }

  \node[rotate=90,align=center] at (-.05\figureheight,.5\figureheight) {\textbf{Marginal likelihood}\\[.6ex]$\log \sigma$};
  \node[rotate=90,align=center] at (-.05\figureheight,-.6\figureheight) {\textbf{Predictive density}\\[.6ex]$\log \sigma$};

  \foreach \x [count=\i from 0] in {MCMC_nlpd,LA_nlpd,EP_nlpd} {
    \pgfplotsset{title={},xlabel={$\log \ell$}}
    \ifthenelse{\i > 0}{\pgfplotsset{yticklabels={}}}{}
    \begin{scope}[shift={(\myxscaler*\i*\figurewidth+\myxshift,\myyscaler*\figureheight)}]
      \input{\data/\x.tex}
    \end{scope}
  }  

  \pgfplotsset{title={},xlabel={$\log \ell$}}
  \begin{scope}[shift={(\myxscaler*3.5*\figurewidth+\myxshift,\myyscaler*\figureheight)}]
    \input{\data/CVI_nlpd.tex}
  \end{scope}

  \draw[blue, shorten >=.2cm,shorten <=.2cm,-Stealth] (CVI_mean_lml_ep_t) to[out=250, in = 50] (CVI_mean_lml_ep_b);
  \draw[blue, shorten >=.2cm,shorten <=.2cm,-Stealth] (CVI_mean_lml_t) to[out=250, in = 200] (CVI_mean_lml_b);
\end{tikzpicture}
\caption{\textbf{Log marginal likelihood / predictive density surfaces} for the {\sc parkinsons} data set by varying kernel magnitude $\sigma$ and lengthscale $\ell$. The colour scale is the same in all plots: $-0.8$~\includegraphics[width=1cm,height=.65em]{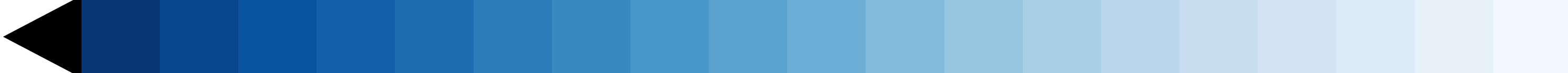}~$0$ (normalized by $n$). Optimal hyperparameters shown by a black marker. EP and our EP-like marginal likelihood estimation match the MCMC baseline better than VI or LA, thus providing a learning proxy. For prediction, our method still leverages the same variational representation as VI.}
\label{fig:parkinsons_contour}
\tikzexternaldisable
\end{figure*}

\begin{figure*}[t!]
\centering\footnotesize
\setlength{\figurewidth}{0.17\textwidth}
\setlength{\figureheight}{1.\figurewidth}
\pgfplotsset{grid style={dashed, white},scale only axis,xlabel near ticks, tick align=inside, ylabel near ticks, axis on top, ticklabel style = {font=\tiny, inner sep=3pt}, ytick={-1,1,3,5}, xtick={-1,1,3,5},ylabel style={yshift=-1em, align=center}, grid=both, minor tick num=1}
\tikzexternalenable  
\tikzsetnextfilename{main-figure5} 
\begin{tikzpicture}[inner sep=0, outer sep=0, remember picture]

  \def\data{./fig/tikz/monks-2}
  \def\myxshift{0.6cm}
  \def\myxscaler{1.1}
  \def\myyscaler{-1.1}

  \foreach \x/\name [count=\i from 0] in {MCMC_mean_lml/MCMC,LA_mean_lml/LA, EP_mean_lml/EP,CVI_mean_lml/VI,CVI_mean_lml_ep/Ours} {
    \pgfplotsset{title=\name,ylabel={},xticklabels={}}
    \ifthenelse{\i > 0}{\pgfplotsset{yticklabels={}}}{}
    \begin{scope}[shift={(\myxscaler*\i*\figurewidth+\myxshift,0)}]
      \input{\data/\x.tex}
    \end{scope}
  }

  \node[rotate=90,align=center] at (-.05\figureheight,.5\figureheight) {\textbf{Marginal likelihood}\\[.6ex]$\log \sigma$};
  \node[rotate=90,align=center] at (-.05\figureheight,-.6\figureheight) {\textbf{Predictive density}\\[.6ex]$\log \sigma$};

  \foreach \x [count=\i from 0] in {MCMC_nlpd,LA_nlpd,EP_nlpd} {
    \pgfplotsset{title={},xlabel={$\log \ell$}}
    \ifthenelse{\i > 0}{\pgfplotsset{yticklabels={}}}{}
    \begin{scope}[shift={(\myxscaler*\i*\figurewidth+\myxshift,\myyscaler*\figureheight)}]
      \input{\data/\x.tex}
    \end{scope}
  }  

  \pgfplotsset{title={},xlabel={$\log \ell$}}
  \begin{scope}[shift={(\myxscaler*3.5*\figurewidth+\myxshift,\myyscaler*\figureheight)}]
    \input{\data/CVI_nlpd.tex}
  \end{scope}

  \draw[blue, shorten >=.2cm,shorten <=.2cm,-Stealth] (CVI_mean_lml_ep_t) to[out=250, in = 50] (CVI_mean_lml_ep_b);
  \draw[blue, shorten >=.2cm,shorten <=.2cm,-Stealth] (CVI_mean_lml_t) to[out=250, in = 200] (CVI_mean_lml_b);
\end{tikzpicture}
\caption{\textbf{Log marginal likelihood / predictive density surfaces} for the {\sc monks-2} data set by varying kernel magnitude $\sigma$ and lengthscale $\ell$. The colour scale is the same in all plots: $-0.9$~\includegraphics[width=1cm,height=.65em]{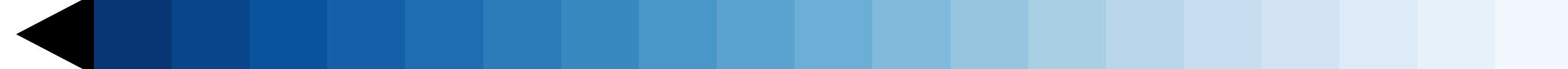}~$-0.3$ (normalized by $n$). Optimal hyperparameters shown by a black marker. EP and our EP-like marginal likelihood estimation match the MCMC baseline better than VI or LA, thus providing a learning proxy. For prediction, our method still leverages the same variational representation as VI.}
\label{fig:monks_contour}
\tikzexternaldisable
\end{figure*}

In addition to the figure for the {\sc ionosphere} data set (\cref{fig:ionosphere_contour}) in the main paper, we include surface plots for four additional data sets in the appendix for more comprehensive comparisons. The marginal likelihood estimation on {\sc sonar}, {\sc usps}, {\sc parkinsons} and {\sc monks-2} are given in \cref{fig:sonar_contour}, \cref{fig:usps_contour}, \cref{fig:parkinsons_contour}, and \cref{fig:monks_contour}, respectively. The {\sc ionosphere} and {\sc sonar} were included to make it easy to compare to previous work, and the other three chosen as an interesting subset covering different type of classification tasks (from general classification to small images).

Similarly to the results on {\sc ionosphere} in the main paper, when using the EP-like marginal likelihood estimation from the VI approximate posterior (Ours), the contour shapes become closer to the MCMC result, and the optimal hyperparameter location of EP-like VI (Ours) is closer to MCMC than VI. These effects help explain the quantitative results in the tables in the main paper and the supplement.


\clearpage

\section{Non-Conjugate Tasks in Bayesian Benchmarks}
We use the {\em Bayesian Benchmarks} suite (\rurl{github.com/secondmind-labs/bayesian_benchmarks}; originally by Salimbeni~\etal) for evaluating the methods in binary classification. Bayesian benchmarks includes common evaluation data sets (typically from UCI) and makes it possible to run a large number of comparisons under a fixed evaluation setup. In the first part, we only include binary classification tasks (Bernoulli likelihood) with $n \leq 1000$. We follow the standard setup of input point normalization and splits in the evaluation suite.

\subsection{Evaluation Metrics}
We conduct 5-fold cross-validation and use test set accuracy and log predictive density to evaluate the test performance of each method (higher is better in both). To compare different methods, we use the paired $t$-test (with $p=0.05$) that compares whether the best-performing method performs statistically significantly better than the others.

\subsection{Experiment Setup}
We initialize the hyperparameters with unit lengthscale and magnitude for all methods. For LA and EP, the hyperparameters are optimized by the default optimizer L-BFGS-B in GPy. For VI and our hybrid training procedure, each E step and M step consists of 20 iterations. In the E-step we set the learning rate of natural gradient descent to be $0.1$. In the M-step we use the Adam optimizer \citep{adam} with learning rate $0.01$. We use the convergence criterion described in the main text, with a maximum number of at most $10\,000$ steps. 

For MCMC, we use log uniform prior to ensure it is the same model as the approximate inference methods. We set the burn-in step to be 200, the number of samples to be 10000, and thin the sample with 2.

\subsection{Experiment Results}
\label{app:classification_additional_tables}
The test set accuracies are given in \cref{table:classification_full_acc_seed}. For test accuracies all methods achieve similar performance, which is to be expected as accuracy alone only captures where the decision boundary has been draw, completely disregarding second-order information. The log predictive density results are included in the main paper (\cref{table:classification_full_lpd_10seed}). Our hybrid training procedure gives the most consistent results and thus achieves the most reliable training.

\begin{table*}[h!]
\centering\scriptsize
\caption{\textbf{Binary classification:} test set accuracy (higher is better) on different data sets from the {\em Bayesian benchmarks} over 5-fold cross-validation with 10 different seeds. Best results and those not statistically significantly different from them under a paired $t$-test are \textbf{bolded}. We provide MCMC results for reference (excluded from bolding). All inference methods perform well overall, while our training objective delivers the most reliable performance.}
\setlength{\tabcolsep}{10pt}
\input{table/seed_test_acc.tex}
\label{table:classification_full_acc_seed}
\end{table*}

\subsection{Ablation Studies}
\label{app:classification_ablation}
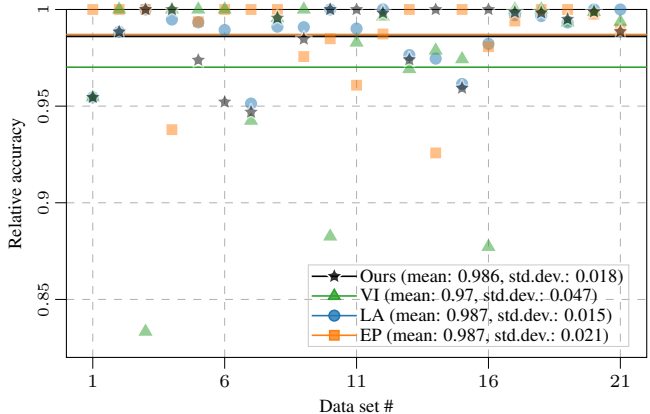
\begin{figure}[h!]
  \tikzexternaldisable
  \centering\scriptsize
  \setlength{\figurewidth}{.45\columnwidth}
  \setlength{\figureheight}{.6\figurewidth}
  \pgfplotsset{scale only axis,ylabel near ticks, y tick label style={rotate=90,font=\tiny},y tick label style={font=\tiny},xtick={1,6,11,16,21,26},grid style={line width=.1pt, draw=gray!10,dashed},grid,legend style={fill=white}}
  \input{./fig/tikz/classification_relative_lpd_acc.tex}\\[-1.5em]
  \caption{Mean relative accuracy compared to best method on each data set of \cref{table:classification_full_acc_ard}. The horizontal lines indicate mean across all data sets; see legend for mean and standard deviation. Our approach yields reliable training.}
  \vspace*{-1em}
  \label{fig:relacc_lpd}
  \tikzexternalenable
\end{figure}

\paragraph{Automatic Relevance Determination Kernel} We run experiments with automatic relevance determination (ARD) kernel to see whether our method would perform well when there are multiple hyperparameters.
To ensure fair comparison we only include results on data sets where all methods have converged.
The log predictive density and test set accuracy are given in \cref{table:classification_full_lpd_ard} and \cref{table:classification_full_acc_ard} respectively. The mean relative accuracy is plotted in \cref{fig:relacc_lpd}. Our training objective performs well overall and achieves reliable training.
 
\begin{table*}[h!]
\centering\scriptsize
\caption{\textbf{Binary classification:} log predictive density (higher is better) on different data sets with ARD kernel from the {\em Bayesian benchmarks} over 5-fold cross-validation. To ensure fair comparison we only include results on data sets where all methods have converged. Best results and those not statistically significantly different from them under a paired $t$-test are \textbf{bolded}.}
\setlength{\tabcolsep}{10pt}
\input{table/ard_test_lpd.tex}
\label{table:classification_full_lpd_ard}
\end{table*}

\begin{table*}[h!]
\centering\scriptsize
\caption{\textbf{Binary classification:} test set accuracy (higher is better) on different data sets with ARD kernel from the {\em Bayesian benchmarks} over 5-fold cross-validation. To ensure fair comparison we only include results on data sets where all methods have converged. Best results and those not statistically significantly different from them under a paired $t$-test are \textbf{bolded}.}
\setlength{\tabcolsep}{10pt}
\input{table/ard_test_acc.tex}
\label{table:classification_full_acc_ard}
\end{table*}

\paragraph{Convergence of EP} To make sure EP fully converged, we initialized it with the trained hyperparameter values of our method and the log predictive density results and test set accuracies are given in \cref{table:classification_full_lpd_ep} and \cref{table:classification_full_acc_ep}. Compared with \cref{table:classification_full_lpd} and \cref{table:classification_full_acc}, regarding log predictive density EP has three more bolded results and regarding test set accuracies EP has two more bolded results. This underlines some of the issues associated with EP and speaks in favour of our method.

\begin{table*}[h!]
\centering\scriptsize
\caption{\textbf{Binary classification:} log predictive density (higher is better) on different data sets from the {\em Bayesian benchmarks} over 5-fold cross-validation. We initialize EP with the trained hyperparameters values of our method. Best results and those not statistically significantly different from them under a paired $t$-test are \textbf{bolded}. The results are largely the same as \cref{table:classification_full_lpd}, which means EP has converged.}
\setlength{\tabcolsep}{10pt}
\input{table/ep_test_lpd.tex}
\label{table:classification_full_lpd_ep}
\end{table*}

\begin{table*}[h!]
\centering\scriptsize
\caption{\textbf{Binary classification:} test set accuracy (higher is better) on different data sets from the {\em Bayesian benchmarks} over 5-fold cross-validation. We initialize EP with the trained hyperparameters values of our method. Best results and those not statistically significantly different from them under a paired $t$-test are \textbf{bolded}. The results are largely the same as \cref{table:classification_full_acc}, which means EP has converged.}
\setlength{\tabcolsep}{10pt}
\input{table/ep_test_acc.tex}
\label{table:classification_full_acc_ep}
\end{table*}

\begin{table*}[h!]
\centering\scriptsize
\caption{\textbf{Binary classification:} log predictive density (higher is better) on different data sets from the {\em Bayesian benchmarks} (mean $\pm$ standard deviation over 5-fold cross-validation). The best results and those not statistically significantly different from them under a paired $t$-test are \textbf{bolded}. For the classification accuracy all methods perform comparably, which is to be expected as accuracy alone only captures where the decision boundary has been draw, completely disregarding second-order information.}
\setlength{\tabcolsep}{10pt}
\input{table/new_isotropic_test_lpd.tex}
\label{table:classification_full_lpd}
\end{table*} 
 
\begin{table*}[h!]
\centering\scriptsize
\caption{\textbf{Binary classification:} test set accuracy (higher is better) on different data sets from the {\em Bayesian benchmarks} (mean $\pm$ standard deviation over 5-fold cross-validation). The best results and those not statistically significantly different from them under a paired $t$-test are \textbf{bolded}. For the classification accuracy all methods perform comparably, which is to be expected as accuracy alone only captures where the decision boundary has been draw, completely disregarding second-order information.}
\setlength{\tabcolsep}{10pt}
\input{table/new_isotropic_test_acc.tex}
\label{table:classification_full_acc}
\end{table*}

\paragraph{Convergence of VI} To ensure that the comparison to VI is as fair as possible, we ran both the vanilla VI approach where we jointly optimize $\vxi$ and $\vtheta$ using L-BFGS-B and the CVI approach by \citet{cvi} that uses natural gradients for the E-step for faster convergence. The speed of convergence was not the main interest, and thus we verified that the VI results obtained by the two different approaches gave the same results---thus confirming that the VI results presented in the tables are all for converged optimization runs.

\section{Robust (Student-$t$) Regression}
For additional insight, we include results for a Student-$t$ likelihood model that allows heavy-tailed noise in the observations. This likelihood is not log-concave, which makes LA and EP more challenging. Previous works in the field do not list many standard data sets for benchmarking, but we have gathered three previously used benchmark problems for comparison.

\subsection{Experiment Setup}

We preprocessed the data and selected an interval of 1000 data points from the original data. We initialize all methods with unit lengthscale, kernel magnitude, and likelihood variance. We fix the degrees of freedom in the likelihood to $\nu=3$. For LA and EP, we follow the methods designed in \citet{aki}. For VI and our hybrid training procedure, E and M steps each have 20 iterations. In the E-step we set the learning rate of natural gradient descent to be $0.001$. In the M-step we use the Adam optimizer \citep{adam} and set the learning rate to $0.001$. For {\sc neal} we set the maximal steps to be $2000$ and for {\sc boston} and {\sc stock} we set the maximal steps to be $5000$.

\begin{table}[b]
\centering\footnotesize
\caption{Data sets information for sparse binary classification.}
\setlength{\tabcolsep}{6pt}
%
\input{table/sparse_dataset.tex}
\label{table:sparse_dataset}
\end{table}

\section{Evaluation on Sparse Approximation}
\label{app:sparse_app}
To further study the proposed approach, we include results for a couple of sparse GP classification problems as proof-of-concept.

\subsection{Experiment Setup}
We initialize all methods with unit lengthscale and magnitude. We use $k$-means to select a set of $500$ inducing points and then fix the location of inducing points. Learning the inducing points would add an additional layer of optimization, and this is not in the scope of this paper. We use 20 iterations for both the E and M steps. In the E-step we set the learning rate of natural gradient descent to $0.1$. In the M-step we use the Adam optimizer \citep{adam} and set the learning rate to $0.01$. The data sets information are given in \cref{table:sparse_dataset}.

\end{document}

%% file: fig/tikz/fig1a.tex
\begin{tikzpicture}

\definecolor{darkgray176}{RGB}{176,176,176}
\definecolor{darkgray178}{RGB}{178,178,178}
\definecolor{darkorange25512714}{RGB}{255,127,14}
\definecolor{lightgray204}{RGB}{204,204,204}
\definecolor{steelblue31119180}{RGB}{31,119,180}

\begin{axis}[
height=\figureheight,
legend cell align={left},
legend style={fill opacity=0.8, draw opacity=1, text opacity=1, draw=none},
tick align=inside,
tick pos=left,
width=\figurewidth,
x grid style={darkgray176},
xmin=1, xmax=5,
xtick style={color=black},
y grid style={darkgray176},
ymin=-0.45, ymax=-0.25,
ytick style={color=black}
]
\addplot [ultra thick, darkorange25512714, dashed, forget plot,clip = false]
table {%
2.9 -0.50
2.9 -0.25
};
\addplot [ultra thick, steelblue31119180, dashed, forget plot,clip = false]
table {%
2.6 -0.50
2.6 -0.25
};
\addplot [smooth,line width=2.8pt, darkgray178]
table {%
-1 -0.657526546378798
-0.7 -0.619832838278508
-0.4 -0.566336369913481
-0.1 -0.502380451225104
0.2 -0.443281277826544
0.5 -0.397944387696695
0.8 -0.361751424078755
1.1 -0.330411565376686
1.4 -0.299996668426016
1.7 -0.279569390927532
2 -0.268254272907398
2.3 -0.263004731404356
2.6 -0.260724448607089
2.9 -0.265662309827204
3.2 -0.276973990756866
3.5 -0.294005566332735
3.8 -0.317531530803751
4.1 -0.345083825862569
4.4 -0.374487773006757
4.7 -0.407762060951404
5 -0.445417953413899
};
\addlegendentry{MCMC}
\addplot [smooth,ultra thick, darkorange25512714]
table {%
-1 -0.852449105361411
-0.7 -0.809336491309543
-0.4 -0.743428283589127
-0.1 -0.662281811371456
0.2 -0.582303813907074
0.5 -0.517204734187883
0.8 -0.466291135462464
1.1 -0.418675269592528
1.4 -0.371916022275593
1.7 -0.333514846300153
2 -0.306120929024164
2.3 -0.287309045041732
2.6 -0.275992617862783
2.9 -0.273372184628913
3.2 -0.280309705954786
3.5 -0.296006881827865
3.8 -0.318328013664613
4.1 -0.344850328827177
4.4 -0.374446231143997
4.7 -0.407505178307188
5 -0.444407047626132
};
\addlegendentry{VI ELBO}
\addplot [smooth,ultra thick, steelblue31119180]
table {%
-1 -0.672032418077585
-0.7 -0.638823029178238
-0.4 -0.582351043071147
-0.1 -0.512230421561897
0.2 -0.452395030727061
0.5 -0.407227568008314
0.8 -0.376132451863325
1.1 -0.344637235842882
1.4 -0.305339168104973
1.7 -0.284345381554071
2 -0.271748895119329
2.3 -0.266051976994096
2.6 -0.263564631765325
2.9 -0.266771847521193
3.2 -0.276988265354363
3.5 -0.294377394382898
3.8 -0.317535740249534
4.1 -0.344461700328236
4.4 -0.374260900083235
4.7 -0.407425096105093
5 -0.44437683421914
};
\addlegendentry{Ours}
\draw (axis cs:3,-0.42) node[
  scale=0.5,
  anchor=base west,
  text=black,
  rotate=0.0
]{\Large Training proxy for\ldots};
\end{axis}

\end{tikzpicture}

%% file: fig/tikz/fig1b.tex
\begin{tikzpicture}

\definecolor{darkgray176}{RGB}{176,176,176}
\definecolor{darkgray178}{RGB}{178,178,178}
\definecolor{darkorange25512714}{RGB}{255,127,14}
\definecolor{steelblue31119180}{RGB}{31,119,180}

\begin{axis}[
height=\figureheight,
tick align=inside,
tick pos=left,
width=\figurewidth,
x grid style={darkgray176},
xmin=1, xmax=5,
xtick style={color=black},
y grid style={darkgray176},
ymin=-0.3, ymax=-0.05,
ytick style={color=black}
]
\addplot [smooth,ultra thick, darkorange25512714]
table {%
-1 -0.645362489720902
-0.7 -0.582560037322829
-0.4 -0.502713084868549
-0.1 -0.420305052235048
0.2 -0.351015691653583
0.5 -0.317479892477142
0.8 -0.283750127527615
1.1 -0.244523809970588
1.4 -0.168930443682583
1.7 -0.129012558066751
2 -0.108155617711889
2.3 -0.100334941902686
2.6 -0.102493704558526
2.9 -0.113103136656933
3.2 -0.130273554447836
3.5 -0.151210479452939
3.8 -0.175659824137126
4.1 -0.204899405978022
4.4 -0.239411233200889
4.7 -0.280240277288176
5 -0.328299507811568
};
\addplot [ultra thick, darkorange25512714, dashed]
table {%
2.9 -0.3
2.9 -0.05
};
\addplot [semithick, darkorange25512714, mark=*, mark size=2.5, mark options={solid}]
table {%
2.9 -0.113103136656933
};
\addplot [ultra thick, steelblue31119180, dashed]
table {%
2.6 -0.3
2.6 -0.05
};
\addplot [semithick, steelblue31119180, mark=*, mark size=2.5, mark options={solid}]
table {%
2.6 -0.102493704558526
};
\draw (axis cs:2.9,-0.08) node[
  scale=0.5,
  anchor=base west,
  text=black,
  rotate=0.0
]{\Large\ldots test performance on holdout data};
\end{axis}

\end{tikzpicture}

%% file: fig/tikz/monks-2/CVI_nlpd.tex
\definecolor{aliceblue242247253}{RGB}{242,247,253}
\definecolor{cornflowerblue107174214}{RGB}{107,174,214}
\definecolor{cornflowerblue89162207}{RGB}{89,162,207}
\definecolor{darkgray176}{RGB}{176,176,176}
\definecolor{darkorange25512714}{RGB}{255,127,14}
\definecolor{lavender210227243}{RGB}{210,227,243}
\definecolor{lavender221234246}{RGB}{221,234,246}
\definecolor{lavender231240249}{RGB}{231,240,249}
\definecolor{lightblue168206228}{RGB}{168,206,228}
\definecolor{lightblue185213234}{RGB}{185,213,234}
\definecolor{midnightblue854116}{RGB}{8,54,116}
\definecolor{midnightblue868137}{RGB}{8,68,137}
\definecolor{powderblue200220239}{RGB}{200,220,239}
\definecolor{skyblue128185218}{RGB}{128,185,218}
\definecolor{skyblue149197223}{RGB}{149,197,223}
\definecolor{steelblue30109178}{RGB}{30,109,178}
\definecolor{steelblue31119180}{RGB}{31,119,180}
\definecolor{steelblue43123186}{RGB}{43,123,186}
\definecolor{steelblue57137193}{RGB}{57,137,193}
\definecolor{steelblue72150200}{RGB}{72,150,200}
\definecolor{teal1996167}{RGB}{19,96,167}
\definecolor{teal882156}{RGB}{8,82,156}

\begin{axis}[
height=\figureheight,
tick pos=left,
width=\figurewidth,
x grid style={darkgray176},
xmin=-1, xmax=5,
xtick style={color=black},
y grid style={darkgray176},
ymin=-1, ymax=5,
ytick style={color=black}
]
\addplot [draw=none, fill=steelblue57137193]
table{%
x  y
3.5 -1
3.8 -1
4.1 -1
4.4 -1
4.7 -1
5 -1
5 -0.7
5 -0.4
5 -0.1
5 0.2
5 0.5
5 0.8
5 0.866700785641344
4.93394724147957 0.8
4.7 0.561407648843911
4.63968392298039 0.5
4.4 0.251488161907083
4.35018905062141 0.2
4.1 -0.0675292906249814
4.06947178974349 -0.1
3.80688112167306 -0.4
3.8 -0.409975282931826
3.59734911207323 -0.7
3.5 -0.919483724543065
3.46360008351239 -1
3.5 -1
};
\addplot [draw=none, fill=steelblue72150200]
table{%
x  y
-0.7 -1
-0.401387118084647 -1
-0.610921143263856 -0.7
-0.7 -0.525872839850859
-0.767649724009323 -0.4
-0.885918500792056 -0.1
-0.954694726888944 0.2
-0.994778141166498 0.5
-1 0.563660783367841
-1 0.5
-1 0.2
-1 -0.1
-1 -0.4
-1 -0.7
-1 -1
-0.7 -1
};
\addplot [draw=none, fill=steelblue72150200]
table{%
x  y
1.1 -1
1.4 -1
1.7 -1
2 -1
2.3 -1
2.6 -1
2.9 -1
3.2 -1
3.46360008351239 -1
3.5 -0.919483724543065
3.59734911207323 -0.7
3.8 -0.409975282931826
3.80688112167306 -0.4
4.06947178974349 -0.1
4.1 -0.0675292906249814
4.35018905062141 0.2
4.4 0.251488161907083
4.63968392298039 0.5
4.7 0.561407648843911
4.93394724147957 0.8
5 0.866700785641344
5 1.1
5 1.4
5 1.7
5 2
5 2.3
5 2.6
5 2.9
5 3.2
5 3.5
5 3.8
5 4.1
5 4.4
5 4.7
5 5
4.7 5
4.58229621392939 5
4.42228059296982 4.7
4.4 4.65941611797549
4.27578733672414 4.4
4.11340302473509 4.1
4.1 4.07579502867969
3.96686700916216 3.8
3.80108147137237 3.5
3.8 3.49806790383929
3.65456792903415 3.2
3.5 2.92430554015234
3.48804839762113 2.9
3.33742665638949 2.6
3.2 2.35922135277619
3.17005552168569 2.3
3.01303558365545 2
2.9 1.80658126035337
2.84408975370515 1.7
2.67692207555554 1.4
2.6 1.27245744397784
2.50505795309519 1.1
2.31969605835521 0.8
2.3 0.768750570837571
2.1419066222835 0.5
2 0.289060609174155
1.94135527218723 0.2
1.72832363355529 -0.1
1.7 -0.138512985582413
1.50286791287801 -0.4
1.4 -0.526643888737113
1.24795093411793 -0.7
1.1 -0.867618513013066
0.969613657284833 -1
1.1 -1
};
\addplot [draw=none, fill=steelblue72150200]
table{%
x  y
-0.973088800481105 1.7
-0.93752398239976 2
-0.916442493173037 2.3
-0.90577832576915 2.6
-0.902125582470991 2.9
-0.901065635307669 3.2
-0.898643670107177 3.5
-0.897098542485723 3.8
-0.900292605468862 4.1
-0.903617777475129 4.4
-0.904331211985759 4.7
-0.900627574022617 5
-1 5
-1 4.7
-1 4.4
-1 4.1
-1 3.8
-1 3.5
-1 3.2
-1 2.9
-1 2.6
-1 2.3
-1 2
-1 1.7
-1 1.47558907772839
-0.973088800481105 1.7
};
\addplot [draw=none, fill=cornflowerblue89162207]
table{%
x  y
-0.4 -1
-0.1 -1
0.2 -1
0.5 -1
0.8 -1
0.969613657284833 -1
1.1 -0.867618513013066
1.24795093411793 -0.7
1.4 -0.526643888737113
1.50286791287801 -0.4
1.7 -0.138512985582413
1.72832363355529 -0.1
1.94135527218723 0.2
2 0.289060609174155
2.1419066222835 0.5
2.3 0.768750570837571
2.31969605835521 0.8
2.50505795309519 1.1
2.6 1.27245744397784
2.67692207555554 1.4
2.84408975370515 1.7
2.9 1.80658126035337
3.01303558365545 2
3.17005552168569 2.3
3.2 2.35922135277619
3.33742665638949 2.6
3.48804839762113 2.9
3.5 2.92430554015234
3.65456792903415 3.2
3.8 3.49806790383929
3.80108147137237 3.5
3.96686700916216 3.8
4.1 4.07579502867969
4.11340302473509 4.1
4.27578733672414 4.4
4.4 4.65941611797549
4.42228059296982 4.7
4.58229621392939 5
4.4 5
4.16593156268969 5
4.1 4.88195610594283
4.01400765735037 4.7
3.85690068613926 4.4
3.8 4.29891498183911
3.70479243883918 4.1
3.54343918405678 3.8
3.5 3.72366120942108
3.39085172831279 3.5
3.2232180002866 3.2
3.2 3.1598105566115
3.06953657788932 2.9
2.9 2.61050907529907
2.89440310494009 2.6
2.73651467350732 2.3
2.6 2.07434550475391
2.55832438802813 2
2.38445427499338 1.7
2.3 1.56642428113258
2.19989834627524 1.4
2.00037777841878 1.1
2 1.09943724480049
1.80217890246942 0.8
1.7 0.658516768858748
1.58239870197982 0.5
1.4 0.269673929921296
1.34173350118695 0.2
1.1 -0.074820215280111
1.07620664957899 -0.1
0.8 -0.38038606171185
0.777828052393891 -0.4
0.5 -0.642102239833821
0.393614369033412 -0.7
0.2 -0.810755891928953
-0.1 -0.769532414285159
-0.169768990225931 -0.7
-0.4 -0.48014559028799
-0.451252576693237 -0.4
-0.606088916338512 -0.1
-0.7 0.176215943121349
-0.708435202051775 0.2
-0.783160581601076 0.5
-0.825440718246898 0.8
-0.846939400830174 1.1
-0.848587710754941 1.4
-0.834542566623308 1.7
-0.818804362151896 2
-0.807637154672476 2.3
-0.79923274200803 2.6
-0.794978888401598 2.9
-0.793319349601961 3.2
-0.789260947089554 3.5
-0.786672348270988 3.8
-0.790522779626535 4.1
-0.795678304235699 4.4
-0.797715579180067 4.7
-0.79561636181783 5
-0.900627574022617 5
-0.904331211985759 4.7
-0.903617777475129 4.4
-0.900292605468861 4.1
-0.897098542485723 3.8
-0.898643670107177 3.5
-0.901065635307669 3.2
-0.902125582470991 2.9
-0.90577832576915 2.6
-0.916442493173037 2.3
-0.93752398239976 2
-0.973088800481104 1.7
-1 1.47558907772839
-1 1.4
-1 1.1
-1 0.8
-1 0.563660783367841
-0.994778141166498 0.5
-0.954694726888944 0.2
-0.885918500792056 -0.1
-0.767649724009323 -0.4
-0.7 -0.525872839850859
-0.610921143263856 -0.7
-0.401387118084647 -1
-0.4 -1
};
\addplot [draw=none, fill=cornflowerblue107174214]
table{%
x  y
-0.1 -0.769532414285159
0.2 -0.810755891928954
0.393614369033412 -0.7
0.5 -0.642102239833821
0.777828052393891 -0.4
0.8 -0.38038606171185
1.07620664957899 -0.1
1.1 -0.074820215280111
1.34173350118695 0.2
1.4 0.269673929921296
1.58239870197982 0.5
1.7 0.658516768858748
1.80217890246942 0.8
2 1.09943724480049
2.00037777841878 1.1
2.19989834627524 1.4
2.3 1.56642428113258
2.38445427499338 1.7
2.55832438802813 2
2.6 2.07434550475391
2.73651467350732 2.3
2.89440310494009 2.6
2.9 2.61050907529907
3.06953657788932 2.9
3.2 3.1598105566115
3.2232180002866 3.2
3.39085172831279 3.5
3.5 3.72366120942108
3.54343918405678 3.8
3.70479243883918 4.1
3.8 4.29891498183911
3.85690068613926 4.4
4.01400765735037 4.7
4.1 4.88195610594283
4.16593156268969 5
4.1 5
3.87729716324804 5
3.8 4.85577731134943
3.72021019070728 4.7
3.56760792371611 4.4
3.5 4.27585131019429
3.40884153126798 4.1
3.25183696346892 3.8
3.2 3.70682225458552
3.0905825658456 3.5
2.92661453275151 3.2
2.9 3.15350305797968
2.761613085494 2.9
2.6 2.62085892648707
2.58792661815823 2.6
2.41566654576461 2.3
2.3 2.11106481244931
2.23165394806487 2
2.04198252194048 1.7
2 1.63600989297291
1.84236757094471 1.4
1.7 1.19795269615871
1.62778495944357 1.1
1.4 0.80382196173868
1.39687972677067 0.8
1.14901110548785 0.5
1.1 0.443057364672972
0.877269348221013 0.2
0.8 0.118968812988055
0.566606946865673 -0.1
0.5 -0.160162935232004
0.2 -0.349145952888071
-0.1 -0.351757170003525
-0.351268511829671 -0.1
-0.4 -0.0417949549538859
-0.505999990409831 0.2
-0.590829714937551 0.5
-0.642025859262873 0.8
-0.671619040126784 1.1
-0.687575822923122 1.4
-0.695923485976857 1.7
-0.7 1.99276836306229
-0.700084741904032 2
-0.7 2.0187017304931
-0.698632816630875 2.3
-0.69173247491338 2.6
-0.686639747645324 2.9
-0.684361512182041 3.2
-0.67926821678047 3.5
-0.676653947689119 3.8
-0.680729866358258 4.1
-0.687408678229579 4.4
-0.690597409393281 4.7
-0.689574699971748 5
-0.7 5
-0.79561636181783 5
-0.797715579180067 4.7
-0.795678304235699 4.4
-0.790522779626535 4.1
-0.786672348270988 3.8
-0.789260947089554 3.5
-0.793319349601961 3.2
-0.794978888401598 2.9
-0.79923274200803 2.6
-0.807637154672476 2.3
-0.818804362151896 2
-0.834542566623308 1.7
-0.848587710754941 1.4
-0.846939400830174 1.1
-0.825440718246898 0.8
-0.783160581601076 0.5
-0.708435202051775 0.2
-0.7 0.176215943121349
-0.606088916338512 -0.1
-0.451252576693237 -0.4
-0.4 -0.48014559028799
-0.169768990225931 -0.7
-0.1 -0.769532414285159
};
\addplot [draw=none, fill=skyblue128185218]
table{%
x  y
-0.1 -0.351757170003525
0.2 -0.349145952888071
0.5 -0.160162935232004
0.566606946865673 -0.1
0.8 0.118968812988055
0.877269348221013 0.2
1.1 0.443057364672972
1.14901110548785 0.5
1.39687972677067 0.8
1.4 0.80382196173868
1.62778495944357 1.1
1.7 1.19795269615871
1.84236757094471 1.4
2 1.63600989297291
2.04198252194048 1.7
2.23165394806487 2
2.3 2.11106481244931
2.41566654576461 2.3
2.58792661815823 2.6
2.6 2.62085892648707
2.761613085494 2.9
2.9 3.15350305797968
2.92661453275151 3.2
3.0905825658456 3.5
3.2 3.70682225458552
3.25183696346892 3.8
3.40884153126798 4.1
3.5 4.27585131019429
3.56760792371611 4.4
3.72021019070728 4.7
3.8 4.85577731134943
3.87729716324804 5
3.8 5
3.60837937410639 5
3.5 4.78198424606871
3.45753755769613 4.7
3.30106272591697 4.4
3.2 4.20333099987322
3.14550669987433 4.1
2.98548979129639 3.8
2.9 3.64029456717598
2.8236392768057 3.5
2.65755350800649 3.2
2.6 3.09775628951449
2.48707652050314 2.9
2.31073820094493 2.6
2.3 2.58206292501778
2.12782877334403 2.3
2 2.09856735837383
1.93590082434325 2
1.73248783168992 1.7
1.7 1.65320125247758
1.51554516481987 1.4
1.4 1.2474316603343
1.2806568860651 1.1
1.1 0.882150531289876
1.02612642796479 0.8
0.8 0.550148206339095
0.749333827216355 0.5
0.5 0.255650688430853
0.421918955708316 0.2
0.2 0.0475623564267083
-0.1 0.0393931064888305
-0.237108004138517 0.2
-0.4 0.465647751087341
-0.410984683020414 0.5
-0.475058065757026 0.8
-0.513441820876019 1.1
-0.537635502474422 1.4
-0.554856401202756 1.7
-0.566572732806502 2
-0.57129254218635 2.3
-0.571277484041814 2.6
-0.568992662310518 2.9
-0.567566860544159 3.2
-0.566569471417752 3.5
-0.568123492758422 3.8
-0.570828368222049 4.1
-0.576562751640207 4.4
-0.577961779349203 4.7
-0.573045607067269 5
-0.689574699971748 5
-0.690597409393281 4.7
-0.687408678229579 4.4
-0.680729866358258 4.1
-0.676653947689119 3.8
-0.67926821678047 3.5
-0.684361512182041 3.2
-0.686639747645324 2.9
-0.69173247491338 2.6
-0.698632816630875 2.3
-0.7 2.0187017304931
-0.700084741904032 2
-0.7 1.99276836306229
-0.695923485976857 1.7
-0.687575822923122 1.4
-0.671619040126784 1.1
-0.642025859262873 0.8
-0.590829714937551 0.5
-0.505999990409831 0.2
-0.4 -0.0417949549538859
-0.351268511829671 -0.1
-0.1 -0.351757170003525
};
\addplot [draw=none, fill=skyblue149197223]
table{%
x  y
-0.1 0.0393931064888305
0.2 0.0475623564267083
0.421918955708316 0.2
0.5 0.255650688430853
0.749333827216355 0.5
0.8 0.550148206339095
1.02612642796479 0.8
1.1 0.882150531289876
1.2806568860651 1.1
1.4 1.2474316603343
1.51554516481987 1.4
1.7 1.65320125247758
1.73248783168992 1.7
1.93590082434325 2
2 2.09856735837383
2.12782877334403 2.3
2.3 2.58206292501778
2.31073820094493 2.6
2.48707652050314 2.9
2.6 3.09775628951449
2.65755350800649 3.2
2.8236392768057 3.5
2.9 3.64029456717598
2.98548979129639 3.8
3.14550669987433 4.1
3.2 4.20333099987322
3.30106272591697 4.4
3.45753755769613 4.7
3.5 4.78198424606871
3.60837937410639 5
3.5 5
3.3297349934517 5
3.2 4.74348708709207
3.18085920974583 4.7
3.03268679920533 4.4
2.9 4.14812298432434
2.87744378426254 4.1
2.72246170603258 3.8
2.6 3.57819940647955
2.56029717081476 3.5
2.39425616617915 3.2
2.3 3.03870103013506
2.22229759003909 2.9
2.03994045736902 2.6
2 2.53630503479093
1.85127685350172 2.3
1.7 2.07493353639895
1.64643380974521 2
1.42917913313154 1.7
1.4 1.66081369092873
1.18773271087328 1.4
1.1 1.29258410111815
0.920953393683706 1.1
0.8 0.965349502694235
0.626579044174047 0.8
0.5 0.674786645271004
0.253316288419387 0.5
0.2 0.462771913160861
-0.1 0.487762737720482
-0.107968269759753 0.5
-0.243172654460904 0.8
-0.324007592156538 1.1
-0.378295053480192 1.4
-0.4 1.53691546271839
-0.413789316428656 1.7
-0.433050157379628 2
-0.443952267741826 2.3
-0.450822493170248 2.6
-0.451345576975712 2.9
-0.450772208906276 3.2
-0.453870726055033 3.5
-0.459593037827724 3.8
-0.460926870085839 4.1
-0.465716825050834 4.4
-0.465326149305126 4.7
-0.45651651416279 5
-0.573045607067269 5
-0.577961779349203 4.7
-0.576562751640207 4.4
-0.570828368222049 4.1
-0.568123492758422 3.8
-0.566569471417752 3.5
-0.567566860544159 3.2
-0.568992662310518 2.9
-0.571277484041814 2.6
-0.57129254218635 2.3
-0.566572732806502 2
-0.554856401202756 1.7
-0.537635502474422 1.4
-0.513441820876019 1.1
-0.475058065757026 0.8
-0.410984683020414 0.5
-0.4 0.465647751087341
-0.237108004138517 0.2
-0.1 0.0393931064888305
};
\addplot [draw=none, fill=lightblue168206228]
table{%
x  y
-0.1 0.487762737720482
0.2 0.462771913160861
0.253316288419387 0.5
0.5 0.674786645271004
0.626579044174047 0.8
0.8 0.965349502694235
0.920953393683706 1.1
1.1 1.29258410111815
1.18773271087328 1.4
1.4 1.66081369092873
1.42917913313154 1.7
1.64643380974521 2
1.7 2.07493353639895
1.85127685350172 2.3
2 2.53630503479093
2.03994045736902 2.6
2.22229759003909 2.9
2.3 3.03870103013506
2.39425616617915 3.2
2.56029717081476 3.5
2.6 3.57819940647955
2.72246170603258 3.8
2.87744378426254 4.1
2.9 4.14812298432434
3.03268679920533 4.4
3.18085920974583 4.7
3.2 4.74348708709207
3.3297349934517 5
3.2 5
3.08319591775472 5
2.93193417104688 4.7
2.9 4.63976609165488
2.79207821423358 4.4
2.63617922150406 4.1
2.6 4.03425534503101
2.48431541226072 3.8
2.3184809449257 3.5
2.3 3.46802714500605
2.15244362725356 3.2
2 2.94474526373587
1.97085538724916 2.9
1.78334530192918 2.6
1.7 2.47237748573775
1.57100861541365 2.3
1.4 2.05996077319101
1.34301800475059 2
1.1 1.7066725571839
1.09177523145255 1.7
0.8 1.41511568886182
0.77759877677076 1.4
0.5 1.17985069634708
0.336886365595712 1.1
0.2 1.03005318698845
0.0866334590075044 1.1
-0.1 1.32832251462334
-0.113809496415877 1.4
-0.154204068442713 1.7
-0.184229030058284 2
-0.209347348122818 2.3
-0.231674145892733 2.6
-0.241791827232274 2.9
-0.249251474763737 3.2
-0.272158342158684 3.5
-0.29345239340322 3.8
-0.288181590874136 4.1
-0.282858293731449 4.4
-0.29244869033008 4.7
-0.291130249157995 5
-0.4 5
-0.45651651416279 5
-0.465326149305126 4.7
-0.465716825050834 4.4
-0.460926870085839 4.1
-0.459593037827724 3.8
-0.453870726055033 3.5
-0.450772208906276 3.2
-0.451345576975712 2.9
-0.450822493170248 2.6
-0.443952267741826 2.3
-0.433050157379628 2
-0.413789316428656 1.7
-0.4 1.53691546271839
-0.378295053480192 1.4
-0.324007592156538 1.1
-0.243172654460904 0.8
-0.107968269759753 0.5
-0.1 0.487762737720482
};
\addplot [draw=none, fill=lightblue168206228]
table{%
x  y
0.2 4.21067659974691
0.303627644598719 4.4
0.256955579495328 4.7
0.2 4.96580960373697
0.181453207505082 4.7
0.13928150287395 4.4
0.2 4.21067659974691
};
\addplot [draw=none, fill=lightblue168206228]
table{%
x  y
0.8 4.39268695330918
0.801232396423631 4.4
0.886237457677512 4.7
0.912148881453731 5
0.8 5
0.642722455562214 5
0.622757676735703 4.7
0.796458951457711 4.4
0.8 4.39268695330918
};
\addplot [draw=none, fill=lightblue185213234]
table{%
x  y
0.2 1.03005318698845
0.336886365595712 1.1
0.5 1.17985069634708
0.77759877677076 1.4
0.8 1.41511568886182
1.09177523145255 1.7
1.1 1.7066725571839
1.34301800475059 2
1.4 2.05996077319101
1.57100861541365 2.3
1.7 2.47237748573775
1.78334530192918 2.6
1.97085538724916 2.9
2 2.94474526373587
2.15244362725356 3.2
2.3 3.46802714500605
2.3184809449257 3.5
2.48431541226072 3.8
2.6 4.03425534503101
2.63617922150406 4.1
2.79207821423358 4.4
2.9 4.63976609165488
2.93193417104688 4.7
3.08319591775472 5
2.9 5
2.87269374276269 5
2.73349407265243 4.7
2.6 4.43460500713725
2.58081426699069 4.4
2.4281069199732 4.1
2.3 3.85703159148165
2.26437375815832 3.8
2.09525082357371 3.5
2 3.33211435590567
1.9051314209887 3.2
1.71966811680823 2.9
1.7 2.86869652013216
1.47586318335071 2.6
1.4 2.49342190962083
1.16914552146703 2.3
1.1 2.22552971980734
0.8 2.16433651963283
0.642854141449314 2.3
0.592594848502194 2.6
0.654300454957084 2.9
0.8 3.19463229817836
0.801401528433515 3.2
0.916810264884713 3.5
1.02086714318073 3.8
1.09495776100568 4.1
1.1 4.13377164083177
1.1469310312543 4.4
1.21421655208569 4.7
1.30687505318079 5
1.1 5
0.912148881453731 5
0.886237457677512 4.7
0.801232396423631 4.4
0.8 4.39268695330918
0.796458951457711 4.4
0.622757676735703 4.7
0.642722455562214 5
0.5 5
0.2 5
-0.1 5
-0.291130249157995 5
-0.29244869033008 4.7
-0.282858293731449 4.4
-0.288181590874136 4.1
-0.29345239340322 3.8
-0.272158342158684 3.5
-0.249251474763737 3.2
-0.241791827232274 2.9
-0.231674145892733 2.6
-0.209347348122818 2.3
-0.184229030058284 2
-0.154204068442713 1.7
-0.113809496415877 1.4
-0.1 1.32832251462334
0.0866334590075044 1.1
0.2 1.03005318698845

0.13928150287395 4.4
0.181453207505082 4.7
0.2 4.96580960373697
0.256955579495328 4.7
0.303627644598719 4.4
0.2 4.21067659974691
0.13928150287395 4.4
};
\addplot [draw=none, fill=powderblue200220239]
table{%
x  y
0.8 2.16433651963283
1.1 2.22552971980734
1.16914552146703 2.3
1.4 2.49342190962082
1.47586318335071 2.6
1.7 2.86869652013216
1.71966811680823 2.9
1.9051314209887 3.2
2 3.33211435590567
2.09525082357371 3.5
2.26437375815832 3.8
2.3 3.85703159148165
2.4281069199732 4.1
2.58081426699069 4.4
2.6 4.43460500713725
2.73349407265243 4.7
2.87269374276269 5
2.68170160146474 5
2.6 4.82393461193561
2.52347416729288 4.7
2.37027795030906 4.4
2.3 4.25700057364826
2.18849060831495 4.1
2.02479459106807 3.8
2 3.75392416263128
1.78515915535508 3.5
1.7 3.36298344092437
1.42031831759103 3.2
1.4 3.18080555432156
1.3871508956594 3.2
1.34457547128482 3.5
1.38356170436409 3.8
1.4 3.86368236341753
1.46141259839273 4.1
1.52171168578756 4.4
1.59399771653877 4.7
1.68028525654456 5
1.4 5
1.30687505318079 5
1.21421655208569 4.7
1.1469310312543 4.4
1.1 4.13377164083177
1.09495776100568 4.1
1.02086714318073 3.8
0.916810264884713 3.5
0.801401528433515 3.2
0.8 3.19463229817836
0.654300454957084 2.9
0.592594848502194 2.6
0.642854141449314 2.3
0.8 2.16433651963283
};
\addplot [draw=none, fill=lavender210227243]
table{%
x  y
1.4 3.18080555432156
1.42031831759103 3.2
1.7 3.36298344092437
1.78515915535508 3.5
2 3.75392416263128
2.02479459106807 3.8
2.18849060831495 4.1
2.3 4.25700057364826
2.37027795030906 4.4
2.52347416729288 4.7
2.6 4.82393461193561
2.68170160146474 5
2.6 5
2.38840510940667 5
2.3 4.80062762914037
2.1221775460576 4.7
2 4.50968753418149
1.96697010816543 4.7
1.99900419395596 5
1.7 5
1.68028525654456 5
1.59399771653877 4.7
1.52171168578756 4.4
1.46141259839273 4.1
1.4 3.86368236341753
1.38356170436409 3.8
1.34457547128482 3.5
1.3871508956594 3.2
1.4 3.18080555432156
};
\addplot [draw=none, fill=lavender221234246]
table{%
x  y
2 4.50968753418149
2.1221775460576 4.7
2.3 4.80062762914037
2.38840510940667 5
2.3 5
2 5
1.99900419395596 5
1.96697010816543 4.7
2 4.50968753418149
};
\addplot [semithick, steelblue31119180, mark=*, mark size=2.5, mark options={solid,fill=black,draw=white}]
table {%
0.5 1.1
};
\addplot [semithick, darkorange25512714, mark=mystar, mark size=2.5, mark options={solid,fill=black,draw=white}]
table {%
0.2 1.7
};
\draw (axis cs:0.7,1.1) node[
  scale=0.5,
  anchor=base west,
  text=black,
  rotate=0.0
]{$-0.521$};
\draw (axis cs:0.4,1.7) node[
  scale=0.5,
  anchor=base west,
  text=black,
  rotate=0.0
]{$-0.501$};
\coordinate (CVI_mean_lml_b) at (axis cs:0.5,1.1);
\coordinate (CVI_mean_lml_ep_b) at (axis cs:0.19999999999999996,1.6999999999999997);
\end{axis}

%% file: table/seed_test_lpd.tex
\begin{tabular}{lc|cccc|c}
\toprule
{} &($n$, $d$)     &          LA         &          EP         &          VI         &         Ours        &         MCMC        \\
\midrule
    {\sc trains}     &       (10, 30)       & $  -0.702{\pm}0.025  $ & $\bf -0.698{\pm}0.033$ & $  -0.702{\pm}0.037  $ & $\bf -0.691{\pm}0.046$ & $  -0.692{\pm}0.025  $ \\
   {\sc balloons}    &       (16, 5)        & $  -0.660{\pm}0.125  $ & $  -0.650{\pm}0.128  $ & $  -0.649{\pm}0.185  $ & $\bf -0.607{\pm}0.227$ & $  -0.684{\pm}0.076  $ \\
  {\sc fertility}    &      (100, 10)       & $\bf -0.388{\pm}0.122$ & $\bf -0.384{\pm}0.149$ & $\bf -0.393{\pm}0.136$ & $\bf -0.397{\pm}0.139$ & $  -0.382{\pm}0.126  $ \\
{\sc pittsburg-bridges-T-OR-D} &       (102, 8)       & $\bf -0.299{\pm}0.081$ & $  -0.321{\pm}0.108  $ & $\bf -0.290{\pm}0.110$ & $\bf -0.293{\pm}0.116$ & $  -0.306{\pm}0.115  $ \\
{\sc acute-nephritis} &       (120, 7)       & $  -0.203{\pm}0.012  $ & $  -0.046{\pm}0.007  $ & $  -0.007{\pm}0.002  $ & $\bf -0.005{\pm}0.002$ & $  -0.005{\pm}0.002  $ \\
{\sc acute-inflammation} &       (120, 7)       & $  -0.184{\pm}0.018  $ & $  -0.052{\pm}0.007  $ & $  -0.007{\pm}0.002  $ & $\bf -0.007{\pm}0.002$ & $  -0.007{\pm}0.003  $ \\
{\sc echocardiogram} &      (131, 11)       & $  -0.424{\pm}0.093  $ & $\bf -0.418{\pm}0.095$ & $\bf -0.425{\pm}0.110$ & $  -0.428{\pm}0.112  $ & $  -0.437{\pm}0.127  $ \\
  {\sc hepatitis}    &      (155, 20)       & $\bf -0.370{\pm}0.071$ & $  -0.372{\pm}0.072  $ & $\bf -0.364{\pm}0.090$ & $  -0.367{\pm}0.094  $ & $  -0.369{\pm}0.091  $ \\
  {\sc parkinsons}   &      (195, 23)       & $  -0.260{\pm}0.031  $ & $  -0.295{\pm}0.056  $ & $  -0.160{\pm}0.050  $ & $\bf -0.141{\pm}0.046$ & $  -0.145{\pm}0.044  $ \\
{\sc breast-cancer-wisc-prog} &      (198, 34)       & $\bf -0.458{\pm}0.075$ & $  -0.473{\pm}0.091  $ & $\bf -0.457{\pm}0.085$ & $\bf -0.460{\pm}0.088$ & $  -0.464{\pm}0.085  $ \\
    {\sc spect}      &      (265, 23)       & $\bf -0.593{\pm}0.049$ & $\bf -0.590{\pm}0.055$ & $  -0.594{\pm}0.054  $ & $  -0.595{\pm}0.054  $ & $  -0.596{\pm}0.051  $ \\
{\sc statlog-heart}  &      (270, 14)       & $  -0.395{\pm}0.064  $ & $\bf -0.389{\pm}0.061$ & $  -0.396{\pm}0.071  $ & $  -0.397{\pm}0.071  $ & $  -0.397{\pm}0.070  $ \\
{\sc haberman-survival} &       (306, 4)       & $\bf -0.530{\pm}0.053$ & $  -0.532{\pm}0.059  $ & $  -0.531{\pm}0.055  $ & $  -0.531{\pm}0.055  $ & $  -0.520{\pm}0.063  $ \\
  {\sc ionosphere}   &      (351, 34)       & $  -0.224{\pm}0.042  $ & $  -0.230{\pm}0.042  $ & $\bf -0.170{\pm}0.048$ & $\bf -0.170{\pm}0.055$ & $  -0.179{\pm}0.058  $ \\
 {\sc horse-colic}   &      (368, 26)       & $  -0.463{\pm}0.059  $ & $\bf -0.452{\pm}0.057$ & $  -0.467{\pm}0.072  $ & $  -0.473{\pm}0.082  $ & $  -0.469{\pm}0.079  $ \\
{\sc congressional-voting} &      (435, 17)       & $  -0.640{\pm}0.028  $ & $\bf -0.639{\pm}0.030$ & $  -0.641{\pm}0.030  $ & $  -0.642{\pm}0.029  $ & $  -0.644{\pm}0.027  $ \\
{\sc cylinder-bands} &      (512, 36)       & $  -0.488{\pm}0.038  $ & $  -0.500{\pm}0.041  $ & $  -0.465{\pm}0.049  $ & $\bf -0.451{\pm}0.052$ & $  -0.451{\pm}0.049  $ \\
{\sc breast-cancer-wisc-diag} &      (569, 31)       & $  -0.085{\pm}0.026  $ & $  -0.140{\pm}0.020  $ & $  -0.077{\pm}0.044  $ & $\bf -0.075{\pm}0.045$ & $  -0.076{\pm}0.043  $ \\
{\sc ilpd-indian-liver} &      (583, 10)       & $\bf -0.513{\pm}0.040$ & $  -0.520{\pm}0.041  $ & $\bf -0.512{\pm}0.043$ & $\bf -0.512{\pm}0.043$ & $  -0.512{\pm}0.042  $ \\
   {\sc monks-2}     &       (601, 7)       & $  -0.491{\pm}0.025  $ & $  -0.512{\pm}0.028  $ & $  -0.464{\pm}0.031  $ & $\bf -0.442{\pm}0.033$ & $  -0.437{\pm}0.032  $ \\
{\sc statlog-australian-credit} &      (690, 15)       & $\bf -0.630{\pm}0.026$ & $  -0.639{\pm}0.036  $ & $\bf -0.630{\pm}0.026$ & $\bf -0.630{\pm}0.026$ & $  -0.630{\pm}0.025  $ \\
{\sc credit-approval} &      (690, 16)       & $\bf -0.342{\pm}0.047$ & $\bf -0.342{\pm}0.050$ & $\bf -0.341{\pm}0.052$ & $  -0.342{\pm}0.052  $ & $  -0.341{\pm}0.052  $ \\
{\sc breast-cancer-wisc} &      (699, 10)       & $\bf -0.094{\pm}0.025$ & $\bf -0.093{\pm}0.023$ & $\bf -0.093{\pm}0.029$ & $  -0.093{\pm}0.029  $ & $  -0.093{\pm}0.029  $ \\
    {\sc blood}      &       (748, 5)       & $\bf -0.478{\pm}0.039$ & $\bf -0.479{\pm}0.040$ & $  -0.478{\pm}0.039  $ & $  -0.478{\pm}0.039  $ & $  -0.478{\pm}0.039  $ \\
     {\sc pima}      &       (768, 9)       & $\bf -0.474{\pm}0.033$ & $\bf -0.476{\pm}0.038$ & $\bf -0.474{\pm}0.035$ & $\bf -0.474{\pm}0.035$ & $  -0.474{\pm}0.035  $ \\
 {\sc mammographic}  &       (961, 6)       & $\bf -0.407{\pm}0.038$ & $\bf -0.407{\pm}0.040$ & $\bf -0.408{\pm}0.040$ & $  -0.408{\pm}0.040  $ & $  -0.408{\pm}0.040  $ \\
{\sc statlog-german-credit} &      (1000, 25)      & $\bf -0.491{\pm}0.030$ & $\bf -0.491{\pm}0.032$ & $\bf -0.492{\pm}0.032$ & $\bf -0.492{\pm}0.032$ & $  -0.492{\pm}0.032  $ \\
\midrule
    {Bold Count}     & {}& $14$           & $13$           & $13$           & $16$           & $/$            \\
\bottomrule\end{tabular}

%% file: fig/tikz/classification_relative_acc.tex
\begin{tikzpicture}

\definecolor{darkgray176}{RGB}{176,176,176}
\definecolor{darkorange25512714}{RGB}{255,127,14}
\definecolor{forestgreen4416044}{RGB}{44,160,44}
\definecolor{lightgray204}{RGB}{204,204,204}
\definecolor{steelblue31119180}{RGB}{31,119,180}

\begin{axis}[
height=\figureheight,
legend cell align={left},
legend style={
  fill opacity=1,
  fill=white,
  draw opacity=1,
  text opacity=1,
  at={(0.97,0.03)},
  anchor=south east,
  draw=lightgray204
},
tick align=outside,
tick pos=left,
width=\figurewidth,
x grid style={darkgray176},
xlabel={Data set \#},
xmin=-0.3, xmax=28.3,
xtick style={color=black},
y grid style={darkgray176},
ylabel={Relative accuracy},
ymin=0.86, ymax=1,
ytick style={color=black}
]
\addplot [semithick, steelblue31119180, opacity=0.5, mark=*, mark size=2, mark options={solid}, only marks, forget plot]
table {%
1 0.875
};
\addplot [semithick, darkorange25512714, opacity=0.5, mark=square*, mark size=1.75, mark options={solid}, only marks, forget plot]
table {%
1 1
};
\addplot [semithick, forestgreen4416044, opacity=0.5, mark=triangle*, mark size=2.5, mark options={solid}, only marks, forget plot]
table {%
1 1
};
\addplot [semithick, black, opacity=0.5, mark=mystar, mark size=3, mark options={solid}, only marks, forget plot]
table {%
1 1
};
\addplot [semithick, steelblue31119180, opacity=0.5, mark=*, mark size=2, mark options={solid}, only marks, forget plot]
table {%
2 1
};
\addplot [semithick, darkorange25512714, opacity=0.5, mark=square*, mark size=1.75, mark options={solid}, only marks, forget plot]
table {%
2 1
};
\addplot [semithick, forestgreen4416044, opacity=0.5, mark=triangle*, mark size=2.5, mark options={solid}, only marks, forget plot]
table {%
2 1
};
\addplot [semithick, black, opacity=0.5, mark=mystar, mark size=3, mark options={solid}, only marks, forget plot]
table {%
2 1
};
\addplot [semithick, steelblue31119180, opacity=0.5, mark=*, mark size=2, mark options={solid}, only marks, forget plot]
table {%
3 1
};
\addplot [semithick, darkorange25512714, opacity=0.5, mark=square*, mark size=1.75, mark options={solid}, only marks, forget plot]
table {%
3 1
};
\addplot [semithick, forestgreen4416044, opacity=0.5, mark=triangle*, mark size=2.5, mark options={solid}, only marks, forget plot]
table {%
3 1
};
\addplot [semithick, black, opacity=0.5, mark=mystar, mark size=3, mark options={solid}, only marks, forget plot]
table {%
3 1
};
\addplot [semithick, steelblue31119180, opacity=0.5, mark=*, mark size=2, mark options={solid}, only marks, forget plot]
table {%
4 0.977346278317152
};
\addplot [semithick, darkorange25512714, opacity=0.5, mark=square*, mark size=1.75, mark options={solid}, only marks, forget plot]
table {%
4 1
};
\addplot [semithick, forestgreen4416044, opacity=0.5, mark=triangle*, mark size=2.5, mark options={solid}, only marks, forget plot]
table {%
4 0.977346278317152
};
\addplot [semithick, black, opacity=0.5, mark=mystar, mark size=3, mark options={solid}, only marks, forget plot]
table {%
4 0.977346278317152
};
\addplot [semithick, steelblue31119180, opacity=0.5, mark=*, mark size=2, mark options={solid}, only marks, forget plot]
table {%
5 1
};
\addplot [semithick, darkorange25512714, opacity=0.5, mark=square*, mark size=1.75, mark options={solid}, only marks, forget plot]
table {%
5 1
};
\addplot [semithick, forestgreen4416044, opacity=0.5, mark=triangle*, mark size=2.5, mark options={solid}, only marks, forget plot]
table {%
5 1
};
\addplot [semithick, black, opacity=0.5, mark=mystar, mark size=3, mark options={solid}, only marks, forget plot]
table {%
5 1
};
\addplot [semithick, steelblue31119180, opacity=0.5, mark=*, mark size=2, mark options={solid}, only marks, forget plot]
table {%
6 1
};
\addplot [semithick, darkorange25512714, opacity=0.5, mark=square*, mark size=1.75, mark options={solid}, only marks, forget plot]
table {%
6 1
};
\addplot [semithick, forestgreen4416044, opacity=0.5, mark=triangle*, mark size=2.5, mark options={solid}, only marks, forget plot]
table {%
6 1
};
\addplot [semithick, black, opacity=0.5, mark=mystar, mark size=3, mark options={solid}, only marks, forget plot]
table {%
6 1
};
\addplot [semithick, steelblue31119180, opacity=0.5, mark=*, mark size=2, mark options={solid}, only marks, forget plot]
table {%
7 0.954576043068641
};
\addplot [semithick, darkorange25512714, opacity=0.5, mark=square*, mark size=1.75, mark options={solid}, only marks, forget plot]
table {%
7 1
};
\addplot [semithick, forestgreen4416044, opacity=0.5, mark=triangle*, mark size=2.5, mark options={solid}, only marks, forget plot]
table {%
7 0.927321668909825
};
\addplot [semithick, black, opacity=0.5, mark=mystar, mark size=3, mark options={solid}, only marks, forget plot]
table {%
7 0.9185733512786
};
\addplot [semithick, steelblue31119180, opacity=0.5, mark=*, mark size=2, mark options={solid}, only marks, forget plot]
table {%
8 0.976744186046512
};
\addplot [semithick, darkorange25512714, opacity=0.5, mark=square*, mark size=1.75, mark options={solid}, only marks, forget plot]
table {%
8 0.976744186046512
};
\addplot [semithick, forestgreen4416044, opacity=0.5, mark=triangle*, mark size=2.5, mark options={solid}, only marks, forget plot]
table {%
8 0.976744186046512
};
\addplot [semithick, black, opacity=0.5, mark=mystar, mark size=3, mark options={solid}, only marks, forget plot]
table {%
8 1
};
\addplot [semithick, steelblue31119180, opacity=0.5, mark=*, mark size=2, mark options={solid}, only marks, forget plot]
table {%
9 1
};
\addplot [semithick, darkorange25512714, opacity=0.5, mark=square*, mark size=1.75, mark options={solid}, only marks, forget plot]
table {%
9 0.93048128342246
};
\addplot [semithick, forestgreen4416044, opacity=0.5, mark=triangle*, mark size=2.5, mark options={solid}, only marks, forget plot]
table {%
9 0.983957219251337
};
\addplot [semithick, black, opacity=0.5, mark=mystar, mark size=3, mark options={solid}, only marks, forget plot]
table {%
9 1
};
\addplot [semithick, steelblue31119180, opacity=0.5, mark=*, mark size=2, mark options={solid}, only marks, forget plot]
table {%
10 0.999676741554873
};
\addplot [semithick, darkorange25512714, opacity=0.5, mark=square*, mark size=1.75, mark options={solid}, only marks, forget plot]
table {%
10 1
};
\addplot [semithick, forestgreen4416044, opacity=0.5, mark=triangle*, mark size=2.5, mark options={solid}, only marks, forget plot]
table {%
10 0.999676741554873
};
\addplot [semithick, black, opacity=0.5, mark=mystar, mark size=3, mark options={solid}, only marks, forget plot]
table {%
10 0.999676741554873
};
\addplot [semithick, steelblue31119180, opacity=0.5, mark=*, mark size=2, mark options={solid}, only marks, forget plot]
table {%
11 1
};
\addplot [semithick, darkorange25512714, opacity=0.5, mark=square*, mark size=1.75, mark options={solid}, only marks, forget plot]
table {%
11 0.989304812834225
};
\addplot [semithick, forestgreen4416044, opacity=0.5, mark=triangle*, mark size=2.5, mark options={solid}, only marks, forget plot]
table {%
11 0.994652406417112
};
\addplot [semithick, black, opacity=0.5, mark=mystar, mark size=3, mark options={solid}, only marks, forget plot]
table {%
11 1
};
\addplot [semithick, steelblue31119180, opacity=0.5, mark=*, mark size=2, mark options={solid}, only marks, forget plot]
table {%
12 0.995555555555556
};
\addplot [semithick, darkorange25512714, opacity=0.5, mark=square*, mark size=1.75, mark options={solid}, only marks, forget plot]
table {%
12 1
};
\addplot [semithick, forestgreen4416044, opacity=0.5, mark=triangle*, mark size=2.5, mark options={solid}, only marks, forget plot]
table {%
12 0.995555555555556
};
\addplot [semithick, black, opacity=0.5, mark=mystar, mark size=3, mark options={solid}, only marks, forget plot]
table {%
12 0.991111111111111
};
\addplot [semithick, steelblue31119180, opacity=0.5, mark=*, mark size=2, mark options={solid}, only marks, forget plot]
table {%
13 1
};
\addplot [semithick, darkorange25512714, opacity=0.5, mark=square*, mark size=1.75, mark options={solid}, only marks, forget plot]
table {%
13 0.991031717098068
};
\addplot [semithick, forestgreen4416044, opacity=0.5, mark=triangle*, mark size=2.5, mark options={solid}, only marks, forget plot]
table {%
13 1
};
\addplot [semithick, black, opacity=0.5, mark=mystar, mark size=3, mark options={solid}, only marks, forget plot]
table {%
13 1
};
\addplot [semithick, steelblue31119180, opacity=0.5, mark=*, mark size=2, mark options={solid}, only marks, forget plot]
table {%
14 0.981931543453365
};
\addplot [semithick, darkorange25512714, opacity=0.5, mark=square*, mark size=1.75, mark options={solid}, only marks, forget plot]
table {%
14 0.982058786105102
};
\addplot [semithick, forestgreen4416044, opacity=0.5, mark=triangle*, mark size=2.5, mark options={solid}, only marks, forget plot]
table {%
14 1
};
\addplot [semithick, black, opacity=0.5, mark=mystar, mark size=3, mark options={solid}, only marks, forget plot]
table {%
14 0.996988590575561
};
\addplot [semithick, steelblue31119180, opacity=0.5, mark=*, mark size=2, mark options={solid}, only marks, forget plot]
table {%
15 0.989861920271572
};
\addplot [semithick, darkorange25512714, opacity=0.5, mark=square*, mark size=1.75, mark options={solid}, only marks, forget plot]
table {%
15 1
};
\addplot [semithick, forestgreen4416044, opacity=0.5, mark=triangle*, mark size=2.5, mark options={solid}, only marks, forget plot]
table {%
15 0.979677966879215
};
\addplot [semithick, black, opacity=0.5, mark=mystar, mark size=3, mark options={solid}, only marks, forget plot]
table {%
15 0.993164824074499
};
\addplot [semithick, steelblue31119180, opacity=0.5, mark=*, mark size=2, mark options={solid}, only marks, forget plot]
table {%
16 1
};
\addplot [semithick, darkorange25512714, opacity=0.5, mark=square*, mark size=1.75, mark options={solid}, only marks, forget plot]
table {%
16 0.980988593155894
};
\addplot [semithick, forestgreen4416044, opacity=0.5, mark=triangle*, mark size=2.5, mark options={solid}, only marks, forget plot]
table {%
16 0.984790874524715
};
\addplot [semithick, black, opacity=0.5, mark=mystar, mark size=3, mark options={solid}, only marks, forget plot]
table {%
16 0.977186311787072
};
\addplot [semithick, steelblue31119180, opacity=0.5, mark=*, mark size=2, mark options={solid}, only marks, forget plot]
table {%
17 0.978112402181646
};
\addplot [semithick, darkorange25512714, opacity=0.5, mark=square*, mark size=1.75, mark options={solid}, only marks, forget plot]
table {%
17 0.968294996442969
};
\addplot [semithick, forestgreen4416044, opacity=0.5, mark=triangle*, mark size=2.5, mark options={solid}, only marks, forget plot]
table {%
17 0.985439886175006
};
\addplot [semithick, black, opacity=0.5, mark=mystar, mark size=3, mark options={solid}, only marks, forget plot]
table {%
17 1
};
\addplot [semithick, steelblue31119180, opacity=0.5, mark=*, mark size=2, mark options={solid}, only marks, forget plot]
table {%
18 0.992815226011102
};
\addplot [semithick, darkorange25512714, opacity=0.5, mark=square*, mark size=1.75, mark options={solid}, only marks, forget plot]
table {%
18 0.994607454401269
};
\addplot [semithick, forestgreen4416044, opacity=0.5, mark=triangle*, mark size=2.5, mark options={solid}, only marks, forget plot]
table {%
18 1
};
\addplot [semithick, black, opacity=0.5, mark=mystar, mark size=3, mark options={solid}, only marks, forget plot]
table {%
18 0.998207771609834
};
\addplot [semithick, steelblue31119180, opacity=0.5, mark=*, mark size=2, mark options={solid}, only marks, forget plot]
table {%
19 0.999979497273138
};
\addplot [semithick, darkorange25512714, opacity=0.5, mark=square*, mark size=1.75, mark options={solid}, only marks, forget plot]
table {%
19 0.990404723828269
};
\addplot [semithick, forestgreen4416044, opacity=0.5, mark=triangle*, mark size=2.5, mark options={solid}, only marks, forget plot]
table {%
19 0.995222864640997
};
\addplot [semithick, black, opacity=0.5, mark=mystar, mark size=3, mark options={solid}, only marks, forget plot]
table {%
19 1
};
\addplot [semithick, steelblue31119180, opacity=0.5, mark=*, mark size=2, mark options={solid}, only marks, forget plot]
table {%
20 0.971584027765225
};
\addplot [semithick, darkorange25512714, opacity=0.5, mark=square*, mark size=1.75, mark options={solid}, only marks, forget plot]
table {%
20 0.943204208166881
};
\addplot [semithick, forestgreen4416044, opacity=0.5, mark=triangle*, mark size=2.5, mark options={solid}, only marks, forget plot]
table {%
20 0.975958496773377
};
\addplot [semithick, black, opacity=0.5, mark=mystar, mark size=3, mark options={solid}, only marks, forget plot]
table {%
20 1
};
\addplot [semithick, steelblue31119180, opacity=0.5, mark=*, mark size=2, mark options={solid}, only marks, forget plot]
table {%
21 1
};
\addplot [semithick, darkorange25512714, opacity=0.5, mark=square*, mark size=1.75, mark options={solid}, only marks, forget plot]
table {%
21 0.997863247863248
};
\addplot [semithick, forestgreen4416044, opacity=0.5, mark=triangle*, mark size=2.5, mark options={solid}, only marks, forget plot]
table {%
21 1
};
\addplot [semithick, black, opacity=0.5, mark=mystar, mark size=3, mark options={solid}, only marks, forget plot]
table {%
21 1
};
\addplot [semithick, steelblue31119180, opacity=0.5, mark=*, mark size=2, mark options={solid}, only marks, forget plot]
table {%
22 0.99496644295302
};
\addplot [semithick, darkorange25512714, opacity=0.5, mark=square*, mark size=1.75, mark options={solid}, only marks, forget plot]
table {%
22 1
};
\addplot [semithick, forestgreen4416044, opacity=0.5, mark=triangle*, mark size=2.5, mark options={solid}, only marks, forget plot]
table {%
22 1
};
\addplot [semithick, black, opacity=0.5, mark=mystar, mark size=3, mark options={solid}, only marks, forget plot]
table {%
22 0.998322147651007
};
\addplot [semithick, steelblue31119180, opacity=0.5, mark=*, mark size=2, mark options={solid}, only marks, forget plot]
table {%
23 1
};
\addplot [semithick, darkorange25512714, opacity=0.5, mark=square*, mark size=1.75, mark options={solid}, only marks, forget plot]
table {%
23 1
};
\addplot [semithick, forestgreen4416044, opacity=0.5, mark=triangle*, mark size=2.5, mark options={solid}, only marks, forget plot]
table {%
23 1
};
\addplot [semithick, black, opacity=0.5, mark=mystar, mark size=3, mark options={solid}, only marks, forget plot]
table {%
23 1
};
\addplot [semithick, steelblue31119180, opacity=0.5, mark=*, mark size=2, mark options={solid}, only marks, forget plot]
table {%
24 0.998292602415399
};
\addplot [semithick, darkorange25512714, opacity=0.5, mark=square*, mark size=1.75, mark options={solid}, only marks, forget plot]
table {%
24 0.998281219764834
};
\addplot [semithick, forestgreen4416044, opacity=0.5, mark=triangle*, mark size=2.5, mark options={solid}, only marks, forget plot]
table {%
24 1
};
\addplot [semithick, black, opacity=0.5, mark=mystar, mark size=3, mark options={solid}, only marks, forget plot]
table {%
24 1
};
\addplot [semithick, steelblue31119180, opacity=0.5, mark=*, mark size=2, mark options={solid}, only marks, forget plot]
table {%
25 0.99661317100166
};
\addplot [semithick, darkorange25512714, opacity=0.5, mark=square*, mark size=1.75, mark options={solid}, only marks, forget plot]
table {%
25 0.99830658550083
};
\addplot [semithick, forestgreen4416044, opacity=0.5, mark=triangle*, mark size=2.5, mark options={solid}, only marks, forget plot]
table {%
25 1
};
\addplot [semithick, black, opacity=0.5, mark=mystar, mark size=3, mark options={solid}, only marks, forget plot]
table {%
25 1
};
\addplot [semithick, steelblue31119180, opacity=0.5, mark=*, mark size=2, mark options={solid}, only marks, forget plot]
table {%
26 1
};
\addplot [semithick, darkorange25512714, opacity=0.5, mark=square*, mark size=1.75, mark options={solid}, only marks, forget plot]
table {%
26 0.99872954335597
};
\addplot [semithick, forestgreen4416044, opacity=0.5, mark=triangle*, mark size=2.5, mark options={solid}, only marks, forget plot]
table {%
26 0.998736092101558
};
\addplot [semithick, black, opacity=0.5, mark=mystar, mark size=3, mark options={solid}, only marks, forget plot]
table {%
26 0.998736092101558
};
\addplot [semithick, steelblue31119180, opacity=0.5, mark=*, mark size=2, mark options={solid}, only marks, forget plot]
table {%
27 1
};
\addplot [semithick, darkorange25512714, opacity=0.5, mark=square*, mark size=1.75, mark options={solid}, only marks, forget plot]
table {%
27 0.990909090909091
};
\addplot [semithick, forestgreen4416044, opacity=0.5, mark=triangle*, mark size=2.5, mark options={solid}, only marks, forget plot]
table {%
27 0.997402597402597
};
\addplot [semithick, black, opacity=0.5, mark=mystar, mark size=3, mark options={solid}, only marks, forget plot]
table {%
27 0.998701298701299
};
\addplot [semithick, steelblue31119180, forget plot]
table {%
-0.3 0.988261319921069
28.3 0.988261319921069
};
\addplot [semithick, darkorange25512714, forget plot]
table {%
-0.3 0.990044831440579
28.3 0.990044831440579
};
\addplot [semithick, forestgreen4416044, forget plot]
table {%
-0.3 0.99157343831666
28.3 0.99157343831666
};
\addplot [semithick, black, forget plot]
table {%
-0.3 0.994370908102317
28.3 0.994370908102317
};
\addplot [semithick, black, mark=mystar, mark size=3, mark options={solid}]
table {%
1 5
};
\addlegendentry{Ours (mean: 0.994, std.dev.: 0.016)}
\addplot [semithick, forestgreen4416044, mark=triangle*, mark size=2.5, mark options={solid}]
table {%
1 5
};
\addlegendentry{VI (mean: 0.992, std.dev.: 0.015)}
\addplot [semithick, steelblue31119180, mark=*, mark size=2, mark options={solid}]
table {%
1 5
};
\addlegendentry{LA (mean: 0.988, std.dev.: 0.025)}
\addplot [semithick, darkorange25512714, mark=square*, mark size=1.75, mark options={solid}]
table {%
1 5
};
\addlegendentry{EP (mean: 0.99, std.dev.: 0.017)}
\end{axis}

\end{tikzpicture}

%% file: table/t_test_lpd.tex
\begin{tabular}{l@{\hspace{-1em}}c|r@{}lr@{}lr@{}lr@{}l}
\toprule
{} &($n$, $d$)     &          \multicolumn{2}{c}{LA}         &          \multicolumn{2}{c}{EP}         &          \multicolumn{2}{c}{VI}         &         \multicolumn{2}{c}{Ours}        \\
\midrule
     {\sc neal}      &       (100, 1)       & $\bf  .317$ & ${\scriptstyle{\pm}.440}    $ & $\bf  .303$ & ${\scriptstyle{\pm}.432} $ & $\bf .295$ & ${\scriptstyle{\pm}.426} $ & $\bf  .301$ & ${\scriptstyle{\pm}.436}  $ \\
    {\sc boston}     &      (506, 13)       & $\bf -.210$ & ${\scriptstyle{\pm}.069}    $ & $\bf -.190$ & ${\scriptstyle{\pm}.053} $ & $   -.206$ & ${\scriptstyle{\pm}.056} $ & $\bf -.195$ & ${\scriptstyle{\pm}.061}  $ \\
    {\sc stock}      &      (1000, 1)       & $    1.910$ & ${\scriptstyle{\pm}.079}    $ & $    1.389$ & ${\scriptstyle{\pm}.242} $ & $   1.917$ & ${\scriptstyle{\pm}.082} $ & $\bf 1.921$ & ${\scriptstyle{\pm}.083}  $ \\
\midrule
    {Bold Count}     & {}& \multicolumn{2}{c}{$2$}            & \multicolumn{2}{c}{$2$}            & \multicolumn{2}{c}{$1$}            & \multicolumn{2}{c}{$3$}            \\
\bottomrule\end{tabular}\phantom{-}

%% file: table/isotropic_sparse_test_lpd.tex
\begin{tabular}{l|ccccc}
\toprule
{}                     &          LA         &          EP         &          VI         &         Ours        \\
\midrule
   {\sc titanic}       & $  -.217\scriptstyle{\pm}.037  $ & $  -.014\scriptstyle{\pm}.004  $ & $\bf -.011\scriptstyle{\pm}.003$ & $  -.037\scriptstyle{\pm}.005  $ \\
     {\sc bank}        & $\bf -.247\scriptstyle{\pm}.006$ & $\bf -.246\scriptstyle{\pm}.007$ & $  -.249\scriptstyle{\pm}.006  $ & $  -.247\scriptstyle{\pm}.007  $ \\
   {\sc twonorm}       & $\bf -.060\scriptstyle{\pm}.007$ & $\bf -.061\scriptstyle{\pm}.008$ & $\bf -.060\scriptstyle{\pm}.008$ & $  -.524\scriptstyle{\pm}.208  $ \\
   {\sc mushroom}      & $  -.129\scriptstyle{\pm}.003  $ & $  -.002\scriptstyle{\pm}.000  $ & $\bf -.001\scriptstyle{\pm}.000$ & $  -.028\scriptstyle{\pm}.001  $ \\
    {\sc magic}        & $\bf -.285\scriptstyle{\pm}.333$ & $  -.693\scriptstyle{\pm}.000  $ & $\bf -.008\scriptstyle{\pm}.001$ & $  -.070\scriptstyle{\pm}.002  $ \\
\midrule
    {Bold Count}       & $3$            & $2$            & $4$            & $0$            \\
\bottomrule\end{tabular}

%% file: fig/tikz/fig4b.tex
\begin{tikzpicture}

\definecolor{aliceblue242247253}{RGB}{242,247,253}
\definecolor{cornflowerblue107174214}{RGB}{107,174,214}
\definecolor{cornflowerblue89162207}{RGB}{89,162,207}
\definecolor{darkgray176}{RGB}{176,176,176}
\definecolor{lavender210227243}{RGB}{210,227,243}
\definecolor{lavender221234246}{RGB}{221,234,246}
\definecolor{lavender231240249}{RGB}{231,240,249}
\definecolor{lightblue168206228}{RGB}{168,206,228}
\definecolor{lightblue185213234}{RGB}{185,213,234}
\definecolor{midnightblue854116}{RGB}{8,54,116}
\definecolor{midnightblue868137}{RGB}{8,68,137}
\definecolor{orange}{RGB}{255,165,0}
\definecolor{powderblue200220239}{RGB}{200,220,239}
\definecolor{skyblue128185218}{RGB}{128,185,218}
\definecolor{skyblue149197223}{RGB}{149,197,223}
\definecolor{steelblue30109178}{RGB}{30,109,178}
\definecolor{steelblue43123186}{RGB}{43,123,186}
\definecolor{steelblue57137193}{RGB}{57,137,193}
\definecolor{steelblue72150200}{RGB}{72,150,200}
\definecolor{teal1996167}{RGB}{19,96,167}
\definecolor{teal882156}{RGB}{8,82,156}

\begin{axis}[
height=\figureheight,
tick align=outside,
tick pos=left,
width=\figurewidth,
x grid style={darkgray176},
xmin=-1, xmax=5,
xtick style={color=black},
y grid style={darkgray176},
ymin=-1, ymax=5,
ytick style={color=black}
]
\addplot [draw=none, fill=steelblue43123186]
table{%
x  y
-0.7 -1
-0.4 -1
-0.2131251495798 -1
-0.374361697534731 -0.7
-0.4 -0.642089437467692
-0.532353285330203 -0.4
-0.642132000541115 -0.1
-0.7 0.141001870680126
-0.720783490436941 0.2
-0.792723031609557 0.5
-0.834076739378187 0.8
-0.853747780679478 1.1
-0.848247130983168 1.4
-0.815745311737944 1.7
-0.772802827585009 2
-0.740213666659031 2.3
-0.719388403520897 2.6
-0.701001479182436 2.9
-0.7 2.92043641060813
-0.69223335178382 3.2
-0.68733433517888 3.5
-0.679671922960993 3.8
-0.672377236757823 4.1
-0.670492235893859 4.4
-0.671379203492741 4.7
-0.676238647923401 5
-0.7 5
-1 5
-1 4.7
-1 4.4
-1 4.1
-1 3.8
-1 3.5
-1 3.2
-1 2.9
-1 2.6
-1 2.3
-1 2
-1 1.7
-1 1.4
-1 1.1
-1 0.8
-1 0.5
-1 0.2
-1 -0.1
-1 -0.4
-1 -0.7
-1 -1
-0.7 -1
};
\addplot [draw=none, fill=steelblue43123186]
table{%
x  y
3.8 -1
4.1 -1
4.4 -1
4.7 -1
5 -1
5 -0.7
5 -0.4
5 -0.1
5 0.2
5 0.357741209332266
4.85611896313701 0.2
4.7 0.0291838090017719
4.58188404950009 -0.1
4.4 -0.298227711850577
4.30649667998318 -0.4
4.1 -0.623361373199528
4.02889977851421 -0.7
3.8 -0.943989697545643
3.74701026064194 -1
3.8 -1
};
\addplot [draw=none, fill=steelblue57137193]
table{%
x  y
-0.1 -1
0.14921294979712 -1
-0.0422558688659263 -0.7
-0.1 -0.589806041215127
-0.203837292561156 -0.4
-0.323850009274082 -0.1
-0.4 0.178066009668042
-0.407440993872572 0.2
-0.477748118870042 0.5
-0.519193860152935 0.8
-0.540553576312039 1.1
-0.542065487400838 1.4
-0.524716425963493 1.7
-0.49989030437207 2
-0.479014082110479 2.3
-0.46409559238273 2.6
-0.454338691831797 2.9
-0.447405076922711 3.2
-0.441273021077361 3.5
-0.439067853291693 3.8
-0.43751585149191 4.1
-0.43515821195295 4.4
-0.43429666040097 4.7
-0.434539367030573 5
-0.676238647923401 5
-0.671379203492741 4.7
-0.670492235893859 4.4
-0.672377236757823 4.1
-0.679671922960993 3.8
-0.68733433517888 3.5
-0.69223335178382 3.2
-0.7 2.92043641060813
-0.701001479182436 2.9
-0.719388403520897 2.6
-0.740213666659031 2.3
-0.772802827585009 2
-0.815745311737944 1.7
-0.848247130983168 1.4
-0.853747780679478 1.1
-0.834076739378187 0.8
-0.792723031609557 0.5
-0.720783490436941 0.2
-0.7 0.141001870680126
-0.642132000541115 -0.1
-0.532353285330203 -0.4
-0.4 -0.642089437467692
-0.374361697534731 -0.7
-0.2131251495798 -1
-0.1 -1
};
\addplot [draw=none, fill=steelblue57137193]
table{%
x  y
3.2 -1
3.5 -1
3.74701026064194 -1
3.8 -0.943989697545643
4.02889977851421 -0.7
4.1 -0.623361373199528
4.30649667998318 -0.4
4.4 -0.298227711850577
4.58188404950009 -0.1
4.7 0.0291838090017719
4.85611896313701 0.2
5 0.357741209332266
5 0.5
5 0.8
5 1.02245499195231
4.78603964521861 0.8
4.7 0.710528021613415
4.49717704924983 0.5
4.4 0.399066870057724
4.20753117098459 0.2
4.1 0.0885861419964858
3.91629961679435 -0.1
3.8 -0.21990458378481
3.6219608772982 -0.4
3.5 -0.524525325202092
3.32168441635676 -0.7
3.2 -0.821945694723621
3.01009326272378 -1
3.2 -1
};
\addplot [draw=none, fill=steelblue72150200]
table{%
x  y
0.2 -1
0.456474889183732 -1
0.229209314275941 -0.7
0.2 -0.655785478210618
0.0516014731145948 -0.4
-0.0798819245991073 -0.1
-0.1 -0.0393694421167288
-0.180389769758556 0.2
-0.246120344921826 0.5
-0.28643566154963 0.8
-0.308431329570621 1.1
-0.313698409658743 1.4
-0.304511225696703 1.7
-0.289169959517963 2
-0.2748532251918 2.3
-0.263557281532114 2.6
-0.255759006348289 2.9
-0.24934804876694 3.2
-0.243739738934206 3.5
-0.241102889434308 3.8
-0.238566173674233 4.1
-0.236246525432281 4.4
-0.235014657243673 4.7
-0.233451131900265 5
-0.4 5
-0.434539367030573 5
-0.43429666040097 4.7
-0.43515821195295 4.4
-0.43751585149191 4.1
-0.439067853291693 3.8
-0.441273021077361 3.5
-0.447405076922711 3.2
-0.454338691831797 2.9
-0.46409559238273 2.6
-0.479014082110479 2.3
-0.49989030437207 2
-0.524716425963493 1.7
-0.542065487400838 1.4
-0.540553576312039 1.1
-0.519193860152935 0.8
-0.477748118870042 0.5
-0.407440993872572 0.2
-0.4 0.178066009668042
-0.323850009274082 -0.1
-0.203837292561156 -0.4
-0.1 -0.589806041215127
-0.0422558688659263 -0.7
0.14921294979712 -1
0.2 -1
};
\addplot [draw=none, fill=steelblue72150200]
table{%
x  y
2.6 -1
2.9 -1
3.01009326272378 -1
3.2 -0.821945694723621
3.32168441635676 -0.7
3.5 -0.524525325202092
3.6219608772982 -0.4
3.8 -0.21990458378481
3.91629961679435 -0.1
4.1 0.0885861419964858
4.20753117098459 0.2
4.4 0.399066870057724
4.49717704924983 0.5
4.7 0.710528021613415
4.78603964521861 0.8
5 1.02245499195231
5 1.1
5 1.4
5 1.40344854084943
4.9966434796075 1.4
4.70485615114817 1.1
4.7 1.09499968380879
4.41338647874086 0.8
4.4 0.786233655668304
4.12107834069135 0.5
4.1 0.478382587394547
3.82684410047746 0.2
3.8 0.172623849731795
3.52851921943865 -0.1
3.5 -0.12876470645191
3.22200637237457 -0.4
3.2 -0.421744209279356
2.9 -0.699745472187175
2.89969913695984 -0.7
2.6 -0.96263216745372
2.54877857418829 -1
2.6 -1
};
\addplot [draw=none, fill=cornflowerblue89162207]
table{%
x  y
0.5 -1
0.8 -1
0.817052895288404 -1
0.8 -0.990035044923015
0.502211612749456 -0.7
0.5 -0.697570483233789
0.298391808611394 -0.4
0.2 -0.221446622955082
0.142524060354404 -0.1
0.0375301346413236 0.2
-0.0338344164626054 0.5
-0.0794010437006862 0.8
-0.1 1.03576766441231
-0.105474198509075 1.1
-0.115107562044004 1.4
-0.111839116315604 1.7
-0.101782906219364 2
-0.1 2.04568686778691
-0.0901021616241089 2.3
-0.0794902369639845 2.6
-0.0710532686433728 2.9
-0.0631494884917713 3.2
-0.0560929349862318 3.5
-0.0508221632840037 3.8
-0.0457868542585921 4.1
-0.0430233857371967 4.4
-0.0413889543647841 4.7
-0.0390810016509926 5
-0.1 5
-0.233451131900265 5
-0.235014657243673 4.7
-0.236246525432281 4.4
-0.238566173674233 4.1
-0.241102889434308 3.8
-0.243739738934206 3.5
-0.24934804876694 3.2
-0.255759006348289 2.9
-0.263557281532114 2.6
-0.2748532251918 2.3
-0.289169959517963 2
-0.304511225696703 1.7
-0.313698409658743 1.4
-0.308431329570621 1.1
-0.28643566154963 0.8
-0.246120344921826 0.5
-0.180389769758556 0.2
-0.1 -0.0393694421167288
-0.0798819245991074 -0.1
0.0516014731145948 -0.4
0.2 -0.655785478210618
0.229209314275941 -0.7
0.456474889183732 -1
0.5 -1
};
\addplot [draw=none, fill=cornflowerblue89162207]
table{%
x  y
2.3 -1
2.54877857418829 -1
2.6 -0.96263216745372
2.89969913695984 -0.7
2.9 -0.699745472187175
3.2 -0.421744209279356
3.22200637237457 -0.4
3.5 -0.12876470645191
3.52851921943865 -0.1
3.8 0.172623849731795
3.82684410047746 0.2
4.1 0.478382587394547
4.12107834069135 0.5
4.4 0.786233655668304
4.41338647874086 0.8
4.7 1.09499968380879
4.70485615114817 1.1
4.9966434796075 1.4
5 1.40344854084943
5 1.7
5 1.7058139151758
4.99434910214268 1.7
4.70192106399842 1.4
4.7 1.39803015767563
4.40991673134918 1.1
4.4 1.08984167967911
4.11747587546473 0.8
4.1 0.782143891542759
3.82356032315896 0.5
3.8 0.476067053399935
3.52594603516997 0.2
3.5 0.173964075354154
3.22027136866349 -0.1
3.2 -0.119867355924883
2.9 -0.398527260611007
2.8982776626421 -0.4
2.6 -0.657372363255754
2.54134586780119 -0.7
2.3 -0.882849156116791
2.09244297595668 -1
2.3 -1
};
\addplot [draw=none, fill=cornflowerblue107174214]
table{%
x  y
0.8 -0.990035044923015
0.817052895288404 -1
1.1 -1
1.4 -1
1.7 -1
2 -1
2.09244297595668 -1
2.3 -0.882849156116791
2.54134586780119 -0.7
2.6 -0.657372363255754
2.8982776626421 -0.4
2.9 -0.398527260611007
3.2 -0.119867355924883
3.22027136866349 -0.1
3.5 0.173964075354154
3.52594603516997 0.2
3.8 0.476067053399935
3.82356032315897 0.5
4.1 0.782143891542759
4.11747587546473 0.8
4.4 1.08984167967911
4.40991673134918 1.1
4.7 1.39803015767563
4.70192106399842 1.4
4.99434910214268 1.7
5 1.7058139151758
5 1.98744816829401
4.72061163790892 1.7
4.7 1.67886454250212
4.42943406173726 1.4
4.4 1.36982087785757
4.13846335713916 1.1
4.1 1.06062044502708
3.84682401193031 0.8
3.8 0.752292946505466
3.55243739398456 0.5
3.5 0.447213220998332
3.2511084728571 0.2
3.2 0.149839170537192
2.93477667546169 -0.1
2.9 -0.132548526644818
2.6 -0.389516603781494
2.58574078116561 -0.4
2.3 -0.611136007361735
2.13866247865963 -0.7
2 -0.780389784584124
1.7 -0.88205568060325
1.4 -0.904185160709034
1.1 -0.832574062254517
0.878675538366536 -0.7
0.8 -0.645624114939565
0.576143653713317 -0.4
0.5 -0.298879929332417
0.388989102270011 -0.1
0.261361029617962 0.2
0.2 0.403552331042577
0.176075583661144 0.5
0.128663683839089 0.8
0.101187979474902 1.1
0.0883056865172634 1.4
0.0861165640996856 1.7
0.0902321773112217 2
0.0967826630836959 2.3
0.104077974226875 2.6
0.111633654280845 2.9
0.120547342984626 3.2
0.131207229111902 3.5
0.141146592943242 3.8
0.148691125862407 4.1
0.151915427312074 4.4
0.153955648012927 4.7
0.155405103464982 5
-0.0390810016509926 5
-0.0413889543647841 4.7
-0.0430233857371967 4.4
-0.0457868542585921 4.1
-0.0508221632840037 3.8
-0.0560929349862318 3.5
-0.0631494884917713 3.2
-0.0710532686433728 2.9
-0.0794902369639845 2.6
-0.0901021616241089 2.3
-0.1 2.04568686778691
-0.101782906219364 2
-0.111839116315604 1.7
-0.115107562044004 1.4
-0.105474198509075 1.1
-0.1 1.03576766441231
-0.0794010437006862 0.8
-0.0338344164626054 0.5
0.0375301346413236 0.2
0.142524060354404 -0.1
0.2 -0.221446622955082
0.298391808611394 -0.4
0.5 -0.697570483233789
0.502211612749456 -0.7
0.8 -0.990035044923015
};
\addplot [draw=none, fill=skyblue128185218]
table{%
x  y
1.1 -0.832574062254517
1.4 -0.904185160709034
1.7 -0.88205568060325
2 -0.780389784584124
2.13866247865963 -0.7
2.3 -0.611136007361735
2.58574078116561 -0.4
2.6 -0.389516603781494
2.9 -0.132548526644818
2.93477667546169 -0.1
3.2 0.149839170537192
3.2511084728571 0.2
3.5 0.447213220998332
3.55243739398456 0.5
3.8 0.752292946505466
3.84682401193031 0.8
4.1 1.06062044502708
4.13846335713916 1.1
4.4 1.36982087785757
4.42943406173726 1.4
4.7 1.67886454250212
4.72061163790892 1.7
5 1.98744816829401
5 2
5 2.27428803897179
4.73479288945817 2
4.7 1.96418890917783
4.44567557484084 1.7
4.4 1.6529092170532
4.15760088381974 1.4
4.1 1.34056687373953
3.86982902759457 1.1
3.8 1.02812472533331
3.58038078895572 0.8
3.5 0.718102298292771
3.28524232453463 0.5
3.2 0.415319846085421
2.97683038026305 0.2
2.9 0.127484655461645
2.63987856594605 -0.1
2.6 -0.134334013643573
2.3 -0.354159604651435
2.21550115426416 -0.4
2 -0.517292371352996
1.7 -0.603502038774273
1.4 -0.598489016727852
1.1 -0.48858820774305
0.975023903588833 -0.4
0.8 -0.241308880746708
0.6934957570329 -0.1
0.522011594977388 0.2
0.5 0.255700839719711
0.427391910504762 0.5
0.369653662392539 0.8
0.335722794667917 1.1
0.319926058392869 1.4
0.318537680519897 1.7
0.327001153890548 2
0.339389353988932 2.3
0.3512856051116 2.6
0.364439569070721 2.9
0.380131452721714 3.2
0.397285385621457 3.5
0.410117254983404 3.8
0.421634284783554 4.1
0.431188095407789 4.4
0.440318078066944 4.7
0.453774327854724 5
0.2 5
0.155405103464982 5
0.153955648012927 4.7
0.151915427312074 4.4
0.148691125862407 4.1
0.141146592943242 3.8
0.131207229111902 3.5
0.120547342984626 3.2
0.111633654280845 2.9
0.104077974226876 2.6
0.0967826630836959 2.3
0.0902321773112217 2
0.0861165640996856 1.7
0.0883056865172634 1.4
0.101187979474902 1.1
0.128663683839089 0.8
0.176075583661144 0.5
0.2 0.403552331042577
0.261361029617962 0.2
0.388989102270011 -0.1
0.5 -0.298879929332417
0.576143653713317 -0.4
0.8 -0.645624114939565
0.878675538366535 -0.7
1.1 -0.832574062254517
};
\addplot [draw=none, fill=skyblue149197223]
table{%
x  y
1.1 -0.48858820774305
1.4 -0.598489016727852
1.7 -0.603502038774273
2 -0.517292371352996
2.21550115426416 -0.4
2.3 -0.354159604651435
2.6 -0.134334013643573
2.63987856594605 -0.1
2.9 0.127484655461645
2.97683038026305 0.2
3.2 0.415319846085421
3.28524232453463 0.5
3.5 0.718102298292771
3.58038078895572 0.8
3.8 1.02812472533331
3.86982902759457 1.1
4.1 1.34056687373953
4.15760088381974 1.4
4.4 1.6529092170532
4.44567557484084 1.7
4.7 1.96418890917783
4.73479288945817 2
5 2.27428803897179
5 2.3
5 2.58776938229018
4.72447898456866 2.3
4.7 2.27460436896428
4.43922752935958 2
4.4 1.9591177152911
4.15614799742576 1.7
4.1 1.64122393367861
3.87446952860893 1.4
3.8 1.32193189361293
3.59206185542403 1.1
3.5 1.00416004192426
3.30485181065185 0.8
3.2 0.693434407255316
3.00555550306319 0.5
2.9 0.398302483764445
2.67978295179937 0.2
2.6 0.130502697903715
2.3 -0.0942608001951389
2.28924919716765 -0.1
2 -0.248370337923538
1.7 -0.317300100681252
1.4 -0.269851272130307
1.11728140416602 -0.1
1.1 -0.0845304203919153
0.876725699780273 0.2
0.8 0.356985810031637
0.742822031347292 0.5
0.66635850842599 0.8
0.623116589459471 1.1
0.607612466489445 1.4
0.616898777731291 1.7
0.639660831209848 2
0.661730625942113 2.3
0.677838075920209 2.6
0.692613318443725 2.9
0.710270215526965 3.2
0.729944573482603 3.5
0.742309270027257 3.8
0.749200287775328 4.1
0.754352556441878 4.4
0.7830326313265 4.7
0.8 4.83347179439025
0.823062810396554 5
0.8 5
0.5 5
0.453774327854724 5
0.440318078066944 4.7
0.431188095407789 4.4
0.421634284783554 4.1
0.410117254983404 3.8
0.397285385621457 3.5
0.380131452721714 3.2
0.364439569070721 2.9
0.3512856051116 2.6
0.339389353988932 2.3
0.327001153890548 2
0.318537680519897 1.7
0.319926058392869 1.4
0.335722794667917 1.1
0.369653662392539 0.8
0.427391910504763 0.5
0.5 0.255700839719711
0.522011594977388 0.2
0.6934957570329 -0.1
0.8 -0.241308880746708
0.975023903588833 -0.4
1.1 -0.48858820774305
};
\addplot [draw=none, fill=lightblue168206228]
table{%
x  y
1.4 -0.269851272130307
1.7 -0.317300100681252
2 -0.248370337923538
2.28924919716765 -0.1
2.3 -0.0942608001951389
2.6 0.130502697903715
2.67978295179937 0.2
2.9 0.398302483764445
3.00555550306319 0.5
3.2 0.693434407255316
3.30485181065185 0.8
3.5 1.00416004192426
3.59206185542403 1.1
3.8 1.32193189361293
3.87446952860893 1.4
4.1 1.64122393367861
4.15614799742576 1.7
4.4 1.9591177152911
4.43922752935958 2
4.7 2.27460436896428
4.72447898456866 2.3
5 2.58776938229018
5 2.6
5 2.9
5 2.97570909039433
4.93088496012777 2.9
4.7 2.6510103575682
4.65447707457782 2.6
4.4 2.32112310006542
4.38159188901598 2.3
4.11043336119082 2
4.1 1.98881525269376
3.8391784293891 1.7
3.8 1.65771889826624
3.56754328920203 1.4
3.5 1.3273349816746
3.29092483717925 1.1
3.2 1.00438412565238
3.0019531419019 0.8
2.9 0.698633645580236
2.68647074339799 0.5
2.6 0.42295770531498
2.30947871683893 0.2
2.3 0.19307152578781
2 0.0487806170986589
1.7 0.00724592605697597
1.4 0.127629141975087
1.32073645596164 0.2
1.12107115776986 0.5
1.1 0.556201725852451
1.01990961010122 0.8
0.966746344590282 1.1
0.948170661685238 1.4
0.957663996686909 1.7
0.982864758514622 2
1.00926045374229 2.3
1.03015302905485 2.6
1.05059573826793 2.9
1.07456419348072 3.2
1.1 3.49340850134824
1.10037631019105 3.5
1.11668685643824 3.8
1.13630220137413 4.1
1.15349015893497 4.4
1.17056878225684 4.7
1.17115869379388 5
1.1 5
0.823062810396554 5
0.8 4.83347179439025
0.7830326313265 4.7
0.754352556441878 4.4
0.749200287775328 4.1
0.742309270027257 3.8
0.729944573482603 3.5
0.710270215526965 3.2
0.692613318443725 2.9
0.677838075920209 2.6
0.661730625942113 2.3
0.639660831209848 2
0.616898777731291 1.7
0.607612466489445 1.4
0.623116589459471 1.1
0.66635850842599 0.8
0.742822031347292 0.5
0.8 0.356985810031637
0.876725699780273 0.2
1.1 -0.0845304203919153
1.11728140416602 -0.1
1.4 -0.269851272130307
};
\addplot [draw=none, fill=lightblue185213234]
table{%
x  y
1.4 0.127629141975087
1.7 0.00724592605697597
2 0.0487806170986589
2.3 0.19307152578781
2.30947871683893 0.2
2.6 0.42295770531498
2.68647074339799 0.5
2.9 0.698633645580236
3.0019531419019 0.8
3.2 1.00438412565238
3.29092483717925 1.1
3.5 1.3273349816746
3.56754328920203 1.4
3.8 1.65771889826624
3.8391784293891 1.7
4.1 1.98881525269376
4.11043336119082 2
4.38159188901598 2.3
4.4 2.32112310006542
4.65447707457782 2.6
4.7 2.6510103575682
4.93088496012777 2.9
5 2.97570909039433
5 3.2
5 3.5
5 3.58374164644159
4.93631896912733 3.5
4.7 3.20612763944829
4.695577666161 3.2
4.44747387270571 2.9
4.4 2.84491766086997
4.20322353718279 2.6
4.1 2.47695456627379
3.96076892182572 2.3
3.8 2.10462850548389
3.71755985550126 2
3.5 1.735241697355
3.47133537077869 1.7
3.21450579361872 1.4
3.2 1.38398999346307
2.93685308666173 1.1
2.9 1.0617547273943
2.62820502817111 0.8
2.6 0.774094384961501
2.3 0.549087823035066
2.19523764584232 0.5
2 0.416942715305981
1.7 0.435025305590889
1.61426360884152 0.5
1.4 0.794795414623279
1.3977732262376 0.8
1.32446792469164 1.1
1.29131206730725 1.4
1.28407604558034 1.7
1.28973959913108 2
1.29901086214333 2.3
1.30846973322222 2.6
1.31950131822975 2.9
1.33367905207828 3.2
1.3465790812217 3.5
1.35780908751878 3.8
1.37171023192964 4.1
1.3861881887147 4.4
1.4 4.62258172443929
1.40947868231664 4.7
1.45523292643952 5
1.4 5
1.17115869379388 5
1.17056878225684 4.7
1.15349015893497 4.4
1.13630220137413 4.1
1.11668685643824 3.8
1.10037631019105 3.5
1.1 3.49340850134824
1.07456419348072 3.2
1.05059573826793 2.9
1.03015302905485 2.6
1.00926045374229 2.3
0.982864758514622 2
0.957663996686909 1.7
0.948170661685238 1.4
0.966746344590282 1.1
1.01990961010122 0.8
1.1 0.556201725852451
1.12107115776986 0.5
1.32073645596164 0.2
1.4 0.127629141975087
};
\addplot [draw=none, fill=powderblue200220239]
table{%
x  y
1.7 0.435025305590889
2 0.416942715305981
2.19523764584232 0.5
2.3 0.549087823035066
2.6 0.774094384961501
2.62820502817111 0.8
2.9 1.0617547273943
2.93685308666173 1.1
3.2 1.38398999346307
3.21450579361872 1.4
3.47133537077869 1.7
3.5 1.735241697355
3.71755985550126 2
3.8 2.10462850548389
3.96076892182572 2.3
4.1 2.47695456627379
4.20322353718279 2.6
4.4 2.84491766086997
4.44747387270571 2.9
4.695577666161 3.2
4.7 3.20612763944829
4.93631896912733 3.5
5 3.58374164644159
5 3.8
5 4.1
5 4.4
5 4.7
5 5
4.7 5
4.4 5
4.1 5
3.8 5
3.5 5
3.2 5
2.9 5
2.6 5
2.45379401326463 5
2.57296748907735 4.7
2.6 4.64374585302203
2.84157128486977 4.4
2.9 4.3493094522723
3.2 4.21366601529013
3.4731667874802 4.1
3.5 4.08892050299455
3.776834352014 3.8
3.8 3.74649796004713
3.86691013639564 3.5
3.83645992039469 3.2
3.8 3.10146950934142
3.72828706774679 2.9
3.57953257396281 2.6
3.5 2.46707765075143
3.39838584488603 2.3
3.20492455135955 2
3.2 1.99352822824163
2.96407790974077 1.7
2.9 1.62331169392794
2.68757400246442 1.4
2.6 1.31231484465472
2.3237492898204 1.1
2.3 1.08305922231807
2 1.03368244154459
1.913493433975 1.1
1.74123496250652 1.4
1.7 1.59057925092313
1.68830416211509 1.7
1.67379586276551 2
1.66676252577613 2.3
1.66543653327154 2.6
1.67171044404229 2.9
1.68948475292674 3.2
1.7 3.30531427177397
1.74281669520972 3.5
1.80496605821117 3.8
1.86496794910961 4.1
1.91854269729508 4.4
1.97777654040215 4.7
2 4.82064118267155
2.09372952295829 5
2 5
1.7 5
1.45523292643952 5
1.40947868231664 4.7
1.4 4.62258172443929
1.3861881887147 4.4
1.37171023192964 4.1
1.35780908751878 3.8
1.3465790812217 3.5
1.33367905207828 3.2
1.31950131822975 2.9
1.30846973322222 2.6
1.29901086214333 2.3
1.28973959913108 2
1.28407604558034 1.7
1.29131206730725 1.4
1.32446792469164 1.1
1.3977732262376 0.8
1.4 0.794795414623279
1.61426360884152 0.5
1.7 0.435025305590889
};
\addplot [draw=none, fill=lavender210227243]
table{%
x  y
2 1.03368244154459
2.3 1.08305922231807
2.3237492898204 1.1
2.6 1.31231484465472
2.68757400246442 1.4
2.9 1.62331169392794
2.96407790974077 1.7
3.2 1.99352822824163
3.20492455135955 2
3.39838584488603 2.3
3.5 2.46707765075143
3.57953257396281 2.6
3.72828706774679 2.9
3.8 3.10146950934142
3.83645992039469 3.2
3.86691013639564 3.5
3.8 3.74649796004713
3.776834352014 3.8
3.5 4.08892050299455
3.4731667874802 4.1
3.2 4.21366601529013
2.9 4.3493094522723
2.84157128486977 4.4
2.6 4.64374585302203
2.57296748907735 4.7
2.45379401326463 5
2.3 5
2.09372952295829 5
2 4.82064118267155
1.97777654040215 4.7
1.91854269729508 4.4
1.86496794910961 4.1
1.80496605821117 3.8
1.74281669520972 3.5
1.7 3.30531427177397
1.68948475292674 3.2
1.67171044404229 2.9
1.66543653327154 2.6
1.66676252577613 2.3
1.67379586276551 2
1.68830416211509 1.7
1.7 1.59057925092313
1.74123496250652 1.4
1.913493433975 1.1
2 1.03368244154459
};
\addplot [draw=orange, fill=orange, mark=*, mark size=0.5, only marks]
table{%
x  y
0 0
0.199321399857706 0.0990177562129225
0.392836786484295 0.196632837520439
0.571100343632043 0.289078050058533
0.732204037893958 0.375363226892275
0.883112513640284 0.45758609794214
1.03314139922304 0.538307731941908
1.18805899305216 0.619412070222198
1.34552563264213 0.701671048114184
1.49493695458398 0.784592924803357
1.62533536026021 0.867018365325768
1.73276592681216 0.947858283612239
1.81945266715868 1.0262850542359
1.88975443873721 1.10168447630292
1.94776646290679 1.17365915329915
1.99666077802346 1.24203723970478
2.03873161083419 1.3068360943812
2.07559409130972 1.36820030461176
2.10836940228763 1.42634240022382
2.13782863025159 1.48149837246727
2.16450077220112 1.53389880443166
2.18875325796516 1.58375266701465
2.21085078229981 1.63124045859395
2.23099638095289 1.67651375401544
2.24935813985391 1.71969863044491
2.26608495089619 1.7609008731673
2.28131463191163 1.80021141674329
2.29517727577525 1.83771108648201
2.30779599824998 1.87347424118108
2.31928653142922 1.9075712912882
2.32975651067017 1.94007026673753
2.33930487944324 1.97103767713306
2.34802157373599 2.00053889719626
2.35598750480863 2.02863826533098
2.36327479427808 2.05539902983624
2.36994719553061 2.08088322988429
2.37606063773494 2.10515156242673
2.38166383969905 2.12826326209301
2.38679895339012 2.15027600671081
2.39150220802244 2.17124585332633
2.39580453427468 2.19122720584585
2.39973215450863 2.21027281372107
2.403307129294 2.22843380030059
2.40654785363741 2.24575971901402
2.40946949852283 2.26229863526178
2.41208439500724 2.27809723175691
2.41440235935228 2.29320093516899
2.41643095857308 2.30765406230346
2.41817571634907 2.32149998473153
2.4196402594423 2.33478131175502
2.42082640457442 2.34754009282259
2.42173418509562 2.35981804199454
2.42236181570995 2.37165678879071
2.42270559195962 2.38309816179983
2.42275971905431 2.39418451387415
2.42251606185788 2.40495910074083
2.42196380425949 2.41546652866481
2.42108900153296 2.42575329173735
2.41987400330961 2.43586842591675
2.4182967170173 2.44586431580131
2.41632967147867 2.45579770224877
2.41393882702791 2.46573095581242
2.41108206096864 2.47573370466491
2.40770723417098 2.48588493940736
2.40374971461872 2.49627576573404
2.39912919552739 2.5070130466291
2.39374559983432 2.51822427941044
2.38747381733197 2.5300642041388
2.38015700974248 2.54272385257342
2.37159836818053 2.55644300764015
2.36155195371971 2.57152719967124
2.34971606509887 2.58836970729874
2.33574209967831 2.60747490657399
2.31930042355805 2.62946364929825
2.30030860608469 2.65499950431452
2.27945406232013 2.6845333784855
2.25870114028812 2.71796263673214
2.24054647450055 2.75480921927866
2.22620010382864 2.79484273369975
2.21538729330915 2.83795054882079
2.2074123801883 2.88378177886562
2.20163240790392 2.93190228931522
2.19756098379346 2.98194818696257
2.19487972135153 3.03365146433749
2.19340276935273 3.08682299743063
2.19303427180934 3.14133025398136
2.19372868074886 3.19709125325776
2.19546006198009 3.25408260187049
2.19819401332603 3.31235105534988
2.20185851507054 3.37201789806313
2.20633359033559 3.43325606959486
2.21146851875427 3.49623203753451
2.21711826104388 3.56101926293804
2.22316761230286 3.6275069440941
2.22950829979437 3.6953639776963
2.23594961378188 3.76412874361441
2.24215188732723 3.83335793921897
2.24771112425443 3.90269602416853
2.25220797333802 3.97188321858347
2.25516258901587 4.0408060922632
2.25607816468802 4.1094977853606
2.25451379195284 4.178057690365
2.2499987404264 4.24652299200003
2.24115742495076 4.31490523487502
2.22710616792986 4.38181156977091
2.21605394374986 4.44176458590117
2.20836065053122 4.4992847877365
2.20084432494708 4.55866279788571
2.18787940081856 4.62218417583658
2.17325217315865 4.68527754837357
2.17264956446199 4.72850545755756
2.1444256760392 4.80662962416131
2.14216724865959 4.84031898665242
2.13495255151871 4.89138530811059
2.12657908720899 4.94903890235097
};
\addplot [semithick, black, mark=*, mark size=1.5, mark options={solid}, only marks]
table {%
2.6 2.3
};\label{dot}
\addplot [line width = 1.5pt, red, mark=x, mark size=3, mark options={solid}, only marks]
table {%
2.40946949852283 2.26229863526178
};\label{cross}
\addplot [semithick, orange, mark=triangle*, mark size=1.5, mark options={solid,rotate=90}, only marks]
table {%
0 0
};\label{triangle}
\end{axis}

\end{tikzpicture}

%% file: fig/tikz/fig4a.tex
\begin{tikzpicture}

\definecolor{darkgray176}{RGB}{176,176,176}
\definecolor{darkorange25512714}{RGB}{255,127,14}
\definecolor{lightgray204}{RGB}{204,204,204}
\definecolor{steelblue31119180}{RGB}{31,119,180}

\begin{axis}[
height=\figureheight,
legend cell align={left},
legend style={fill opacity=0.8, draw opacity=1, text opacity=1, draw=lightgray204},
tick align=outside,
tick pos=left,
width=\figurewidth,
x grid style={darkgray176},
xmin=-14.95, xmax=313.95,
xtick style={color=black},
y grid style={darkgray176},
ymin=-110, ymax=-60,
ytick style={color=black}
]
\addplot [semithick, steelblue31119180]
table {%
0 -279.999926803682
1 -140.355251624262
2 -128.793169189164
3 -120.139418650593
4 -113.48413884233
5 -107.85886732214
6 -102.474400170755
7 -96.9379579467374
8 -91.4472154010779
9 -86.6452035734309
10 -82.9537157308557
11 -80.3043731888245
12 -78.4200156901442
13 -77.0485961048675
14 -76.0183113585817
15 -75.221714256364
16 -74.5919298222692
17 -74.0861075254673
18 -73.6757102306024
19 -73.3410129187642
20 -73.0679178272389
21 -72.8460261113114
22 -72.6674173792278
23 -72.5258551185815
24 -72.4162653014701
25 -72.3343966659301
26 -72.2766021179649
27 -72.2396992508443
28 -72.2208810338172
29 -72.2176573718481
30 -72.2278151621682
31 -72.2493892016126
32 -72.2806393880138
33 -72.3200315996217
34 -72.3662208035531
35 -72.4180356114375
36 -72.4744638650756
37 -72.5346390324428
38 -72.5978273078593
39 -72.6634153833637
40 -72.7308989097146
41 -72.7998717006859
42 -72.8700157547021
43 -72.9410921746216
44 -73.0129330621997
45 -73.0854344525273
46 -73.1585503401027
47 -73.2322878365114
48 -73.3067034935506
49 -73.3819008276282
50 -73.4580290932491
51 -73.5352833767937
52 -73.6139061181888
53 -73.6941902197125
54 -73.7764839715125
55 -73.8611981179066
56 -73.9488155153705
57 -74.0399040044849
58 -74.1351333521374
59 -74.2352974436715
60 -74.34134335723
61 -74.4544095942026
62 -74.5758766614033
63 -74.7074345427323
64 -74.8511735786556
65 -75.0097082312574
66 -75.1863476725223
67 -75.3853338369645
68 -75.6121773180277
69 -75.8741338671004
70 -76.1808719705891
71 -76.5453476052094
72 -76.9846735862723
73 -77.5198323384535
74 -78.1702870341447
75 -78.9359408770759
76 -79.7720058036624
77 -80.606883982032
78 -81.4086136781499
79 -82.1880522827052
80 -82.9583232413419
81 -83.7273262868805
82 -84.4998364789585
83 -85.2779787610073
84 -86.0618972969894
85 -86.8504219520375
86 -87.6420042255211
87 -88.4356530232751
88 -89.2321158986768
89 -90.0354592794687
90 -90.8533425361177
91 -91.6947019510702
92 -92.5651782147058
93 -93.4630937837201
94 -94.3796242348351
95 -95.3061016160012
96 -96.2424494141864
97 -97.1958591733176
98 -98.1787206017321
99 -99.2123197414914
100 -100.325421767439
101 -101.547153726674
102 -102.908987807475
103 -104.503800558283
104 -106.318623058036
105 -107.495834494286
106 -108.423610145327
107 -109.551776926458
108 -111.33255041026
109 -113.147108736332
110 -112.840329598914
111 -117.003140379851
112 -115.89437879956
113 -116.886960178376
114 -118.316456690202
115 -119.765705899619
116 -121.858181392913
117 -123.808143840918
118 -124.860059462679
119 -125.521497471206
120 -126.411721657319
121 -129.363741630223
122 -128.249856842628
123 -125.574958114014
124 -135.009338319331
125 -130.57309637446
126 -130.588613426629
127 -137.074235724185
128 -133.199924068693
129 -133.396907165886
130 -137.543715746616
131 -136.160405505628
132 -136.029801449296
133 -142.163807295184
134 -140.455505914523
135 -142.374543890056
136 -139.770455966896
137 -141.832730664438
138 -143.095193010667
139 -144.84798005949
140 -142.661633396343
141 -147.331818140821
142 -144.213621603912
143 -149.137792782995
144 -146.027399418669
145 -150.549117448609
146 -146.744241842023
147 -151.363607274676
148 -148.166284299149
149 -154.897121462729
150 -152.658826044886
151 -150.196770010889
152 -169.178208487346
153 -153.937496477788
154 -151.974719738888
155 -165.332904599055
156 -157.999462505003
157 -158.858908438895
158 -157.697597540462
159 -159.188274509644
160 -158.93363449847
161 -161.620291883707
162 -158.463123584047
163 -162.305251284461
164 -163.169729821903
165 -165.728679250466
166 -162.888198324981
167 -165.065779193532
168 -162.941006467938
169 -162.345637824376
170 -198.569872326378
171 -177.824871848199
172 -175.497957418065
173 -167.534146602649
174 -178.698428106357
175 -168.074946463622
176 -205.245635895334
177 -187.842652059697
178 -182.439615437071
179 -182.541303087072
180 -193.935931276491
181 -176.896454486388
182 -181.075369169052
183 -191.795335131679
184 -181.188066416751
185 -197.596340345355
186 -184.792311098176
187 -190.61131391854
188 -199.485809867415
189 -187.305080387993
190 -199.290204809714
191 -206.812154905778
192 -197.817372953189
193 -201.310211987014
194 -186.076838467929
195 -190.637180157003
196 -202.605641576147
197 -212.957484445422
198 -210.967735272736
199 -204.501038347321
200 -207.258892602881
201 -206.607094189967
202 -193.061322206103
203 -185.235131349063
204 -183.794853165743
205 -188.223323138826
206 -192.748972795962
207 -184.290700636143
208 -184.431439815987
209 -183.988986119844
210 -185.27491868437
211 -182.587082417562
212 -174.103675366417
213 -171.8779097007
214 -174.453877662567
215 -174.36189435537
216 -248.549567359847
217 -190.728717524824
218 -192.418631759578
219 -196.416409606441
220 -184.594229040031
221 -183.096093494125
222 -232.122831409674
223 -233.918182480092
224 -218.623391741262
225 -205.761747429856
226 -196.330810607004
227 -215.521843902524
228 -214.71951524499
229 -213.691928614344
230 -213.366793565529
231 -216.439082849335
232 -218.473929954877
233 -211.916591964393
234 -214.778733507308
235 -217.38215886824
236 -242.765089132205
237 -235.029207155634
238 -229.50983031374
239 -231.607045426264
240 -236.448681062778
241 -241.047447089221
242 -232.501077336765
243 -226.83145882949
244 -227.587315946741
245 -215.062531221585
246 -209.124603126688
247 -206.965061676408
248 -205.88011335525
249 -205.37145580333
250 -203.758886338216
251 -205.017807618577
252 -206.458696243585
253 -211.409501442357
254 -216.010644651183
255 -219.555170447947
256 -225.431070597812
257 -217.758342640916
258 -222.572247245165
259 -227.154657943918
260 -225.216850539858
261 -221.773199788141
262 -222.794162521127
263 -229.228240824162
264 -225.838381670083
265 -219.650394051869
266 -216.371154571896
267 -220.151691550994
268 -224.361929435966
269 -230.500323557369
270 -235.288559595941
271 -237.36851154081
272 -241.505885614414
273 -235.643309722283
274 -222.491912477998
275 -211.701285458719
276 -210.637868883428
277 -209.123299777303
278 -203.405671340856
279 -204.485352772071
280 -206.925383040301
281 -210.861005400632
282 -214.335268974471
283 -218.784152748572
284 -222.972294034649
285 -226.028637718038
286 -218.49354301137
287 -220.727188133418
288 -225.621982672585
289 -228.192313574987
290 -232.427503333771
291 -237.375098461602
292 -238.298316646947
293 -243.067426514503
294 -242.227191001423
295 -224.466651793191
296 -218.47502784345
297 -216.820628343218
298 -222.162938389623
299 -227.450254955598
};
\addlegendentry{$\mathcal{L}_\text{VI}$}
\addplot [semithick, darkorange25512714]
table {%
0 -194.081210556846
1 -139.319532395384
2 -127.936544401454
3 -119.157689975829
4 -112.357660365371
5 -106.560947613453
6 -100.983473325397
7 -95.2503426335486
8 -89.5932308410655
9 -84.6941443029622
10 -80.9722979063196
11 -78.3128614571282
12 -76.4021893997631
13 -74.9793200549607
14 -73.8771053775546
15 -72.9949147750388
16 -72.2711593099358
17 -71.6664543240041
18 -71.1544050968355
19 -70.7165827722618
20 -70.3396659114774
21 -70.0137198390328
22 -69.7311096283624
23 -69.4857924532822
24 -69.2728508100839
25 -69.0881834647609
26 -68.9283003861333
27 -68.7901858299163
28 -68.6712056399901
29 -68.5690429331581
30 -68.4816518177704
31 -68.4072224851815
32 -68.3441534836531
33 -68.2910286222628
34 -68.2465970109047
35 -68.2097553969849
36 -68.1795323386912
37 -68.1550739565462
38 -68.1356311003904
39 -68.1205478084577
40 -68.1092509495979
41 -68.101240946374
42 -68.0960834832372
43 -68.0934021121278
44 -68.0928716770366
45 -68.0942124877358
46 -68.0971851809013
47 -68.101586212824
48 -68.1072439334258
49 -68.1140151959396
50 -68.1217824613601
51 -68.1304513618491
52 -68.1399486929385
53 -68.1502208106283
54 -68.1612324163578
55 -68.1729657201361
56 -68.185419979967
57 -68.1986114234507
58 -68.2125735649683
59 -68.2273579378837
60 -68.2430352638378
61 -68.2596970770735
62 -68.2774578036762
63 -68.2964572514487
64 -68.3168633734971
65 -68.338874985808
66 -68.3627237730966
67 -68.3886742777159
68 -68.4170194205078
69 -68.448067174664
70 -68.4821113117369
71 -68.519377881359
72 -68.5599516888572
73 -68.603758258848
74 -68.6508689627956
75 -68.7022868313565
76 -68.7592792336705
77 -68.8194063204353
78 -68.87745888278
79 -68.9316192422189
80 -68.9832415261556
81 -69.0339218081913
82 -69.0847716160586
83 -69.1364308341895
84 -69.189162931985
85 -69.2430568795552
86 -69.2982261387233
87 -69.3548626645587
88 -69.4131103039253
89 -69.4729442917599
90 -69.5343566840457
91 -69.597844317728
92 -69.664196221613
93 -69.7334969892101
94 -69.8055098979824
95 -69.8807342662738
96 -69.9603436188932
97 -70.0458152219965
98 -70.1383830654326
99 -70.2398498645042
100 -70.3529923741678
101 -70.4808420882928
102 -70.625147397671
103 -70.7886049976137
104 -70.9661808329203
105 -71.1262395791962
106 -71.2559025837926
107 -71.3721027821032
108 -71.4928535046061
109 -71.766817404391
110 -72.0290813120641
111 -72.3202149110808
112 -72.6028911927129
113 -72.7336799101392
114 -72.948019196268
115 -73.1870499733291
116 -73.2991456453201
117 -73.6328086437703
118 -73.942510761533
119 -74.0480905000612
120 -74.1688930253419
121 -74.7008013422224
122 -74.63374994839
123 -75.6304628515865
124 -75.0272728688749
125 -74.9609822663296
126 -75.085883568962
127 -75.7539947130344
128 -75.3302949294799
129 -75.2685919078455
130 -75.4652929684697
131 -75.5370133484487
132 -75.5019574267794
133 -75.9375240493295
134 -76.2396368813195
135 -76.0484939995168
136 -76.2186682259698
137 -76.0178042741599
138 -76.2500930663728
139 -76.462646329123
140 -76.6131755223123
141 -76.4717480153787
142 -76.9568746685948
143 -76.7849269074109
144 -77.3373322915653
145 -77.1223206007452
146 -77.5287322189396
147 -77.2618843716296
148 -77.7069293638293
149 -77.5479426757975
150 -78.3724394330534
151 -78.2674160888944
152 -79.2581126476698
153 -79.653077070754
154 -78.9372211087088
155 -79.3392080843004
156 -79.6866727865905
157 -79.1388138292415
158 -79.2954047356435
159 -79.2171493822525
160 -79.3390348070631
161 -79.2588950720274
162 -79.8904361207841
163 -79.3490820084911
164 -79.9198345489783
165 -80.1054404784038
166 -80.2624297063929
167 -80.090421280628
168 -80.5015376988636
169 -80.4405564695537
170 -81.1544891365263
171 -83.4595079021668
172 -82.6658066632156
173 -82.2526436183761
174 -81.0992081970732
175 -82.1728167500669
176 -81.5880583738045
177 -83.7097832945048
178 -83.7317324768617
179 -83.1393485474899
180 -83.2788425961903
181 -84.3255018161294
182 -82.6382408745221
183 -82.7615777722333
184 -83.5619532950023
185 -83.2584748935957
186 -84.6283289561009
187 -83.9351182367053
188 -84.5637662725441
189 -85.2444618374984
190 -84.3101692292086
191 -86.2018088069071
192 -86.3871967910003
193 -86.3966991741323
194 -89.0249003654885
195 -85.4877345244172
196 -84.2245572285626
197 -86.0505429566991
198 -87.9837806254634
199 -87.6954614006098
200 -87.481274844752
201 -87.7754086257226
202 -90.1169134489836
203 -87.4295303408761
204 -86.6334271540346
205 -86.8351068289719
206 -86.5466810137789
207 -88.3727123760223
208 -86.8166726009657
209 -87.0735196716174
210 -87.4667140917019
211 -87.6887475511654
212 -91.6843057574821
213 -90.0985562678602
214 -85.7151464534379
215 -86.7486257067173
216 -83.1283029802971
217 -86.4973109714844
218 -84.9075095600505
219 -85.4458912177799
220 -88.5240502929289
221 -87.5723655983431
222 -84.071523562817
223 -87.4220407765105
224 -89.8328985528688
225 -89.0841977297004
226 -88.0631175757841
227 -86.5803252524794
228 -88.739908515805
229 -89.0206882633723
230 -88.8923036845054
231 -89.4490023904463
232 -90.5134789282763
233 -89.7986862567604
234 -90.6464589491908
235 -90.3157518480241
236 -91.787410248499
237 -93.4753908313316
238 -93.596550658815
239 -92.9925083328564
240 -94.2676024534726
241 -96.4305731962193
242 -95.4134645910803
243 -95.479506210822
244 -94.615628131587
245 -95.604872931809
246 -93.7903602595188
247 -93.3164004435594
248 -92.0749881366562
249 -91.4601904440867
250 -92.5510218474186
251 -90.9977887444426
252 -91.4535670696492
253 -92.5671292412541
254 -90.7794308839509
255 -91.2105879401923
256 -91.0613363661053
257 -92.3711109335901
258 -92.0564858014936
259 -91.9875312288237
260 -93.1729913278579
261 -92.6579043083966
262 -92.4226377028693
263 -93.2187291244596
264 -93.558473614107
265 -94.8649578391874
266 -94.0590323866402
267 -93.8588136555896
268 -91.6982091499547
269 -92.6440829260574
270 -93.2669895088863
271 -95.0538468896752
272 -94.6394701378151
273 -95.740358466019
274 -96.0208059091055
275 -98.054136456446
276 -93.6733335850049
277 -93.9482583360923
278 -96.2832187994052
279 -95.6571686301224
280 -92.6019597522837
281 -92.3148331017123
282 -92.948565809467
283 -92.0458686848834
284 -91.3453983156202
285 -92.5323942320395
286 -93.2900641831576
287 -92.735839480625
288 -92.5753389393151
289 -92.6773142097358
290 -92.3366051838193
291 -92.9605863280412
292 -93.8492428479817
293 -93.7956193310326
294 -94.5605622814238
295 -95.7766733778295
296 -95.1687424576511
297 -94.5069961183101
298 -92.9471261019834
299 -94.1305401725053
};
\addlegendentry{$\mathcal{L}_\text{EP}$}
\addplot [semithick, red, densely dashed, forget plot]
table {%
44 -110
44 -60
};
\draw (axis cs:49,-65) node[
  scale=0.5,
  anchor=base west,
  text=black,
  rotate=0.0
]{When $\mathcal{L}_\text{EP}$ starts to decrease};
\end{axis}

\end{tikzpicture}

%% file: table/seed_test_acc.tex
\begin{tabular}{lc|cccc|c}
\toprule
{} &($n$, $d$)     &          LA         &          EP         &          VI         &         Ours        &         MCMC        \\
\midrule
    {\sc trains}     &       (10, 30)       & $  0.620{\pm}0.394   $ & $\bf 0.680{\pm}0.371 $ & $  0.660{\pm}0.353   $ & $\bf 0.710{\pm}0.348 $ & $  0.730{\pm}0.349   $ \\
   {\sc balloons}    &       (16, 5)        & $\bf 0.618{\pm}0.275 $ & $  0.607{\pm}0.282   $ & $\bf 0.615{\pm}0.285 $ & $\bf 0.650{\pm}0.291 $ & $  0.625{\pm}0.272   $ \\
  {\sc fertility}    &      (100, 10)       & $\bf 0.879{\pm}0.062 $ & $\bf 0.880{\pm}0.062 $ & $\bf 0.879{\pm}0.062 $ & $\bf 0.877{\pm}0.063 $ & $  0.877{\pm}0.061   $ \\
{\sc pittsburg-bridges-T-OR-D} &       (102, 8)       & $\bf 0.874{\pm}0.070 $ & $\bf 0.868{\pm}0.075 $ & $\bf 0.875{\pm}0.072 $ & $\bf 0.877{\pm}0.068 $ & $  0.868{\pm}0.069   $ \\
{\sc acute-nephritis} &       (120, 7)       & $\bf 1.000{\pm}0.000 $ & $\bf 1.000{\pm}0.000 $ & $\bf 1.000{\pm}0.000 $ & $\bf 1.000{\pm}0.000 $ & $  1.000{\pm}0.000   $ \\
{\sc acute-inflammation} &       (120, 7)       & $\bf 1.000{\pm}0.000 $ & $\bf 1.000{\pm}0.000 $ & $\bf 1.000{\pm}0.000 $ & $\bf 1.000{\pm}0.000 $ & $  1.000{\pm}0.000   $ \\
{\sc echocardiogram} &      (131, 11)       & $  0.820{\pm}0.064   $ & $\bf 0.840{\pm}0.056 $ & $  0.808{\pm}0.069   $ & $  0.808{\pm}0.068   $ & $  0.797{\pm}0.073   $ \\
  {\sc hepatitis}    &      (155, 20)       & $\bf 0.830{\pm}0.059 $ & $  0.819{\pm}0.061   $ & $\bf 0.833{\pm}0.059 $ & $\bf 0.832{\pm}0.060 $ & $  0.830{\pm}0.057   $ \\
  {\sc parkinsons}   &      (195, 23)       & $\bf 0.951{\pm}0.031 $ & $  0.888{\pm}0.050   $ & $  0.942{\pm}0.032   $ & $\bf 0.949{\pm}0.034 $ & $  0.950{\pm}0.031   $ \\
{\sc breast-cancer-wisc-prog} &      (198, 34)       & $  0.808{\pm}0.058   $ & $  0.793{\pm}0.070   $ & $\bf 0.815{\pm}0.057 $ & $\bf 0.814{\pm}0.057 $ & $  0.808{\pm}0.060   $ \\
    {\sc spect}      &      (265, 23)       & $\bf 0.706{\pm}0.055 $ & $\bf 0.703{\pm}0.057 $ & $\bf 0.703{\pm}0.056 $ & $\bf 0.705{\pm}0.056 $ & $  0.707{\pm}0.058   $ \\
{\sc statlog-heart}  &      (270, 14)       & $  0.830{\pm}0.049   $ & $\bf 0.838{\pm}0.047 $ & $  0.830{\pm}0.049   $ & $  0.828{\pm}0.050   $ & $  0.822{\pm}0.050   $ \\
{\sc haberman-survival} &       (306, 4)       & $\bf 0.741{\pm}0.048 $ & $\bf 0.740{\pm}0.050 $ & $\bf 0.741{\pm}0.049 $ & $\bf 0.741{\pm}0.048 $ & $  0.746{\pm}0.047   $ \\
  {\sc ionosphere}   &      (351, 34)       & $  0.927{\pm}0.029   $ & $  0.930{\pm}0.035   $ & $\bf 0.942{\pm}0.030 $ & $\bf 0.945{\pm}0.027 $ & $  0.943{\pm}0.026   $ \\
 {\sc horse-colic}   &      (368, 26)       & $  0.810{\pm}0.042   $ & $\bf 0.817{\pm}0.040 $ & $  0.805{\pm}0.044   $ & $  0.806{\pm}0.043   $ & $  0.805{\pm}0.045   $ \\
{\sc congressional-voting} &      (435, 17)       & $\bf 0.599{\pm}0.046 $ & $\bf 0.597{\pm}0.046 $ & $\bf 0.598{\pm}0.047 $ & $\bf 0.597{\pm}0.049 $ & $  0.600{\pm}0.046   $ \\
{\sc cylinder-bands} &      (512, 36)       & $  0.778{\pm}0.041   $ & $  0.763{\pm}0.045   $ & $  0.782{\pm}0.041   $ & $\bf 0.794{\pm}0.040 $ & $  0.795{\pm}0.040   $ \\
{\sc breast-cancer-wisc-diag} &      (569, 31)       & $\bf 0.969{\pm}0.015 $ & $\bf 0.974{\pm}0.015 $ & $\bf 0.970{\pm}0.015 $ & $\bf 0.972{\pm}0.014 $ & $  0.973{\pm}0.016   $ \\
{\sc ilpd-indian-liver} &      (583, 10)       & $\bf 0.718{\pm}0.035 $ & $\bf 0.715{\pm}0.039 $ & $\bf 0.719{\pm}0.035 $ & $\bf 0.719{\pm}0.035 $ & $  0.719{\pm}0.036   $ \\
   {\sc monks-2}     &       (601, 7)       & $  0.758{\pm}0.036   $ & $  0.738{\pm}0.034   $ & $  0.760{\pm}0.035   $ & $\bf 0.773{\pm}0.035 $ & $  0.778{\pm}0.034   $ \\
{\sc statlog-australian-credit} &      (690, 15)       & $\bf 0.677{\pm}0.035 $ & $  0.667{\pm}0.032   $ & $\bf 0.677{\pm}0.035 $ & $\bf 0.677{\pm}0.035 $ & $  0.678{\pm}0.035   $ \\
{\sc credit-approval} &      (690, 16)       & $\bf 0.859{\pm}0.031 $ & $  0.856{\pm}0.029   $ & $\bf 0.860{\pm}0.030 $ & $\bf 0.860{\pm}0.030 $ & $  0.859{\pm}0.030   $ \\
{\sc breast-cancer-wisc} &      (699, 10)       & $  0.967{\pm}0.013   $ & $\bf 0.969{\pm}0.012 $ & $  0.967{\pm}0.013   $ & $  0.967{\pm}0.013   $ & $  0.967{\pm}0.013   $ \\
    {\sc blood}      &       (748, 5)       & $\bf 0.783{\pm}0.033 $ & $\bf 0.784{\pm}0.032 $ & $\bf 0.783{\pm}0.034 $ & $\bf 0.783{\pm}0.034 $ & $  0.779{\pm}0.031   $ \\
     {\sc pima}      &       (768, 9)       & $\bf 0.764{\pm}0.030 $ & $\bf 0.765{\pm}0.031 $ & $\bf 0.764{\pm}0.030 $ & $\bf 0.764{\pm}0.030 $ & $  0.763{\pm}0.030   $ \\
 {\sc mammographic}  &       (961, 6)       & $\bf 0.823{\pm}0.026 $ & $\bf 0.823{\pm}0.026 $ & $\bf 0.823{\pm}0.026 $ & $\bf 0.823{\pm}0.026 $ & $  0.823{\pm}0.026   $ \\
{\sc statlog-german-credit} &      (1000, 25)      & $\bf 0.769{\pm}0.027 $ & $  0.766{\pm}0.028   $ & $\bf 0.768{\pm}0.027 $ & $\bf 0.768{\pm}0.027 $ & $  0.767{\pm}0.027   $ \\
\midrule
    {Bold Count}     & {}& $18$           & $17$           & $19$           & $23$           & $/$            \\
\bottomrule\end{tabular}

%% file: table/ard_test_lpd.tex
\begin{tabular}{lc|cccc}
\toprule
{} &($n$, $d$)     &          LA         &          EP         &          VI         &         Ours        \\
\midrule
  {\sc fertility}    &      (100, 10)       & $\bf -0.417{\pm}0.096$ & $\bf -0.407{\pm}0.120$ & $\bf -0.439{\pm}0.116$ & $  -0.456{\pm}0.115  $ \\
{\sc pittsburg-bridges-T-OR-D} &       (102, 8)       & $\bf -0.320{\pm}0.057$ & $\bf -0.325{\pm}0.085$ & $\bf -0.325{\pm}0.055$ & $\bf -0.320{\pm}0.061$ \\
{\sc acute-inflammation} &       (120, 7)       & $  -0.112{\pm}0.005  $ & $  -0.029{\pm}0.004  $ & $\bf -1.668{\pm}2.754$ & $\bf -0.005{\pm}0.001$ \\
  {\sc parkinsons}   &      (195, 23)       & $  -0.260{\pm}0.029  $ & $  -0.214{\pm}0.086  $ & $\bf -0.062{\pm}0.062$ & $\bf -0.048{\pm}0.042$ \\
{\sc breast-cancer-wisc-prog} &      (198, 34)       & $\bf -0.498{\pm}0.053$ & $\bf -0.476{\pm}0.068$ & $\bf -0.516{\pm}0.080$ & $\bf -0.644{\pm}0.092$ \\
    {\sc spect}      &      (265, 23)       & $\bf -0.608{\pm}0.059$ & $\bf -0.614{\pm}0.064$ & $\bf -0.643{\pm}0.056$ & $  -0.646{\pm}0.068  $ \\
{\sc statlog-heart}  &      (270, 14)       & $  -0.433{\pm}0.066  $ & $\bf -0.398{\pm}0.056$ & $  -0.450{\pm}0.072  $ & $  -0.472{\pm}0.082  $ \\
{\sc haberman-survival} &       (306, 4)       & $\bf -0.535{\pm}0.049$ & $\bf -0.531{\pm}0.053$ & $\bf -0.539{\pm}0.050$ & $\bf -0.540{\pm}0.051$ \\
  {\sc ionosphere}   &      (351, 34)       & $\bf -0.243{\pm}0.058$ & $\bf -0.253{\pm}0.032$ & $\bf -0.208{\pm}0.086$ & $\bf -0.231{\pm}0.061$ \\
{\sc congressional-voting} &      (435, 17)       & $\bf -0.642{\pm}0.030$ & $\bf -0.639{\pm}0.033$ & $  -0.687{\pm}0.043  $ & $\bf -0.644{\pm}0.081$ \\
{\sc cylinder-bands} &      (512, 36)       & $\bf -0.468{\pm}0.063$ & $  -0.501{\pm}0.061  $ & $\bf -0.454{\pm}0.068$ & $  -0.478{\pm}0.064  $ \\
{\sc breast-cancer-wisc-diag} &      (569, 31)       & $\bf -0.098{\pm}0.024$ & $  -0.122{\pm}0.031  $ & $\bf -0.075{\pm}0.044$ & $\bf -0.073{\pm}0.046$ \\
{\sc ilpd-indian-liver} &      (583, 10)       & $  -0.531{\pm}0.030  $ & $\bf -0.518{\pm}0.035$ & $\bf -0.519{\pm}0.036$ & $\bf -0.518{\pm}0.037$ \\
   {\sc monks-2}     &       (601, 7)       & $  -0.458{\pm}0.029  $ & $  -0.485{\pm}0.038  $ & $  -0.424{\pm}0.037  $ & $\bf -0.399{\pm}0.036$ \\
{\sc statlog-australian-credit} &      (690, 15)       & $\bf -0.633{\pm}0.022$ & $\bf -0.627{\pm}0.009$ & $\bf -0.633{\pm}0.022$ & $\bf -0.632{\pm}0.023$ \\
{\sc credit-approval} &      (690, 16)       & $  -0.294{\pm}0.102  $ & $  -0.148{\pm}0.015  $ & $\bf -0.905{\pm}1.126$ & $\bf -0.070{\pm}0.028$ \\
{\sc breast-cancer-wisc} &      (699, 10)       & $\bf -0.102{\pm}0.026$ & $\bf -0.101{\pm}0.028$ & $\bf -0.102{\pm}0.034$ & $\bf -0.104{\pm}0.035$ \\
    {\sc blood}      &       (748, 5)       & $\bf -0.477{\pm}0.047$ & $\bf -0.477{\pm}0.049$ & $\bf -0.477{\pm}0.048$ & $\bf -0.477{\pm}0.048$ \\
     {\sc pima}      &       (768, 9)       & $\bf -0.473{\pm}0.034$ & $\bf -0.466{\pm}0.042$ & $\bf -0.474{\pm}0.037$ & $\bf -0.474{\pm}0.036$ \\
 {\sc mammographic}  &       (961, 6)       & $\bf -0.399{\pm}0.039$ & $\bf -0.401{\pm}0.041$ & $\bf -0.400{\pm}0.040$ & $\bf -0.400{\pm}0.040$ \\
{\sc statlog-german-credit} &      (1000, 25)      & $\bf -0.488{\pm}0.044$ & $\bf -0.485{\pm}0.046$ & $  -0.501{\pm}0.053  $ & $\bf -0.503{\pm}0.054$ \\
\midrule
    {Bold Count}     & {}& $15$           & $15$           & $17$           & $17$           \\
\bottomrule\end{tabular}

%% file: table/ard_test_acc.tex
\begin{tabular}{lc|cccc}
\toprule
{} &($n$, $d$)     &          LA         &          EP         &          VI         &         Ours        \\
\midrule
  {\sc fertility}    &      (100, 10)       & $\bf 0.840{\pm}0.049 $ & $\bf 0.880{\pm}0.060 $ & $\bf 0.840{\pm}0.049 $ & $\bf 0.840{\pm}0.049 $ \\
{\sc pittsburg-bridges-T-OR-D} &       (102, 8)       & $\bf 0.853{\pm}0.003 $ & $\bf 0.863{\pm}0.036 $ & $\bf 0.863{\pm}0.019 $ & $\bf 0.853{\pm}0.003 $ \\
{\sc acute-inflammation} &       (120, 7)       & $\bf 1.000{\pm}0.000 $ & $\bf 1.000{\pm}0.000 $ & $\bf 0.833{\pm}0.173 $ & $\bf 1.000{\pm}0.000 $ \\
  {\sc parkinsons}   &      (195, 23)       & $\bf 0.985{\pm}0.021 $ & $\bf 0.928{\pm}0.071 $ & $\bf 0.990{\pm}0.021 $ & $\bf 0.990{\pm}0.021 $ \\
{\sc breast-cancer-wisc-prog} &      (198, 34)       & $\bf 0.768{\pm}0.029 $ & $\bf 0.768{\pm}0.048 $ & $\bf 0.773{\pm}0.053 $ & $\bf 0.753{\pm}0.049 $ \\
    {\sc spect}      &      (265, 23)       & $\bf 0.702{\pm}0.056 $ & $\bf 0.709{\pm}0.054 $ & $\bf 0.709{\pm}0.057 $ & $\bf 0.675{\pm}0.066 $ \\
{\sc statlog-heart}  &      (270, 14)       & $  0.796{\pm}0.039   $ & $\bf 0.837{\pm}0.027 $ & $  0.789{\pm}0.049   $ & $\bf 0.793{\pm}0.054 $ \\
{\sc haberman-survival} &       (306, 4)       & $\bf 0.732{\pm}0.043 $ & $\bf 0.738{\pm}0.038 $ & $\bf 0.735{\pm}0.044 $ & $\bf 0.735{\pm}0.044 $ \\
  {\sc ionosphere}   &      (351, 34)       & $\bf 0.923{\pm}0.026 $ & $\bf 0.909{\pm}0.026 $ & $\bf 0.932{\pm}0.016 $ & $  0.917{\pm}0.016   $ \\
{\sc congressional-voting} &      (435, 17)       & $\bf 0.607{\pm}0.027 $ & $\bf 0.598{\pm}0.030 $ & $\bf 0.536{\pm}0.091 $ & $\bf 0.607{\pm}0.034 $ \\
{\sc cylinder-bands} &      (512, 36)       & $\bf 0.791{\pm}0.050 $ & $  0.768{\pm}0.043   $ & $\bf 0.785{\pm}0.078 $ & $\bf 0.799{\pm}0.048 $ \\
{\sc breast-cancer-wisc-diag} &      (569, 31)       & $\bf 0.975{\pm}0.010 $ & $\bf 0.963{\pm}0.022 $ & $\bf 0.972{\pm}0.013 $ & $\bf 0.974{\pm}0.013 $ \\
{\sc ilpd-indian-liver} &      (583, 10)       & $\bf 0.707{\pm}0.033 $ & $\bf 0.724{\pm}0.033 $ & $  0.702{\pm}0.023   $ & $\bf 0.705{\pm}0.030 $ \\
   {\sc monks-2}     &       (601, 7)       & $  0.765{\pm}0.040   $ & $  0.727{\pm}0.044   $ & $  0.769{\pm}0.038   $ & $\bf 0.785{\pm}0.031 $ \\
{\sc statlog-australian-credit} &      (690, 15)       & $\bf 0.651{\pm}0.032 $ & $\bf 0.677{\pm}0.018 $ & $\bf 0.659{\pm}0.035 $ & $\bf 0.649{\pm}0.033 $ \\
{\sc credit-approval} &      (690, 16)       & $\bf 0.962{\pm}0.020 $ & $  0.961{\pm}0.013   $ & $\bf 0.859{\pm}0.201 $ & $\bf 0.980{\pm}0.011 $ \\
{\sc breast-cancer-wisc} &      (699, 10)       & $\bf 0.961{\pm}0.011 $ & $\bf 0.958{\pm}0.011 $ & $\bf 0.964{\pm}0.010 $ & $\bf 0.963{\pm}0.012 $ \\
    {\sc blood}      &       (748, 5)       & $\bf 0.786{\pm}0.038 $ & $\bf 0.789{\pm}0.036 $ & $\bf 0.789{\pm}0.036 $ & $\bf 0.787{\pm}0.037 $ \\
     {\sc pima}      &       (768, 9)       & $\bf 0.766{\pm}0.025 $ & $\bf 0.771{\pm}0.028 $ & $\bf 0.766{\pm}0.025 $ & $\bf 0.767{\pm}0.023 $ \\
 {\sc mammographic}  &       (961, 6)       & $\bf 0.831{\pm}0.012 $ & $\bf 0.829{\pm}0.016 $ & $\bf 0.830{\pm}0.013 $ & $\bf 0.830{\pm}0.013 $ \\
{\sc statlog-german-credit} &      (1000, 25)      & $\bf 0.787{\pm}0.028 $ & $\bf 0.779{\pm}0.027 $ & $\bf 0.782{\pm}0.028 $ & $\bf 0.778{\pm}0.028 $ \\
\midrule
    {Bold Count}     & {}& $19$           & $18$           & $18$           & $20$           \\
\bottomrule\end{tabular}

%% file: fig/tikz/classification_relative_lpd_acc.tex
\begin{tikzpicture}

\definecolor{darkgray176}{RGB}{176,176,176}
\definecolor{darkorange25512714}{RGB}{255,127,14}
\definecolor{forestgreen4416044}{RGB}{44,160,44}
\definecolor{lightgray204}{RGB}{204,204,204}
\definecolor{steelblue31119180}{RGB}{31,119,180}

\begin{axis}[
height=\figureheight,
legend cell align={left},
legend style={
  fill opacity=0.8,
  draw opacity=1,
  text opacity=1,
  at={(0.97,0.03)},
  anchor=south east,
  draw=lightgray204
},
tick align=outside,
tick pos=left,
width=\figurewidth,
x grid style={darkgray176},
xlabel={Data set \#},
xmin=0, xmax=22,
xtick style={color=black},
y grid style={darkgray176},
ylabel={Relative accuracy},
ymin=0.82, ymax=1,
ytick style={color=black}
]
\addplot [semithick, steelblue31119180, opacity=0.5, mark=*, mark size=2, mark options={solid}, only marks, forget plot]
table {%
1 0.954545454545454
};
\addplot [semithick, darkorange25512714, opacity=0.5, mark=square*, mark size=1.75, mark options={solid}, only marks, forget plot]
table {%
1 1
};
\addplot [semithick, forestgreen4416044, opacity=0.5, mark=triangle*, mark size=2.5, mark options={solid}, only marks, forget plot]
table {%
1 0.954545454545454
};
\addplot [semithick, black, opacity=0.5, mark=mystar, mark size=3, mark options={solid}, only marks, forget plot]
table {%
1 0.954545454545454
};
\addplot [semithick, steelblue31119180, opacity=0.5, mark=*, mark size=2, mark options={solid}, only marks, forget plot]
table {%
2 0.98841059602649
};
\addplot [semithick, darkorange25512714, opacity=0.5, mark=square*, mark size=1.75, mark options={solid}, only marks, forget plot]
table {%
2 1
};
\addplot [semithick, forestgreen4416044, opacity=0.5, mark=triangle*, mark size=2.5, mark options={solid}, only marks, forget plot]
table {%
2 1
};
\addplot [semithick, black, opacity=0.5, mark=mystar, mark size=3, mark options={solid}, only marks, forget plot]
table {%
2 0.98841059602649
};
\addplot [semithick, steelblue31119180, opacity=0.5, mark=*, mark size=2, mark options={solid}, only marks, forget plot]
table {%
3 1
};
\addplot [semithick, darkorange25512714, opacity=0.5, mark=square*, mark size=1.75, mark options={solid}, only marks, forget plot]
table {%
3 1
};
\addplot [semithick, forestgreen4416044, opacity=0.5, mark=triangle*, mark size=2.5, mark options={solid}, only marks, forget plot]
table {%
3 0.833333333333333
};
\addplot [semithick, black, opacity=0.5, mark=mystar, mark size=3, mark options={solid}, only marks, forget plot]
table {%
3 1
};
\addplot [semithick, steelblue31119180, opacity=0.5, mark=*, mark size=2, mark options={solid}, only marks, forget plot]
table {%
4 0.994818652849741
};
\addplot [semithick, darkorange25512714, opacity=0.5, mark=square*, mark size=1.75, mark options={solid}, only marks, forget plot]
table {%
4 0.937823834196891
};
\addplot [semithick, forestgreen4416044, opacity=0.5, mark=triangle*, mark size=2.5, mark options={solid}, only marks, forget plot]
table {%
4 1
};
\addplot [semithick, black, opacity=0.5, mark=mystar, mark size=3, mark options={solid}, only marks, forget plot]
table {%
4 1
};
\addplot [semithick, steelblue31119180, opacity=0.5, mark=*, mark size=2, mark options={solid}, only marks, forget plot]
table {%
5 0.993364299933643
};
\addplot [semithick, darkorange25512714, opacity=0.5, mark=square*, mark size=1.75, mark options={solid}, only marks, forget plot]
table {%
5 0.993696084936961
};
\addplot [semithick, forestgreen4416044, opacity=0.5, mark=triangle*, mark size=2.5, mark options={solid}, only marks, forget plot]
table {%
5 1
};
\addplot [semithick, black, opacity=0.5, mark=mystar, mark size=3, mark options={solid}, only marks, forget plot]
table {%
5 0.97378898473789
};
\addplot [semithick, steelblue31119180, opacity=0.5, mark=*, mark size=2, mark options={solid}, only marks, forget plot]
table {%
6 0.989361702127659
};
\addplot [semithick, darkorange25512714, opacity=0.5, mark=square*, mark size=1.75, mark options={solid}, only marks, forget plot]
table {%
6 1
};
\addplot [semithick, forestgreen4416044, opacity=0.5, mark=triangle*, mark size=2.5, mark options={solid}, only marks, forget plot]
table {%
6 1
};
\addplot [semithick, black, opacity=0.5, mark=mystar, mark size=3, mark options={solid}, only marks, forget plot]
table {%
6 0.952127659574468
};
\addplot [semithick, steelblue31119180, opacity=0.5, mark=*, mark size=2, mark options={solid}, only marks, forget plot]
table {%
7 0.951327433628319
};
\addplot [semithick, darkorange25512714, opacity=0.5, mark=square*, mark size=1.75, mark options={solid}, only marks, forget plot]
table {%
7 1
};
\addplot [semithick, forestgreen4416044, opacity=0.5, mark=triangle*, mark size=2.5, mark options={solid}, only marks, forget plot]
table {%
7 0.942477876106195
};
\addplot [semithick, black, opacity=0.5, mark=mystar, mark size=3, mark options={solid}, only marks, forget plot]
table {%
7 0.946902654867257
};
\addplot [semithick, steelblue31119180, opacity=0.5, mark=*, mark size=2, mark options={solid}, only marks, forget plot]
table {%
8 0.99112002291607
};
\addplot [semithick, darkorange25512714, opacity=0.5, mark=square*, mark size=1.75, mark options={solid}, only marks, forget plot]
table {%
8 1
};
\addplot [semithick, forestgreen4416044, opacity=0.5, mark=triangle*, mark size=2.5, mark options={solid}, only marks, forget plot]
table {%
8 0.995560011458035
};
\addplot [semithick, black, opacity=0.5, mark=mystar, mark size=3, mark options={solid}, only marks, forget plot]
table {%
8 0.995560011458035
};
\addplot [semithick, steelblue31119180, opacity=0.5, mark=*, mark size=2, mark options={solid}, only marks, forget plot]
table {%
9 0.990843123704216
};
\addplot [semithick, darkorange25512714, opacity=0.5, mark=square*, mark size=1.75, mark options={solid}, only marks, forget plot]
table {%
9 0.975639253628196
};
\addplot [semithick, forestgreen4416044, opacity=0.5, mark=triangle*, mark size=2.5, mark options={solid}, only marks, forget plot]
table {%
9 1
};
\addplot [semithick, black, opacity=0.5, mark=mystar, mark size=3, mark options={solid}, only marks, forget plot]
table {%
9 0.984709744298549
};
\addplot [semithick, steelblue31119180, opacity=0.5, mark=*, mark size=2, mark options={solid}, only marks, forget plot]
table {%
10 1
};
\addplot [semithick, darkorange25512714, opacity=0.5, mark=square*, mark size=1.75, mark options={solid}, only marks, forget plot]
table {%
10 0.984848484848485
};
\addplot [semithick, forestgreen4416044, opacity=0.5, mark=triangle*, mark size=2.5, mark options={solid}, only marks, forget plot]
table {%
10 0.882575757575757
};
\addplot [semithick, black, opacity=0.5, mark=mystar, mark size=3, mark options={solid}, only marks, forget plot]
table {%
10 1
};
\addplot [semithick, steelblue31119180, opacity=0.5, mark=*, mark size=2, mark options={solid}, only marks, forget plot]
table {%
11 0.990111277908833
};
\addplot [semithick, darkorange25512714, opacity=0.5, mark=square*, mark size=1.75, mark options={solid}, only marks, forget plot]
table {%
11 0.960826363571378
};
\addplot [semithick, forestgreen4416044, opacity=0.5, mark=triangle*, mark size=2.5, mark options={solid}, only marks, forget plot]
table {%
11 0.982891319369981
};
\addplot [semithick, black, opacity=0.5, mark=mystar, mark size=3, mark options={solid}, only marks, forget plot]
table {%
11 1
};
\addplot [semithick, steelblue31119180, opacity=0.5, mark=*, mark size=2, mark options={solid}, only marks, forget plot]
table {%
12 1
};
\addplot [semithick, darkorange25512714, opacity=0.5, mark=square*, mark size=1.75, mark options={solid}, only marks, forget plot]
table {%
12 0.987329682923723
};
\addplot [semithick, forestgreen4416044, opacity=0.5, mark=triangle*, mark size=2.5, mark options={solid}, only marks, forget plot]
table {%
12 0.996386731185534
};
\addplot [semithick, black, opacity=0.5, mark=mystar, mark size=3, mark options={solid}, only marks, forget plot]
table {%
12 0.998185406850885
};
\addplot [semithick, steelblue31119180, opacity=0.5, mark=*, mark size=2, mark options={solid}, only marks, forget plot]
table {%
13 0.976405683807662
};
\addplot [semithick, darkorange25512714, opacity=0.5, mark=square*, mark size=1.75, mark options={solid}, only marks, forget plot]
table {%
13 1
};
\addplot [semithick, forestgreen4416044, opacity=0.5, mark=triangle*, mark size=2.5, mark options={solid}, only marks, forget plot]
table {%
13 0.969199136842962
};
\addplot [semithick, black, opacity=0.5, mark=mystar, mark size=3, mark options={solid}, only marks, forget plot]
table {%
13 0.974023858963397
};
\addplot [semithick, steelblue31119180, opacity=0.5, mark=*, mark size=2, mark options={solid}, only marks, forget plot]
table {%
14 0.974568096115057
};
\addplot [semithick, darkorange25512714, opacity=0.5, mark=square*, mark size=1.75, mark options={solid}, only marks, forget plot]
table {%
14 0.925791458388143
};
\addplot [semithick, forestgreen4416044, opacity=0.5, mark=triangle*, mark size=2.5, mark options={solid}, only marks, forget plot]
table {%
14 0.978795053933176
};
\addplot [semithick, black, opacity=0.5, mark=mystar, mark size=3, mark options={solid}, only marks, forget plot]
table {%
14 1
};
\addplot [semithick, steelblue31119180, opacity=0.5, mark=*, mark size=2, mark options={solid}, only marks, forget plot]
table {%
15 0.961456102783726
};
\addplot [semithick, darkorange25512714, opacity=0.5, mark=square*, mark size=1.75, mark options={solid}, only marks, forget plot]
table {%
15 1
};
\addplot [semithick, forestgreen4416044, opacity=0.5, mark=triangle*, mark size=2.5, mark options={solid}, only marks, forget plot]
table {%
15 0.974304068522484
};
\addplot [semithick, black, opacity=0.5, mark=mystar, mark size=3, mark options={solid}, only marks, forget plot]
table {%
15 0.9593147751606
};
\addplot [semithick, steelblue31119180, opacity=0.5, mark=*, mark size=2, mark options={solid}, only marks, forget plot]
table {%
16 0.982248520710059
};
\addplot [semithick, darkorange25512714, opacity=0.5, mark=square*, mark size=1.75, mark options={solid}, only marks, forget plot]
table {%
16 0.980769230769231
};
\addplot [semithick, forestgreen4416044, opacity=0.5, mark=triangle*, mark size=2.5, mark options={solid}, only marks, forget plot]
table {%
16 0.877218934911243
};
\addplot [semithick, black, opacity=0.5, mark=mystar, mark size=3, mark options={solid}, only marks, forget plot]
table {%
16 1
};
\addplot [semithick, steelblue31119180, opacity=0.5, mark=*, mark size=2, mark options={solid}, only marks, forget plot]
table {%
17 0.997036910712953
};
\addplot [semithick, darkorange25512714, opacity=0.5, mark=square*, mark size=1.75, mark options={solid}, only marks, forget plot]
table {%
17 0.994041845642234
};
\addplot [semithick, forestgreen4416044, opacity=0.5, mark=triangle*, mark size=2.5, mark options={solid}, only marks, forget plot]
table {%
17 1
};
\addplot [semithick, black, opacity=0.5, mark=mystar, mark size=3, mark options={solid}, only marks, forget plot]
table {%
17 0.998518455356477
};
\addplot [semithick, steelblue31119180, opacity=0.5, mark=*, mark size=2, mark options={solid}, only marks, forget plot]
table {%
18 0.996619437102245
};
\addplot [semithick, darkorange25512714, opacity=0.5, mark=square*, mark size=1.75, mark options={solid}, only marks, forget plot]
table {%
18 1
};
\addplot [semithick, forestgreen4416044, opacity=0.5, mark=triangle*, mark size=2.5, mark options={solid}, only marks, forget plot]
table {%
18 1
};
\addplot [semithick, black, opacity=0.5, mark=mystar, mark size=3, mark options={solid}, only marks, forget plot]
table {%
18 0.998309718551122
};
\addplot [semithick, steelblue31119180, opacity=0.5, mark=*, mark size=2, mark options={solid}, only marks, forget plot]
table {%
19 0.993282382221439
};
\addplot [semithick, darkorange25512714, opacity=0.5, mark=square*, mark size=1.75, mark options={solid}, only marks, forget plot]
table {%
19 1
};
\addplot [semithick, forestgreen4416044, opacity=0.5, mark=triangle*, mark size=2.5, mark options={solid}, only marks, forget plot]
table {%
19 0.993282382221439
};
\addplot [semithick, black, opacity=0.5, mark=mystar, mark size=3, mark options={solid}, only marks, forget plot]
table {%
19 0.99496729291016
};
\addplot [semithick, steelblue31119180, opacity=0.5, mark=*, mark size=2, mark options={solid}, only marks, forget plot]
table {%
20 1
};
\addplot [semithick, darkorange25512714, opacity=0.5, mark=square*, mark size=1.75, mark options={solid}, only marks, forget plot]
table {%
20 0.997507270461155
};
\addplot [semithick, forestgreen4416044, opacity=0.5, mark=triangle*, mark size=2.5, mark options={solid}, only marks, forget plot]
table {%
20 0.998753635230578
};
\addplot [semithick, black, opacity=0.5, mark=mystar, mark size=3, mark options={solid}, only marks, forget plot]
table {%
20 0.998753635230578
};
\addplot [semithick, steelblue31119180, opacity=0.5, mark=*, mark size=2, mark options={solid}, only marks, forget plot]
table {%
21 1
};
\addplot [semithick, darkorange25512714, opacity=0.5, mark=square*, mark size=1.75, mark options={solid}, only marks, forget plot]
table {%
21 0.989834815756036
};
\addplot [semithick, forestgreen4416044, opacity=0.5, mark=triangle*, mark size=2.5, mark options={solid}, only marks, forget plot]
table {%
21 0.993646759847522
};
\addplot [semithick, black, opacity=0.5, mark=mystar, mark size=3, mark options={solid}, only marks, forget plot]
table {%
21 0.98856416772554
};
\addplot [semithick, steelblue31119180, forget plot]
table {%
4.44089209850063e-16 0.986929509385408
22 0.986929509385408
};
\addplot [semithick, darkorange25512714, forget plot]
table {%
4.44089209850063e-16 0.987052777386783
22 0.987052777386783
};
\addplot [semithick, forestgreen4416044, forget plot]
table {%
4.44089209850063e-16 0.970141450242081
22 0.970141450242081
};
\addplot [semithick, black, forget plot]
table {%
4.44089209850063e-16 0.986032496012234
22 0.986032496012234
};
\addplot [semithick, black, mark=mystar, mark size=3, mark options={solid}]
table {%
1 5
};
\addlegendentry{Ours (mean: 0.986, std.dev.: 0.018)}
\addplot [semithick, forestgreen4416044, mark=triangle*, mark size=2.5, mark options={solid}]
table {%
1 5
};
\addlegendentry{VI (mean: 0.97, std.dev.: 0.047)}
\addplot [semithick, steelblue31119180, mark=*, mark size=2, mark options={solid}]
table {%
1 5
};
\addlegendentry{LA (mean: 0.987, std.dev.: 0.015)}
\addplot [semithick, darkorange25512714, mark=square*, mark size=1.75, mark options={solid}]
table {%
1 5
};
\addlegendentry{EP (mean: 0.987, std.dev.: 0.021)}
\end{axis}

\end{tikzpicture}

%% file: table/ep_test_lpd.tex
\begin{tabular}{lc|cccc}
\toprule
{} &($n$, $d$)     &          LA         &    EP (new init.)   &          VI         &         Ours        \\
\midrule
    {\sc trains}     &       (10, 30)       & $\bf -0.695{\pm}0.011$ & $\bf -0.680{\pm}0.042$ & $\bf -0.692{\pm}0.023$ & $\bf -0.681{\pm}0.042$ \\
   {\sc balloons}    &       (16, 5)        & $\bf -0.707{\pm}0.146$ & $\bf -0.672{\pm}0.263$ & $\bf -0.711{\pm}0.239$ & $\bf -0.657{\pm}0.267$ \\
  {\sc fertility}    &      (100, 10)       & $\bf -0.379{\pm}0.099$ & $\bf -0.379{\pm}0.103$ & $\bf -0.378{\pm}0.103$ & $\bf -0.379{\pm}0.103$ \\
{\sc pittsburg-bridges-T-OR-D} &       (102, 8)       & $\bf -0.306{\pm}0.044$ & $\bf -0.295{\pm}0.058$ & $\bf -0.295{\pm}0.057$ & $\bf -0.296{\pm}0.059$ \\
{\sc acute-nephritis} &       (120, 7)       & $  -0.203{\pm}0.010  $ & $  -0.010{\pm}0.003  $ & $  -0.006{\pm}0.002  $ & $\bf -0.005{\pm}0.001$ \\
{\sc acute-inflammation} &       (120, 7)       & $  -0.172{\pm}0.033  $ & $  -0.013{\pm}0.002  $ & $  -0.008{\pm}0.001  $ & $\bf -0.007{\pm}0.001$ \\
{\sc echocardiogram} &      (131, 11)       & $\bf -0.420{\pm}0.073$ & $\bf -0.421{\pm}0.086$ & $\bf -0.420{\pm}0.085$ & $\bf -0.421{\pm}0.086$ \\
  {\sc hepatitis}    &      (155, 20)       & $\bf -0.369{\pm}0.043$ & $\bf -0.362{\pm}0.052$ & $\bf -0.361{\pm}0.051$ & $\bf -0.361{\pm}0.052$ \\
  {\sc parkinsons}   &      (195, 23)       & $  -0.267{\pm}0.030  $ & $  -0.153{\pm}0.050  $ & $  -0.164{\pm}0.062  $ & $\bf -0.145{\pm}0.054$ \\
{\sc breast-cancer-wisc-prog} &      (198, 34)       & $\bf -0.460{\pm}0.063$ & $\bf -0.458{\pm}0.070$ & $\bf -0.458{\pm}0.071$ & $\bf -0.458{\pm}0.070$ \\
    {\sc spect}      &      (265, 23)       & $\bf -0.587{\pm}0.045$ & $\bf -0.590{\pm}0.051$ & $\bf -0.589{\pm}0.050$ & $\bf -0.590{\pm}0.051$ \\
{\sc statlog-heart}  &      (270, 14)       & $\bf -0.392{\pm}0.071$ & $\bf -0.395{\pm}0.080$ & $\bf -0.394{\pm}0.080$ & $\bf -0.395{\pm}0.080$ \\
{\sc haberman-survival} &       (306, 4)       & $\bf -0.535{\pm}0.050$ & $\bf -0.537{\pm}0.054$ & $\bf -0.537{\pm}0.054$ & $\bf -0.537{\pm}0.054$ \\
  {\sc ionosphere}   &      (351, 34)       & $  -0.221{\pm}0.021  $ & $  -0.173{\pm}0.031  $ & $\bf -0.169{\pm}0.031$ & $\bf -0.169{\pm}0.033$ \\
 {\sc horse-colic}   &      (368, 26)       & $\bf -0.465{\pm}0.058$ & $\bf -0.469{\pm}0.068$ & $\bf -0.466{\pm}0.068$ & $\bf -0.469{\pm}0.068$ \\
{\sc congressional-voting} &      (435, 17)       & $\bf -0.636{\pm}0.026$ & $\bf -0.636{\pm}0.029$ & $\bf -0.636{\pm}0.029$ & $\bf -0.636{\pm}0.029$ \\
{\sc cylinder-bands} &      (512, 36)       & $  -0.482{\pm}0.030  $ & $\bf -0.442{\pm}0.038$ & $  -0.457{\pm}0.038  $ & $\bf -0.441{\pm}0.040$ \\
{\sc breast-cancer-wisc-diag} &      (569, 31)       & $\bf -0.079{\pm}0.020$ & $\bf -0.069{\pm}0.040$ & $\bf -0.071{\pm}0.038$ & $\bf -0.071{\pm}0.045$ \\
{\sc ilpd-indian-liver} &      (583, 10)       & $\bf -0.516{\pm}0.033$ & $\bf -0.516{\pm}0.037$ & $\bf -0.516{\pm}0.037$ & $\bf -0.516{\pm}0.037$ \\
   {\sc monks-2}     &       (601, 7)       & $  -0.496{\pm}0.037  $ & $\bf -0.445{\pm}0.042$ & $  -0.468{\pm}0.045  $ & $\bf -0.443{\pm}0.044$ \\
{\sc statlog-australian-credit} &      (690, 15)       & $\bf -0.627{\pm}0.022$ & $\bf -0.627{\pm}0.023$ & $\bf -0.627{\pm}0.023$ & $\bf -0.627{\pm}0.023$ \\
{\sc credit-approval} &      (690, 16)       & $\bf -0.342{\pm}0.008$ & $\bf -0.341{\pm}0.009$ & $\bf -0.341{\pm}0.008$ & $\bf -0.341{\pm}0.009$ \\
{\sc breast-cancer-wisc} &      (699, 10)       & $\bf -0.091{\pm}0.030$ & $\bf -0.091{\pm}0.037$ & $\bf -0.091{\pm}0.037$ & $\bf -0.091{\pm}0.037$ \\
    {\sc blood}      &       (748, 5)       & $\bf -0.475{\pm}0.043$ & $\bf -0.476{\pm}0.044$ & $\bf -0.476{\pm}0.044$ & $\bf -0.476{\pm}0.044$ \\
     {\sc pima}      &       (768, 9)       & $\bf -0.470{\pm}0.031$ & $\bf -0.471{\pm}0.033$ & $\bf -0.471{\pm}0.033$ & $\bf -0.471{\pm}0.033$ \\
 {\sc mammographic}  &       (961, 6)       & $\bf -0.408{\pm}0.037$ & $\bf -0.408{\pm}0.039$ & $\bf -0.408{\pm}0.039$ & $\bf -0.408{\pm}0.039$ \\
{\sc statlog-german-credit} &      (1000, 25)      & $\bf -0.493{\pm}0.040$ & $\bf -0.494{\pm}0.043$ & $\bf -0.493{\pm}0.043$ & $\bf -0.494{\pm}0.043$ \\
\midrule
    {Bold Count}     & {}& $21$           & $23$           & $22$           & $27$           \\
\bottomrule\end{tabular}

%% file: table/ep_test_acc.tex
\begin{tabular}{lc|cccc}
\toprule
{} &($n$, $d$)     &          LA         &    EP (new init.)   &          VI         &         Ours        \\
\midrule
    {\sc trains}     &       (10, 30)       & $\bf 0.700{\pm}0.400 $ & $\bf 0.800{\pm}0.245 $ & $\bf 0.800{\pm}0.245 $ & $\bf 0.800{\pm}0.245 $ \\
   {\sc balloons}    &       (16, 5)        & $\bf 0.667{\pm}0.298 $ & $\bf 0.667{\pm}0.298 $ & $\bf 0.667{\pm}0.298 $ & $\bf 0.667{\pm}0.298 $ \\
  {\sc fertility}    &      (100, 10)       & $\bf 0.880{\pm}0.060 $ & $\bf 0.880{\pm}0.060 $ & $\bf 0.880{\pm}0.060 $ & $\bf 0.880{\pm}0.060 $ \\
{\sc pittsburg-bridges-T-OR-D} &       (102, 8)       & $\bf 0.863{\pm}0.036 $ & $\bf 0.863{\pm}0.036 $ & $\bf 0.863{\pm}0.036 $ & $\bf 0.863{\pm}0.036 $ \\
{\sc acute-nephritis} &       (120, 7)       & $\bf 1.000{\pm}0.000 $ & $\bf 1.000{\pm}0.000 $ & $\bf 1.000{\pm}0.000 $ & $\bf 1.000{\pm}0.000 $ \\
{\sc acute-inflammation} &       (120, 7)       & $\bf 1.000{\pm}0.000 $ & $\bf 1.000{\pm}0.000 $ & $\bf 1.000{\pm}0.000 $ & $\bf 1.000{\pm}0.000 $ \\
{\sc echocardiogram} &      (131, 11)       & $\bf 0.808{\pm}0.061 $ & $\bf 0.778{\pm}0.063 $ & $\bf 0.785{\pm}0.076 $ & $\bf 0.778{\pm}0.063 $ \\
  {\sc hepatitis}    &      (155, 20)       & $\bf 0.813{\pm}0.024 $ & $\bf 0.832{\pm}0.024 $ & $\bf 0.813{\pm}0.024 $ & $\bf 0.832{\pm}0.024 $ \\
  {\sc parkinsons}   &      (195, 23)       & $\bf 0.959{\pm}0.048 $ & $\bf 0.959{\pm}0.048 $ & $\bf 0.944{\pm}0.050 $ & $\bf 0.959{\pm}0.048 $ \\
{\sc breast-cancer-wisc-prog} &      (198, 34)       & $\bf 0.793{\pm}0.051 $ & $\bf 0.793{\pm}0.051 $ & $\bf 0.793{\pm}0.051 $ & $\bf 0.793{\pm}0.051 $ \\
    {\sc spect}      &      (265, 23)       & $\bf 0.706{\pm}0.031 $ & $\bf 0.706{\pm}0.023 $ & $\bf 0.702{\pm}0.028 $ & $\bf 0.706{\pm}0.023 $ \\
{\sc statlog-heart}  &      (270, 14)       & $\bf 0.830{\pm}0.036 $ & $\bf 0.826{\pm}0.034 $ & $\bf 0.830{\pm}0.036 $ & $\bf 0.826{\pm}0.034 $ \\
{\sc haberman-survival} &       (306, 4)       & $\bf 0.725{\pm}0.039 $ & $\bf 0.725{\pm}0.039 $ & $\bf 0.725{\pm}0.039 $ & $\bf 0.725{\pm}0.039 $ \\
  {\sc ionosphere}   &      (351, 34)       & $\bf 0.932{\pm}0.025 $ & $\bf 0.946{\pm}0.028 $ & $\bf 0.949{\pm}0.026 $ & $\bf 0.946{\pm}0.028 $ \\
 {\sc horse-colic}   &      (368, 26)       & $\bf 0.799{\pm}0.035 $ & $\bf 0.802{\pm}0.039 $ & $  0.791{\pm}0.045   $ & $\bf 0.802{\pm}0.039 $ \\
{\sc congressional-voting} &      (435, 17)       & $\bf 0.605{\pm}0.024 $ & $\bf 0.591{\pm}0.016 $ & $\bf 0.595{\pm}0.013 $ & $\bf 0.591{\pm}0.016 $ \\
{\sc cylinder-bands} &      (512, 36)       & $  0.785{\pm}0.037   $ & $\bf 0.803{\pm}0.030 $ & $\bf 0.791{\pm}0.034 $ & $\bf 0.803{\pm}0.031 $ \\
{\sc breast-cancer-wisc-diag} &      (569, 31)       & $  0.972{\pm}0.010   $ & $\bf 0.975{\pm}0.009 $ & $\bf 0.979{\pm}0.009 $ & $\bf 0.977{\pm}0.011 $ \\
{\sc ilpd-indian-liver} &      (583, 10)       & $\bf 0.719{\pm}0.025 $ & $\bf 0.719{\pm}0.028 $ & $\bf 0.715{\pm}0.027 $ & $\bf 0.719{\pm}0.028 $ \\
   {\sc monks-2}     &       (601, 7)       & $\bf 0.740{\pm}0.049 $ & $\bf 0.762{\pm}0.045 $ & $\bf 0.744{\pm}0.046 $ & $\bf 0.762{\pm}0.045 $ \\
{\sc statlog-australian-credit} &      (690, 15)       & $\bf 0.678{\pm}0.028 $ & $\bf 0.678{\pm}0.028 $ & $\bf 0.678{\pm}0.028 $ & $\bf 0.678{\pm}0.028 $ \\
{\sc credit-approval} &      (690, 16)       & $\bf 0.859{\pm}0.026 $ & $\bf 0.862{\pm}0.026 $ & $\bf 0.864{\pm}0.024 $ & $\bf 0.862{\pm}0.026 $ \\
{\sc breast-cancer-wisc} &      (699, 10)       & $\bf 0.969{\pm}0.016 $ & $\bf 0.969{\pm}0.016 $ & $\bf 0.969{\pm}0.016 $ & $\bf 0.969{\pm}0.016 $ \\
    {\sc blood}      &       (748, 5)       & $\bf 0.785{\pm}0.042 $ & $\bf 0.786{\pm}0.044 $ & $\bf 0.786{\pm}0.044 $ & $\bf 0.786{\pm}0.044 $ \\
     {\sc pima}      &       (768, 9)       & $\bf 0.764{\pm}0.027 $ & $\bf 0.767{\pm}0.026 $ & $\bf 0.767{\pm}0.026 $ & $\bf 0.767{\pm}0.026 $ \\
 {\sc mammographic}  &       (961, 6)       & $\bf 0.824{\pm}0.019 $ & $\bf 0.823{\pm}0.019 $ & $\bf 0.823{\pm}0.019 $ & $\bf 0.823{\pm}0.019 $ \\
{\sc statlog-german-credit} &      (1000, 25)      & $\bf 0.770{\pm}0.039 $ & $\bf 0.769{\pm}0.037 $ & $\bf 0.768{\pm}0.038 $ & $\bf 0.769{\pm}0.037 $ \\
\midrule
    {Bold Count}     & {}& $25$           & $27$           & $26$           & $27$           \\
\bottomrule\end{tabular}

%% file: table/new_isotropic_test_lpd.tex
\begin{tabular}{rlc|cccc}
\toprule
{\#} &{} &($n$, $d$)     &          LA         &          EP         &          VI         &         Ours        \\
\midrule
         1           &     {\sc trains}     &       (10, 30)       & $\bf -0.695{\pm}0.011$ & $\bf -0.687{\pm}0.023$ & $\bf -0.692{\pm}0.023$ & $\bf -0.681{\pm}0.042$ \\
         2           &    {\sc balloons}    &       (16, 5)        & $\bf -0.707{\pm}0.146$ & $\bf -0.684{\pm}0.161$ & $\bf -0.711{\pm}0.239$ & $\bf -0.657{\pm}0.267$ \\
         3           &   {\sc fertility}    &      (100, 10)       & $\bf -0.379{\pm}0.099$ & $\bf -0.384{\pm}0.138$ & $\bf -0.378{\pm}0.103$ & $\bf -0.379{\pm}0.103$ \\
         4           & {\sc pittsburg-bridges-T-OR-D} &       (102, 8)       & $\bf -0.306{\pm}0.044$ & $\bf -0.316{\pm}0.060$ & $\bf -0.295{\pm}0.057$ & $\bf -0.296{\pm}0.059$ \\
         5           & {\sc acute-nephritis} &       (120, 7)       & $  -0.203{\pm}0.010  $ & $  -0.047{\pm}0.009  $ & $  -0.006{\pm}0.002  $ & $\bf -0.005{\pm}0.001$ \\
         6           & {\sc acute-inflammation} &       (120, 7)       & $  -0.172{\pm}0.033  $ & $  -0.055{\pm}0.005  $ & $  -0.008{\pm}0.001  $ & $\bf -0.007{\pm}0.001$ \\
         7           & {\sc echocardiogram} &      (131, 11)       & $\bf -0.420{\pm}0.073$ & $\bf -0.412{\pm}0.084$ & $\bf -0.420{\pm}0.085$ & $\bf -0.421{\pm}0.086$ \\
         8           &   {\sc hepatitis}    &      (155, 20)       & $\bf -0.369{\pm}0.043$ & $\bf -0.376{\pm}0.046$ & $\bf -0.361{\pm}0.051$ & $\bf -0.361{\pm}0.052$ \\
         9           &   {\sc parkinsons}   &      (195, 23)       & $  -0.267{\pm}0.030  $ & $  -0.302{\pm}0.090  $ & $  -0.164{\pm}0.062  $ & $\bf -0.145{\pm}0.054$ \\
         10          & {\sc breast-cancer-wisc-prog} &      (198, 34)       & $\bf -0.460{\pm}0.063$ & $\bf -0.478{\pm}0.083$ & $\bf -0.458{\pm}0.071$ & $\bf -0.458{\pm}0.070$ \\
         11          &     {\sc spect}      &      (265, 23)       & $\bf -0.587{\pm}0.045$ & $\bf -0.587{\pm}0.052$ & $\bf -0.589{\pm}0.050$ & $\bf -0.590{\pm}0.051$ \\
         12          & {\sc statlog-heart}  &      (270, 14)       & $\bf -0.392{\pm}0.071$ & $\bf -0.380{\pm}0.058$ & $\bf -0.394{\pm}0.080$ & $\bf -0.395{\pm}0.080$ \\
         13          & {\sc haberman-survival} &       (306, 4)       & $\bf -0.535{\pm}0.050$ & $\bf -0.539{\pm}0.059$ & $\bf -0.537{\pm}0.054$ & $\bf -0.537{\pm}0.054$ \\
         14          &   {\sc ionosphere}   &      (351, 34)       & $  -0.221{\pm}0.021  $ & $  -0.227{\pm}0.016  $ & $\bf -0.169{\pm}0.031$ & $\bf -0.169{\pm}0.033$ \\
         15          &  {\sc horse-colic}   &      (368, 26)       & $  -0.465{\pm}0.058  $ & $\bf -0.455{\pm}0.060$ & $\bf -0.466{\pm}0.068$ & $\bf -0.469{\pm}0.068$ \\
         16          & {\sc congressional-voting} &      (435, 17)       & $\bf -0.636{\pm}0.026$ & $\bf -0.633{\pm}0.027$ & $\bf -0.636{\pm}0.029$ & $\bf -0.636{\pm}0.029$ \\
         17          & {\sc cylinder-bands} &      (512, 36)       & $  -0.482{\pm}0.030  $ & $  -0.495{\pm}0.030  $ & $  -0.457{\pm}0.038  $ & $\bf -0.441{\pm}0.040$ \\
         18          & {\sc breast-cancer-wisc-diag} &      (569, 31)       & $\bf -0.079{\pm}0.020$ & $  -0.140{\pm}0.015  $ & $\bf -0.071{\pm}0.038$ & $\bf -0.071{\pm}0.045$ \\
         19          & {\sc ilpd-indian-liver} &      (583, 10)       & $\bf -0.516{\pm}0.033$ & $\bf -0.521{\pm}0.033$ & $\bf -0.516{\pm}0.037$ & $\bf -0.516{\pm}0.037$ \\
         20          &    {\sc monks-2}     &       (601, 7)       & $  -0.496{\pm}0.037  $ & $  -0.518{\pm}0.042  $ & $  -0.468{\pm}0.045  $ & $\bf -0.443{\pm}0.044$ \\
         21          & {\sc statlog-australian-credit} &      (690, 15)       & $\bf -0.627{\pm}0.022$ & $\bf -0.636{\pm}0.030$ & $\bf -0.627{\pm}0.023$ & $\bf -0.627{\pm}0.023$ \\
         22          & {\sc credit-approval} &      (690, 16)       & $\bf -0.342{\pm}0.008$ & $\bf -0.343{\pm}0.012$ & $\bf -0.341{\pm}0.008$ & $\bf -0.341{\pm}0.009$ \\
         23          & {\sc breast-cancer-wisc} &      (699, 10)       & $\bf -0.091{\pm}0.030$ & $\bf -0.092{\pm}0.028$ & $\bf -0.091{\pm}0.037$ & $\bf -0.091{\pm}0.037$ \\
         24          &     {\sc blood}      &       (748, 5)       & $\bf -0.475{\pm}0.043$ & $\bf -0.476{\pm}0.045$ & $\bf -0.476{\pm}0.044$ & $\bf -0.476{\pm}0.044$ \\
         25          &      {\sc pima}      &       (768, 9)       & $\bf -0.470{\pm}0.031$ & $\bf -0.472{\pm}0.035$ & $\bf -0.471{\pm}0.033$ & $\bf -0.471{\pm}0.033$ \\
         26          &  {\sc mammographic}  &       (961, 6)       & $\bf -0.408{\pm}0.037$ & $\bf -0.408{\pm}0.041$ & $\bf -0.408{\pm}0.039$ & $\bf -0.408{\pm}0.039$ \\
         27          & {\sc statlog-german-credit} &      (1000, 25)      & $\bf -0.493{\pm}0.040$ & $\bf -0.493{\pm}0.041$ & $\bf -0.493{\pm}0.043$ & $\bf -0.494{\pm}0.043$ \\
\midrule
{} &    {Bold Count}     & {}& $20$           & $20$           & $22$           & $27$           \\
\bottomrule\end{tabular}

%% file: table/new_isotropic_test_acc.tex
\begin{tabular}{lc|ccccc}
\toprule
{} &($n$, $d$)     &          LA         &          EP         &          VI         &         Ours        \\
\midrule
    {\sc trains}     &       (10, 30)       & $\bf 0.700{\pm}0.400 $ & $\bf 0.800{\pm}0.245 $ & $\bf 0.800{\pm}0.245 $ & $\bf 0.800{\pm}0.245 $ \\
   {\sc balloons}    &       (16, 5)        & $\bf 0.667{\pm}0.298 $ & $\bf 0.667{\pm}0.298 $ & $\bf 0.667{\pm}0.298 $ & $\bf 0.667{\pm}0.298 $ \\
  {\sc fertility}    &      (100, 10)       & $\bf 0.880{\pm}0.060 $ & $\bf 0.880{\pm}0.060 $ & $\bf 0.880{\pm}0.060 $ & $\bf 0.880{\pm}0.060 $ \\
{\sc pittsburg-bridges-T-OR-D} &       (102, 8)       & $\bf 0.863{\pm}0.036 $ & $\bf 0.883{\pm}0.066 $ & $\bf 0.863{\pm}0.036 $ & $\bf 0.863{\pm}0.036 $ \\
{\sc acute-nephritis} &       (120, 7)       & $\bf 1.000{\pm}0.000 $ & $\bf 1.000{\pm}0.000 $ & $\bf 1.000{\pm}0.000 $ & $\bf 1.000{\pm}0.000 $ \\
{\sc acute-inflammation} &       (120, 7)       & $\bf 1.000{\pm}0.000 $ & $\bf 1.000{\pm}0.000 $ & $\bf 1.000{\pm}0.000 $ & $\bf 1.000{\pm}0.000 $ \\
{\sc echocardiogram} &      (131, 11)       & $  0.808{\pm}0.061   $ & $\bf 0.847{\pm}0.043 $ & $\bf 0.785{\pm}0.076 $ & $  0.778{\pm}0.063   $ \\
  {\sc hepatitis}    &      (155, 20)       & $\bf 0.813{\pm}0.024 $ & $\bf 0.813{\pm}0.032 $ & $\bf 0.813{\pm}0.024 $ & $\bf 0.832{\pm}0.024 $ \\
  {\sc parkinsons}   &      (195, 23)       & $\bf 0.959{\pm}0.048 $ & $  0.892{\pm}0.075   $ & $\bf 0.944{\pm}0.050 $ & $\bf 0.959{\pm}0.048 $ \\
{\sc breast-cancer-wisc-prog} &      (198, 34)       & $\bf 0.793{\pm}0.051 $ & $\bf 0.793{\pm}0.069 $ & $\bf 0.793{\pm}0.051 $ & $\bf 0.793{\pm}0.051 $ \\
    {\sc spect}      &      (265, 23)       & $\bf 0.706{\pm}0.031 $ & $\bf 0.698{\pm}0.027 $ & $\bf 0.702{\pm}0.028 $ & $\bf 0.706{\pm}0.023 $ \\
{\sc statlog-heart}  &      (270, 14)       & $\bf 0.830{\pm}0.036 $ & $\bf 0.833{\pm}0.039 $ & $\bf 0.830{\pm}0.036 $ & $\bf 0.826{\pm}0.034 $ \\
{\sc haberman-survival} &       (306, 4)       & $\bf 0.725{\pm}0.039 $ & $\bf 0.719{\pm}0.036 $ & $\bf 0.725{\pm}0.039 $ & $\bf 0.725{\pm}0.039 $ \\
  {\sc ionosphere}   &      (351, 34)       & $\bf 0.932{\pm}0.025 $ & $\bf 0.932{\pm}0.024 $ & $\bf 0.946{\pm}0.024 $ & $\bf 0.946{\pm}0.028 $ \\
 {\sc horse-colic}   &      (368, 26)       & $\bf 0.799{\pm}0.035 $ & $\bf 0.807{\pm}0.032 $ & $\bf 0.791{\pm}0.045 $ & $\bf 0.802{\pm}0.039 $ \\
{\sc congressional-voting} &      (435, 17)       & $\bf 0.605{\pm}0.024 $ & $\bf 0.593{\pm}0.014 $ & $\bf 0.595{\pm}0.013 $ & $\bf 0.591{\pm}0.016 $ \\
{\sc cylinder-bands} &      (512, 36)       & $  0.785{\pm}0.037   $ & $\bf 0.777{\pm}0.044 $ & $\bf 0.791{\pm}0.034 $ & $\bf 0.803{\pm}0.031 $ \\
{\sc breast-cancer-wisc-diag} &      (569, 31)       & $\bf 0.972{\pm}0.010 $ & $\bf 0.974{\pm}0.019 $ & $\bf 0.977{\pm}0.012 $ & $\bf 0.977{\pm}0.011 $ \\
{\sc ilpd-indian-liver} &      (583, 10)       & $\bf 0.719{\pm}0.025 $ & $\bf 0.712{\pm}0.020 $ & $\bf 0.717{\pm}0.029 $ & $\bf 0.719{\pm}0.028 $ \\
   {\sc monks-2}     &       (601, 7)       & $\bf 0.740{\pm}0.049 $ & $  0.719{\pm}0.039   $ & $\bf 0.744{\pm}0.046 $ & $\bf 0.762{\pm}0.045 $ \\
{\sc statlog-australian-credit} &      (690, 15)       & $\bf 0.678{\pm}0.028 $ & $\bf 0.677{\pm}0.021 $ & $\bf 0.678{\pm}0.028 $ & $\bf 0.678{\pm}0.028 $ \\
{\sc credit-approval} &      (690, 16)       & $\bf 0.859{\pm}0.026 $ & $\bf 0.864{\pm}0.020 $ & $\bf 0.864{\pm}0.024 $ & $\bf 0.862{\pm}0.026 $ \\
{\sc breast-cancer-wisc} &      (699, 10)       & $\bf 0.969{\pm}0.016 $ & $\bf 0.969{\pm}0.016 $ & $\bf 0.969{\pm}0.016 $ & $\bf 0.969{\pm}0.016 $ \\
    {\sc blood}      &       (748, 5)       & $\bf 0.785{\pm}0.042 $ & $\bf 0.785{\pm}0.041 $ & $\bf 0.786{\pm}0.044 $ & $\bf 0.786{\pm}0.044 $ \\
     {\sc pima}      &       (768, 9)       & $\bf 0.764{\pm}0.027 $ & $\bf 0.766{\pm}0.025 $ & $\bf 0.767{\pm}0.026 $ & $\bf 0.767{\pm}0.026 $ \\
 {\sc mammographic}  &       (961, 6)       & $\bf 0.824{\pm}0.019 $ & $\bf 0.823{\pm}0.017 $ & $\bf 0.823{\pm}0.019 $ & $\bf 0.823{\pm}0.019 $ \\
{\sc statlog-german-credit} &      (1000, 25)      & $\bf 0.770{\pm}0.039 $ & $\bf 0.763{\pm}0.038 $ & $\bf 0.768{\pm}0.038 $ & $\bf 0.769{\pm}0.037 $ \\
\midrule
    {Bold Count}     & {}& $25$           & $25$           & $27$           & $26$           \\
\bottomrule\end{tabular}

%% file: table/sparse_dataset.tex
%

\begin{tabular}{l|ccccc}
\toprule
Data set &
{\sc titanic}     &  
{\sc bank}      & 
{\sc twonorm}     &
{\sc mushroom}    & 
{\sc magic}      \\
\midrule
($n$, $d$) 
&      (2201, 4)       
&      (4521, 17)      
&      (7400, 21)      
&      (8124, 22)      
&     (19020, 11) \\
\bottomrule
\end{tabular}

%% file: arxiv.bbl
\begin{thebibliography}{37}
\providecommand{\natexlab}[1]{#1}
\providecommand{\url}[1]{\texttt{#1}}
\expandafter\ifx\csname urlstyle\endcsname\relax
  \providecommand{\doi}[1]{doi: #1}\else
  \providecommand{\doi}{doi: \begingroup \urlstyle{rm}\Url}\fi

\bibitem[Adam et~al.(2021)Adam, Chang, Khan, and Solin]{dual}
Adam, V., Chang, P.~E., Khan, M.~E., and Solin, A.
\newblock Dual parameterization of sparse variational {G}aussian processes.
\newblock In \emph{Advances in Neural Information Processing Systems 34
  (NeurIPS)}, volume~34, pp.\  11474--11486. Curran Associates, Inc., 2021.

\bibitem[Amari(1998)]{amari1998natural}
Amari, S.-I.
\newblock Natural gradient works efficiently in learning.
\newblock \emph{Neural Computation}, 10\penalty0 (2):\penalty0 251--276, 1998.

\bibitem[Bui et~al.(2017)Bui, Yan, and Turner]{powerep}
Bui, T.~D., Yan, J., and Turner, R.~E.
\newblock A unifying framework for {G}aussian process pseudo-point
  approximations using power expectation propagation.
\newblock \emph{Journal of Machine Learning Research (JMLR)}, 18:\penalty0
  3649--3720, 2017.

\bibitem[Chang et~al.(2020)Chang, Wilkinson, Khan, and Solin]{chang2020fast}
Chang, P.~E., Wilkinson, W.~J., Khan, M.~E., and Solin, A.
\newblock Fast variational learning in state-space {G}aussian process models.
\newblock In \emph{International Workshop on Machine Learning for Signal
  Processing (MLSP)}. IEEE, 2020.

\bibitem[Csat{\'o}(2002)]{csato2002gaussian}
Csat{\'o}, L.
\newblock \emph{Gaussian Processes: {I}terative Sparse Approximations}.
\newblock PhD thesis, Aston University, Birmingham, UK, 2002.

\bibitem[Dua \& Graff(2017)Dua and Graff]{UCI}
Dua, D. and Graff, C.
\newblock {UCI} machine learning repository, 2017.
\newblock \url{http://archive.ics.uci.edu/ml}.

\bibitem[Gorinova et~al.(2020)Gorinova, Moore, and Hoffman]{gorinova20a}
Gorinova, M., Moore, D., and Hoffman, M.
\newblock Automatic reparameterisation of probabilistic programs.
\newblock In \emph{Proceedings of the 37th International Conference on Machine
  Learning}, volume 119 of \emph{Proceedings of Machine Learning Research},
  pp.\  3648--3657. PMLR, 2020.

\bibitem[{GPy}(since 2012)]{gpy2014}
{GPy}.
\newblock {GPy}: A {G}aussian process framework in python.
\newblock \url{http://github.com/SheffieldML/GPy}, since 2012.

\bibitem[Hensman et~al.(2013)Hensman, Fusi, and Lawrence]{hensman2013gaussian}
Hensman, J., Fusi, N., and Lawrence, N.~D.
\newblock Gaussian processes for big data.
\newblock In \emph{Proceedings of the 29th Conference on Uncertainty in
  Artificial Intelligence (UAI)}, pp.\  282--290. AUAI Press, 2013.

\bibitem[Jyl{{\"a}}nki et~al.(2011)Jyl{{\"a}}nki, Vanhatalo, and Vehtari]{aki}
Jyl{{\"a}}nki, P., Vanhatalo, J., and Vehtari, A.
\newblock Robust {G}aussian process regression with a {S}tudent-$t$ likelihood.
\newblock \emph{Journal of Machine Learning Research (JMLR)}, 12:\penalty0
  3227--3257, 2011.

\bibitem[Khan \& Lin(2017)Khan and Lin]{cvi}
Khan, M.~E. and Lin, W.
\newblock Conjugate-computation variational inference: Converting variational
  inference in non-conjugate models to inferences in conjugate models.
\newblock In \emph{Proceedings of the 20th International Conference on
  Artificial Intelligence and Statistics (AISTATS)}, volume~54 of
  \emph{Proceedings of Machine Learning Research}, pp.\  878--887. PMRL, 2017.

\bibitem[Khan et~al.(2013)Khan, Aravkin, Friedlander, and Seeger]{khan2013fast}
Khan, M.~E., Aravkin, A., Friedlander, M., and Seeger, M.
\newblock Fast dual variational inference for non-conjugate latent {G}aussian
  models.
\newblock In \emph{Proceedings of the 30th International Conference on Machine
  Learning (ICML)}, volume~28 of \emph{Proceedings of Machine Learning
  Research}, pp.\  951--959. PMLR, 2013.

\bibitem[Kingma \& Ba(2015)Kingma and Ba]{adam}
Kingma, D.~P. and Ba, J.
\newblock Adam: A method for stochastic optimization.
\newblock In \emph{International Conference on Learning Representations
  (ICLR)}, 2015.

\bibitem[Kuss \& Rasmussen(2005)Kuss and Rasmussen]{05classification}
Kuss, M. and Rasmussen, C.~E.
\newblock Assessing approximate inference for binary {G}aussian process
  classification.
\newblock \emph{Journal of Machine Learning Research (JMLR)}, 6:\penalty0
  1679--1704, 2005.

\bibitem[Li et~al.(2015)Li, Hern\'{a}ndez-Lobato, and Turner]{li2015stochastic}
Li, Y., Hern\'{a}ndez-Lobato, J.~M., and Turner, R.~E.
\newblock Stochastic expectation propagation.
\newblock In \emph{Advances in Neural Information Processing Systems 28}, pp.\
  2323--2331. Curran Associates, Inc., 2015.

\bibitem[Lotfi et~al.(2022)Lotfi, Izmailov, Benton, Goldblum, and
  Wilson]{lotfi2022bayesian}
Lotfi, S., Izmailov, P., Benton, G., Goldblum, M., and Wilson, A.~G.
\newblock {B}ayesian model selection, the marginal likelihood, and
  generalization.
\newblock In \emph{Proceedings of the 39th International Conference on Machine
  Learning (ICML)}, volume 162 of \emph{Proceedings of Machine Learning
  Research}, pp.\  14223--14247. PMLR, 2022.

\bibitem[Matthews et~al.(2017)Matthews, {van der Wilk}, Nickson, Fujii,
  {Boukouvalas}, {Le{\'o}n-Villagr{\'a}}, Ghahramani, and Hensman]{GPflow:2017}
Matthews, A. G. d.~G., {van der Wilk}, M., Nickson, T., Fujii, K.,
  {Boukouvalas}, A., {Le{\'o}n-Villagr{\'a}}, P., Ghahramani, Z., and Hensman,
  J.
\newblock {GP}flow: A {G}aussian process library using {T}ensor{F}low.
\newblock \emph{Journal of Machine Learning Research (JMLR)}, 18\penalty0
  (40):\penalty0 1--6, 2017.

\bibitem[Minka(2001)]{EP}
Minka, T.~P.
\newblock Expectation propagation for approximate {B}ayesian inference.
\newblock In \emph{Proceedings of the 17th Conference on Uncertainty in
  Artificial Intelligence (UAI)}, Proceedings of Machine Learning Research,
  pp.\  362--369, 2001.

\bibitem[Murray et~al.(2010)Murray, Adams, and MacKay]{murray2010elliptical}
Murray, I., Adams, R., and MacKay, D.
\newblock Elliptical slice sampling.
\newblock In \emph{Proceedings of the Thirteenth International Conference on
  Artificial Intelligence and Statistics (AISTATS)}, pp.\  541--548. JMLR
  Workshop and Conference Proceedings, 2010.

\bibitem[Neal(2001)]{AIS}
Neal, R.~M.
\newblock Annealed importance sampling.
\newblock \emph{Statistics and Computing}, 11:\penalty0 125--139, 2001.

\bibitem[Neal(2022)]{odata}
Neal, R.~M.
\newblock Software for flexible {B}ayesian modeling and {M}arkov {C}hain
  sampling, 2022.
\newblock \url{http://www.cs.toronto.edu/~radford/fbm.software.html}.

\bibitem[Nickisch \& Rasmussen(2008)Nickisch and Rasmussen]{08classification}
Nickisch, H. and Rasmussen, C.~E.
\newblock Approximations for binary {G}aussian process classification.
\newblock \emph{Journal of Machine Learning Research (JMLR)}, 9:\penalty0
  2035--2078, 2008.

\bibitem[Opper \& Archambeau(2009)Opper and Archambeau]{VI}
Opper, M. and Archambeau, C.
\newblock The variational {G}aussian approximation revisited.
\newblock \emph{Neural Computation}, 21\penalty0 (3):\penalty0 786--92, 2009.

\bibitem[Qui{\~{n}}onero-Candela \& Rasmussen(2005)Qui{\~{n}}onero-Candela and
  Rasmussen]{05overview}
Qui{\~{n}}onero-Candela, J. and Rasmussen, C.~E.
\newblock A unifying view of sparse approximate {G}aussian process regression.
\newblock \emph{Journal of Machine Learning Research (JMLR)}, 6:\penalty0
  1939--1959, 2005.

\bibitem[Rasmussen \& Nickisch(2010)Rasmussen and
  Nickisch]{rasmussen2010gaussian}
Rasmussen, C.~E. and Nickisch, H.
\newblock Gaussian processes for machine learning ({GPML}) toolbox.
\newblock \emph{Journal of Machine Learning Research (JMLR)}, 11:\penalty0
  3011--3015, 2010.

\bibitem[Rasmussen \& Williams(2006)Rasmussen and Williams]{gpbook}
Rasmussen, C.~E. and Williams, C. K.~I.
\newblock \emph{{G}aussian Processes for Machine Learning}.
\newblock {MIT} Press, 2006.

\bibitem[Salimbeni et~al.(2018)Salimbeni, Eleftheriadis, and
  Hensman]{salimbeni2018natural}
Salimbeni, H., Eleftheriadis, S., and Hensman, J.
\newblock Natural gradients in practice: {N}on-conjugate variational inference
  in {G}aussian process models.
\newblock In \emph{Proceedings of the Twenty-First International Conference on
  Artificial Intelligence and Statistics (AISTATS)}, volume~84 of
  \emph{Proceedings of Machine Learning Research}, pp.\  689--697. PMLR, 2018.

\bibitem[Seeger(2003)]{seeger2003bayesian}
Seeger, M.
\newblock \emph{Bayesian {G}aussian Process Models: {PAC}-{B}ayesian
  Generalisation Error Bounds and Sparse Approximations}.
\newblock PhD thesis, University of Edinburgh, Edinburgh, UK, 2003.

\bibitem[Solin \& S\"arkk\"a(2015)Solin and S\"arkk\"a]{solin2015state}
Solin, A. and S\"arkk\"a, S.
\newblock State space methods for efficient inference in {S}tudent-$t$ process
  regression.
\newblock In \emph{Proceedings of the Eighteenth International Conference on
  Artificial Intelligence and Statistics (AISTATS)}, volume~38 of
  \emph{Proceedings of Machine Learning Research}, pp.\  885--893. PMLR, 2015.

\bibitem[Titsias(2009)]{VFE}
Titsias, M.
\newblock Variational learning of inducing variables in sparse {Gaussian}
  processes.
\newblock In \emph{Proceedings of the Twelth International Conference on
  Artificial Intelligence and Statistics}, volume~5 of \emph{Proceedings of
  Machine Learning Research}, pp.\  567--574, Hilton Clearwater Beach Resort,
  Clearwater Beach, Florida USA, 16--18 Apr 2009. PMLR.

\bibitem[Vanhatalo et~al.(2013)Vanhatalo, Riihim{\"a}ki, Hartikainen,
  Jyl{\"a}nki, Tolvanen, and Vehtari]{vanhatalo2013gpstuff}
Vanhatalo, J., Riihim{\"a}ki, J., Hartikainen, J., Jyl{\"a}nki, P., Tolvanen,
  V., and Vehtari, A.
\newblock {GPstuff}: {B}ayesian modeling with {G}aussian processes.
\newblock \emph{Journal of Machine Learning Research (JMLR)}, 14\penalty0
  (1):\penalty0 1175--1179, 2013.

\bibitem[Vapnick(1998)]{vapnick1998statistical}
Vapnick, V.~N.
\newblock \emph{Statistical Learning Theory}.
\newblock Wiley, New York, 1998.

\bibitem[Vehtari et~al.(2016)Vehtari, Mononen, Tolvanen, Sivula, and
  Winther]{vehtari2016bayesian}
Vehtari, A., Mononen, T., Tolvanen, V., Sivula, T., and Winther, O.
\newblock Bayesian leave-one-out cross-validation approximations for {G}aussian
  latent variable models.
\newblock \emph{Journal of Machine Learning Research (JMLR)}, 17\penalty0
  (1):\penalty0 3581--3618, 2016.

\bibitem[Vehtari et~al.(2020)Vehtari, Gelman, Sivula, Jylänki, Tran, Sahai,
  Blomstedt, Cunningham, Schiminovich, and Robert]{epasawayoflife}
Vehtari, A., Gelman, A., Sivula, T., Jylänki, P., Tran, D., Sahai, S.,
  Blomstedt, P., Cunningham, J.~P., Schiminovich, D., and Robert, C.~P.
\newblock Expectation propagation as a way of life.
\newblock \emph{Journal of Machine Learning Research (JMLR)}, 21:\penalty0
  1--53, 2020.

\bibitem[Wilkinson et~al.(2023)Wilkinson, S{\"a}rkk{\"a}, and
  Solin]{wilkinson2021bayes}
Wilkinson, W.~J., S{\"a}rkk{\"a}, S., and Solin, A.
\newblock {Bayes}--{Newton} methods for approximate {B}ayesian inference with
  {PSD} guarantees.
\newblock \emph{Journal of Machine Learning Research (JMLR)}, 24\penalty0
  (83):\penalty0 1--50, 2023.

\bibitem[Williams \& Barber(1998)Williams and Barber]{LA}
Williams, C.~K. and Barber, D.
\newblock Bayesian classification with {Gaussian} processes.
\newblock \emph{IEEE Transactions on Pattern Analysis and Machine
  Intelligence}, 20\penalty0 (12):\penalty0 1342--1351, 1998.

\bibitem[Williams \& Seeger(2001)Williams and Seeger]{williams2001using}
Williams, C.~K. and Seeger, M.
\newblock Using the {N}ystr{\"o}m method to speed up kernel machines.
\newblock In \emph{Advances in Neural Information Processing Systems 13
  (NIPS)}, pp.\  682--688. MIT Press, 2001.

\end{thebibliography}
